\documentclass[12pt, a4paper, openany]{article}
\usepackage[utf8]{inputenc}
\usepackage[affil-it]{authblk}
\usepackage[T1]{fontenc}
\usepackage[usestackEOL]{stackengine}
\usepackage{algorithm} 
\usepackage{algpseudocode} 
\usepackage[dvipsnames]{xcolor}

\usepackage{fullpage}
\usepackage{eso-pic}
\usepackage{setspace}
\usepackage{subfig}
\usepackage[titletoc]{appendix}
\usepackage{type1cm}
\usepackage{lettrine}
\usepackage{float}
\usepackage{amsmath}
\usepackage{amsfonts}
\usepackage{comment}
\usepackage{adjustbox}
\usepackage{listings}
\usepackage{xcolor}
\definecolor{codegreen}{rgb}{0,0.6,0}
\definecolor{codegray}{rgb}{0.5,0.5,0.5}
\definecolor{codepurple}{rgb}{0.58,0,0.82}
\definecolor{backcolour}{rgb}{0.95,0.95,0.92}
\lstdefinestyle{mystyle}{
    backgroundcolor=\color{backcolour},   
    commentstyle=\color{codegreen},
    keywordstyle=\color{magenta},
    numberstyle=\tiny\color{codegray},
    stringstyle=\color{codepurple},
    basicstyle=\ttfamily\footnotesize,
    breakatwhitespace=false,         
    breaklines=true,                 
    captionpos=b,                    
    keepspaces=true,                 
    numbers=left,                    
    numbersep=5pt,                  
    showspaces=false,                
    showstringspaces=false,
    showtabs=false,                  
    tabsize=2
}
\usepackage[numbers]{natbib}
\usepackage{graphicx}
\usepackage{amsmath,amssymb,amsfonts}
\usepackage{verbatim}
\usepackage{geometry}
\usepackage{float}
\usepackage{hyperref}
\usepackage{neuralnetwork}
\usepackage{algorithm}
\usepackage{xcolor}
\usepackage{tikz}
\usetikzlibrary{shapes,backgrounds,calc,positioning} 
\usepackage{amsmath} 
\usepackage{tikz}
\usepackage{listofitems} 
\usetikzlibrary{arrows.meta} 
\usepackage[outline]{contour} 
\contourlength{1.4pt}
\setcitestyle{square}
\tikzset{>=latex} 
\usepackage{xcolor}
\colorlet{myred}{red!80!black}
\colorlet{myblue}{blue!80!black}
\colorlet{mygreen}{green!60!black}
\colorlet{myorange}{orange!70!red!60!black}
\colorlet{mydarkred}{red!30!black}
\colorlet{mydarkblue}{blue!40!black}
\colorlet{mydarkgreen}{green!30!black}
\tikzstyle{node}=[thick,circle,draw=myblue,minimum size=22,inner sep=0.5,outer sep=0.6]
\tikzstyle{node in}=[node,green!20!black,draw=mygreen!30!black,fill=mygreen!25]
\tikzstyle{node hidden}=[node,blue!20!black,draw=myblue!30!black,fill=myblue!20]
\tikzstyle{node convol}=[node,orange!20!black,draw=myorange!30!black,fill=myorange!20]
\tikzstyle{node out}=[node,red!20!black,draw=myred!30!black,fill=myred!20]
\tikzstyle{connect}=[thick,mydarkblue] 
\tikzstyle{connect arrow}=[-{Latex[length=4,width=3.5]},thick,mydarkblue,shorten <=0.5,shorten >=1]
\tikzset{ 
  node 1/.style={node in},
  node 2/.style={node hidden},
  node 3/.style={node out},
}

\usepackage{notoccite}

\begin{document}

\title{Unsupervised physics-informed neural network in reaction-diffusion biology models (Ulcerative colitis and Crohn's disease cases) 

A preliminary study}

\author[2]{Ahmed Rebai}

\author[2]{Louay Boukhris}

\author[2]{Radhi Toujani}

\author[2]{Ahmed Gueddiche}

\author[2]{Fayad Ali Banna}

\author[2]{Fares Souissi}

\author[2]{Ahmed Lasram}

\author[2]{Elyes Ben Rayana}

\author[1]{\textbf{Hatem Zaag}}

\affil[1]{Université Sorbonne Paris Nord, LAGA, CNRS (UMR 7539), F-93430, Villetaneuse, France.}

\affil[2]{AI Factory, Value Digital Services, Tunisia, "value.com.tn".}

\maketitle

\setlength{\baselineskip}{13pt plus.2pt}

\begin{abstract}
We propose to explore the potential of physics-informed neural networks (PINNs) in solving a class of partial differential equations (PDEs) used to model the propagation of chronic inflammatory bowel diseases, such as Crohn's disease and ulcerative colitis. An unsupervised approach was privileged during the deep neural network training. Given the complexity of the underlying biological system, characterized by intricate feedback loops and limited availability of high-quality data, the aim of this study is to explore the potential of PINNs in solving PDEs. In addition to providing this exploratory assessment, we also aim to emphasize the principles of reproducibility and transparency in our approach, with a specific focus on ensuring the robustness and generalizability through the use of artificial intelligence. We will quantify the relevance of the PINN method with several linear and non-linear PDEs in relation to biology. However, it is important to note that the final solution is dependent on the initial conditions, chosen boundary conditions, and neural network architectures. \footnote{This work was carried out under the scientific direction of the mathematician Pr. Hatem Zaag.} \footnote{Corresponding author: Ahmed Rebai, ahmed.rebai@value.com.tn}
\end{abstract}


\newpage
\tableofcontents

\section{Introduction}
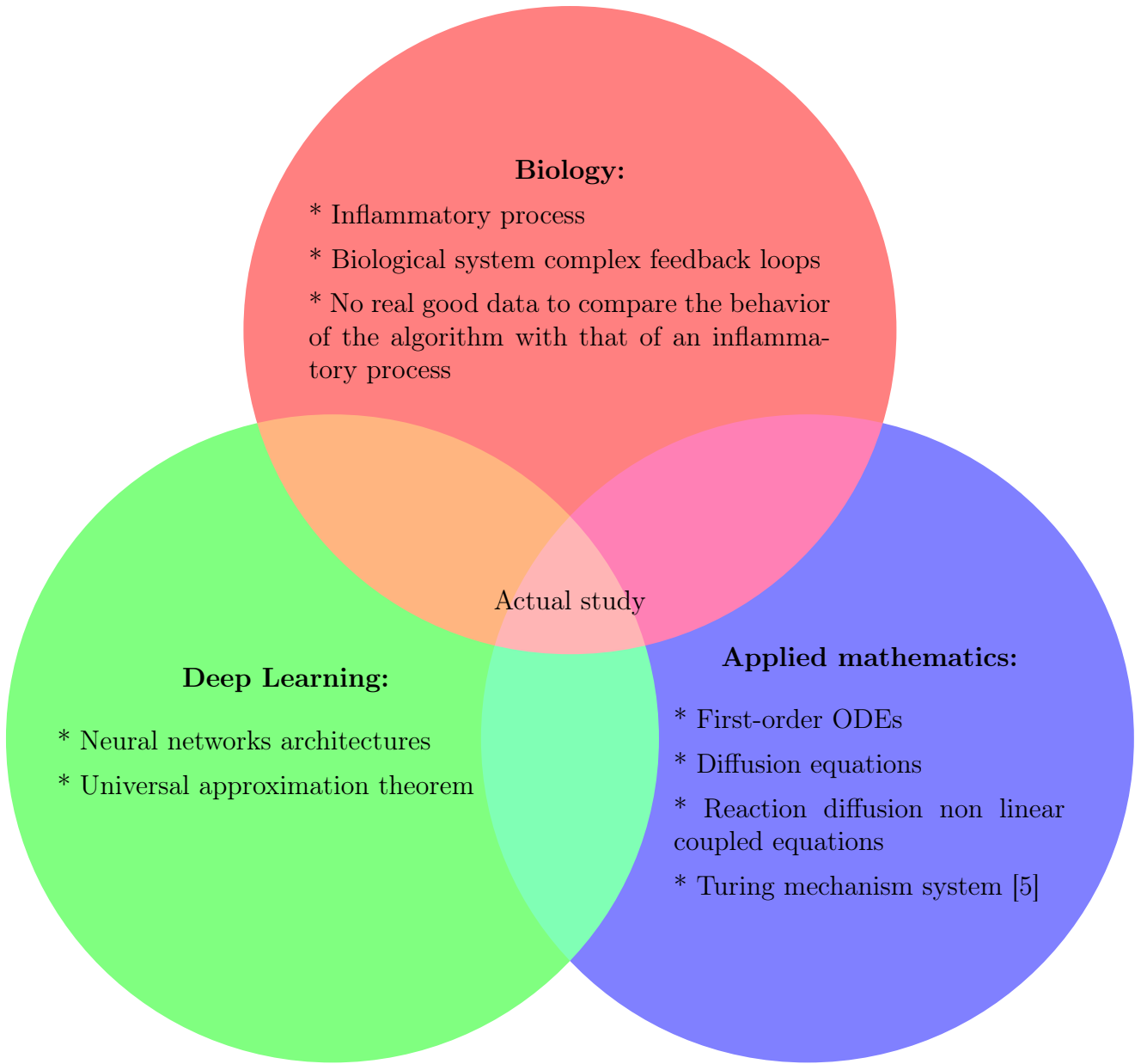
\begin{figure}[H]
\begin{center}
\begin{tikzpicture}[
roundnode/.style={circle, minimum size=7mm},
squarednode/.style={rectangle, draw=red!60, fill=red!5, very thick, minimum size=5mm},
]
  \begin{scope}[blend group = soft light]
    \fill[red!50!white]   ( 90:4.2) circle (5);
    \fill[green!50!white] (210:4.2) circle (5);
    \fill[blue!50!white]  (330:4.2) circle (5);
  \end{scope}

  \node[roundnode] at (90:5.1) (a)  {
    \begin{minipage}{8cm}
      \begin{center}
        \textbf{Biology:}
      \end{center}  
      \item * Inflammatory process
      \item * Biological system complex feedback loops
      \item * No real good data to compare the behavior of the algorithm with that of an inflammatory process
    \end{minipage}
  }
  ;
 
  \node[roundnode] at (205:4.8) (b)  {
    \begin{minipage}{7cm}
      \begin{center}
        \textbf{Deep Learning:}
      \end{center}  
      \item * Neural networks architectures
      \item * Universal approximation theorem
    \end{minipage}
  }
  ;
  
  \node[roundnode] at (330:5.3) (c)  {
    \begin{minipage}{6cm}
      \begin{center}
        \textbf{Applied mathematics:}
      \end{center}  
      \item * First-order ODEs
      \item * Diffusion equations
      \item * Reaction diffusion non linear coupled equations
      \item * Turing mechanism system \citep{Nadin2021}
    \end{minipage}
  }
  ;
  
  \node {Actual study};
\end{tikzpicture}

\end{center}
\caption{This figure illustrates the interplay between deep neural networks and differential equation models in this multidisciplinary study. Deep neural networks are capable of handling large amounts of data but can be computationally moderate, while differential equation models on the other hand, require less data but may be computationally intensive. The aim of the study is to leverage the advantages of both approaches to solve partial differential equations modeling biological phenomena using deep learning techniques, minimizing data requirements and obtaining a computationally efficient model.}
\label{fig:intersection}
\end{figure}

The current work focuses on a new multidisciplinary field at the intersection of three disciplines: artificial intelligence (AI) via deep learning, applied mathematics via partial differential equations, and the biology of the inflammatory bowel diseases (as illustrated in Figure \ref{fig:intersection}). While writing this paper, we encountered several difficulties due to the unique nature of this multidisciplinary subject, which is both innovative and AI hyped, resulting in a genuine debate in the community between partisans who are optimistic about the potential of this new technique \citep{Cuomo2022, KARNIADAKIS2021} and non-partisans who point out its limitations \citep{NEURIPS2021_FAILURE_MODES, Fuks_2020}. For this, we believe it is best to begin with a brief overview of each discipline before discussing progress at the various intersections between these disciplines, followed by a discussion of the resulting controversy.

\section{PINN in biology: crossroads of several disciplines}
\subsection{Artificial intelligence}
In its general definition, \textbf{artificial intelligence} allows computers to partially or totally perform intelligent tasks usually associated with human intelligence. Nowadays, artificial intelligence learns and generalizes patterns in high-dimensional and highly non-linear spaces without being specifically guided \citep{LeCun2015, Wang2020}. This learning processes is based on various types of data (tabular data, image data, sound data, text data...) and is leading to success in various fields such as nuclear energy where AI has been used recently to control a fusion reactor \citep{Degrave2022} and Earth science where AI allowed to predict the weather in short-term within the "nowcasting" project with a deep reinforcement learning algorithm \citep{Ravuri2021}. Continuing with this progress, in this paper we will see how neural networks could also learn the dynamics of a complex biological system from the structure of the partial differential equations describing these dynamics.

\subsection{Partial differential equations}
Many engineering fields use \textbf{partial differential equations} as models, including combustion theory, weather prediction, financial markets and industrial machine design. A partial differential equation or a system of partial differential equations can be solved analytically \citep{analytical-resol-review}, numerically \citep{numerical-resol-review1, numerical-resol-review2} and now with artificial intelligence using techniques such as deep neural networks (DNNs) \citep{Cuomo2022, ai-resol-review1}. In practice, the analytical method works for some simple equations, but its application is difficult in most cases of coupled and nonlinear PDE systems. The numerical resolution technique is preferred for this and often requires the use of expensive commercial numerical solvers such as finite element method (FEM) or finite difference method (FDM). It can be summarized in five steps: modeling, meshing, discretizing, numerical computing and post-processing:

\begin{itemize}
    \item \textbf{Modeling}: Mathematical modeling of physical or chemical or biological processes.
    \item \textbf{Mesh}: The creation of a mesh or a grid called also computational domain which consists of equivalent system of multiple sub-domains (finite differences or elements or volumes). Given the mesh, the basis functions are predetermined. This step is characterized by its great temporal complexity. At this level, the PINN method could offer a solution to reduce the execution time since it only requires faster random sampling of the working domain.
    \item \textbf{Discretization}: Discretize the governing equations by turning it into a system of equations simply by approximating the derivatives. The used functions are linear (i.e polynomial functions) which sometimes does not capture the non-linear character of the underlying phenomena. Given the non-linear nature of neural networks, they could help overcome this deficiency.
   \item \textbf{Solution}: Solving the set of linear equations by numerical computing using extensive parallel IT ressources like CPUs, GPUs and large RAMs. 
   \item \textbf{Postprocess}: Finding the desired quantities like position or velocity by analyzing the obtained data. The use of machine learning models can make this analysis more refined and robust.  
\end{itemize}

As previously stated, numerical methods have some drawbacks such as high time consumption, repetitiveness and lack of autonomy. In fact, creating a mesh to simulate an airplane turbo-reactor can take months in some industrial cases. Also, numerical solving is repetitive because the 5 steps must be reproduced each time the domain is changed. Furthermore, unlike AI models, this procedure does not learn from previous trials even if we keep the same domain or the same grid. However, the basis functions do not always allow for the reflection of non-linear phenomena that cause real or artificial blow-up phenomena such as numerical explosion \citep{Merle2021}. Finally, when we consider how difficult it is to reduce human intervention, it is clear that this technique lacks the autonomy sought during the normal use of artificial intelligence. As a result, several unanswered questions may arise, such as: Is it possible for solvers to learn the basis functions from partial differential equations automatically? Is it feasible to develop autonomous flow solvers for fluid mechanics?

\subsection{Biology of the inflammatory bowel diseases}
Now let's move to \textbf{biology} which is the science of living organisms extending from the molecular level to the mesoscopic ecosystems. In this paper, we focus on the modelling of the inflammatory process hitting the bowel. Crohn's disease and ulcerative colitis are both inflammatory bowel diseases but they are different indeed \citep{LeBerre2020, PMID24393595}. Ulcerative colitis (UC) is a chronic inflammatory bowel disease resulting from an overreaction of the natural defenses of the digestive immune system, with an estimated prevalence of 1 in 1500 people with an annual incidence of 6 to 8 new cases per 100,000 inhabitants in Australia \citep{busingye2021prevalence}, Western Europe and the United States. In Tunisia, the incidence is estimated at 2.11 per 100,000 inhabitants per year \citep{Mosli2021, Karoui2009}. UC is not a rare disease in tunisian adults, but in children. It is characterized by smooth ulceration of the inner lining of the colon. The inflammation begins in the  lower region of the colon, just above the anus, and progresses upward at varying distances. One of the most important indicators of the severity of this disease is the spatial distribution of the intestinal lesions associated with an introduced gastro-enterologist's severity score. While individuals with moderate to high severity scores have a concentration of lesions around the rectum, those with low severity scores frequently have a homogeneous spatial distribution of colonic lesions. UC appears in lesions such as bleeding rectal and colon ulcers. It is a currently incurable disease characterized by varying intensities of inflammatory relapse with interspersed  remission periods. This puts the patient at higher risk of colon cancer than the general population thus the potential removal of the organ (colectomy). Currently available treatments  aim to control pain, reduce the frequency and duration of relapses, and thereby relieve symptoms. Crohn's disease is a type of painful inflammatory bowel disease (IBD) that is not well understood. In Tunisia, this serious disease affects both children over 10 and adults \citep{Mosli2021}. It consists of the appearance of several asymmetrical segments of deep lesions separated by intact areas. In the worst cases, these areas can turn into fissures or even holes in the wall of the intestine. Unlike other IBDs, it affects any part of the gastrointestinal tract, from top (the mouth) to bottom (the anus), in contiguous or isolated parts. The inflammation can affect the inner lining and even go beyond the entire thickness of the intestinal wall; It is manifested by a blood vessels dilation and tissues fluid loss. It is usually present in the lower part of the small intestine that connects to the colon. The inflamed portion of the intestine affects the deep panniculus and is not adjacent to it, but rather is distributed throughout the gastrointestinal tract, with an erratic inflammation pattern. The diagnosis of this disease requires advanced technological tools which present difficulties in the collection of data to predict the spread. For that, the mathematical modeling has been increasingly utilized as a tool to understand the complex and dynamic processes involved in both diseases as shown in \citep{Nadin2021, Collard2021}. 

In the evaluation and management of both Crohn's disease and ulcerative colitis, doctors typically use a combination of biological, clinical, and spatial indicators to assess a patient's condition, predict its progression, and determine the most appropriate treatment. Clinical indicators may include a physician's examination and questioning of the patient, as well as video examination of the colon through colonoscopy. Biopsy samples taken during colonoscopy can also provide valuable histological images. In addition, biological or chemical indicators such as the measurement of calprotectin levels in stool (as an indicator of inflammation) and analysis of the intestinal microbiota through DNA and RNA analysis can provide important insights into the disease. Additionally, analysis of RNA expression in the intestine can also be used as an indicator.

\begin{table}[H]
\centering
\begin{tabular}{|c|c|c|}
\hline
\textbf{Data type} & \textbf{Data requirements} & \textbf{Tasks} \\
\hline
\multicolumn{1}{|c|}{Clinical data}   & Doctor's questionnaires                     & Classification         \\ 
\hline
\multicolumn{1}{|c|}{Biological data} & Physico-clinical analysis & Scoring and Classification \\ 
\hline
\multicolumn{1}{|c|}{Images and videos data}   & Computer vision treatment                             & Classification and PDE                    \\ \hline
\end{tabular}
\caption{A summary of clinical, biological, and image/video data characterization in IBDs medical tasks.}
\label{table:table1}
\end{table}




\subsection{The importance of spatial information}
\begin{figure}[H]
   \centering
    \includegraphics[scale=0.45]{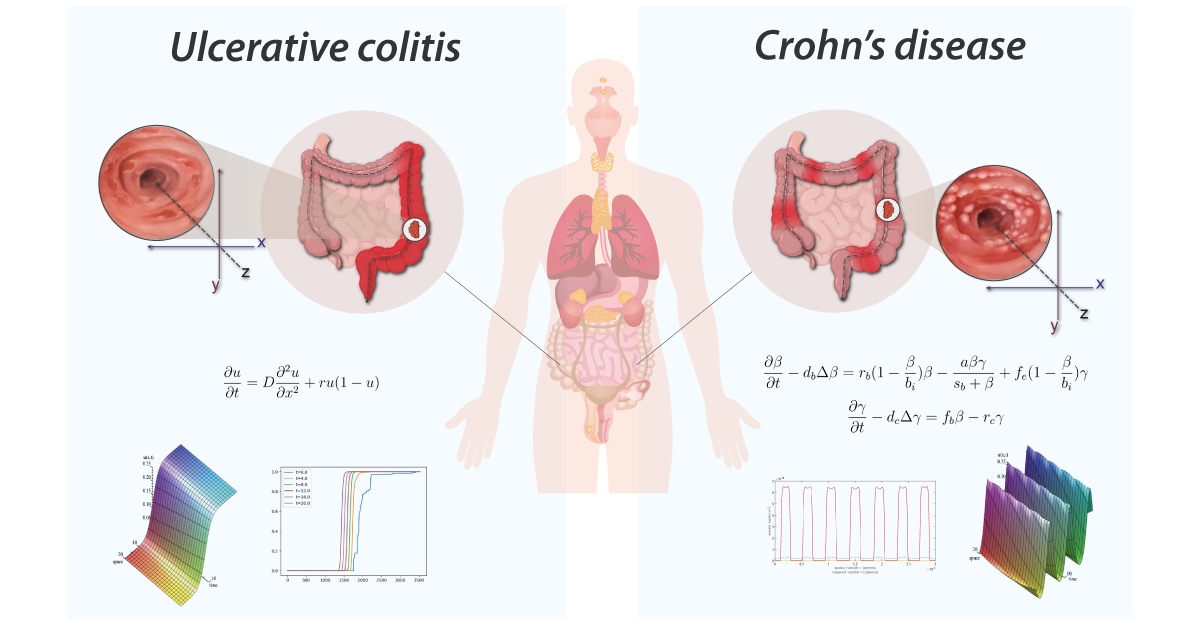}
    \caption{The figure shows how Crohn's disease differs from ulceratives colitis in terms of propagation, effect on the human body and mathematical modeling of both diseases. In the right, the system of partial differential equations is derived from this paper \citep{Nadin2021}. In the left, the Fisher kpp equation is used for the ulcerative colitis disease as shown in this study \citep{Toledo2021}.}
   \label{fig:bodymaths}
\end{figure}

Gastro-enterologists and surgeons are hindered from having spatial information on anatomical sites since these indicators are not spatial, and the provided information is never localized in a specific position. The diagnosis of these diseases is based on the analysis of colonoscopy videos. Thus, physicians assess the severity of the disease according to the presence of inflammation, bleeding or ulcers on the intestinal wall which requires an advanced level of expertise. In the same way, the extent of the lesions is currently ignored in medical practice, for lack of a validated method for analyzing this information. This same remark applies to other indicators like numerical score of severity \citep{Kraszewski2021}, the speed of inflammation propagation, the choice of treatment \citep{perforation_colonoscopy}. Gastroenterologists recognize the significance of spatial information in the development of complications such as esophageal and colon cancer in patients with Crohn's disease and ulcerative colitis. However, current guidelines fail to fully consider the quantity and distribution of lesions, often focusing solely on the most severe lesion identified. This is due to the scarcity of software tools and scientific literature. Additionally, the intricate feedback loops and technical challenges in collecting high-quality data for the calibration of numerical and mathematical models (see figure \ref{fig:bodymaths}) further highlights the need for innovative methods. This necessity is the driving force behind our current study, which aims to address the limitations in current approaches and provide a more comprehensive understanding of the disease.

Endoscopic video analysis \citep{AlAli2021} plays a crucial role in evaluating the severity of ulcerative colitis and monitoring the progression of the disease. Colonoscopy is widely used as the reference examination to assess the intensity of the disease and the extent of intestinal lesions. During this routine procedure, a gastroenterologist inserts a camera-equipped endoscope into the colon to visualize the inner lining and take biopsies if necessary. It should be noted that this technique has a very strong impact on the quality of life of the patients.

Wireless capsule endoscopy (WCE) \citep{AlAli2021} is another commonly used technique in which patients swallow a small, intelligent capsule that contains a camera and a light source. The capsule sends images of the intestinal mucosa to a wearable sensor, making it a less invasive alternative to colonoscopy. This method is especially useful for accessing regions of the small intestine that are difficult to reach with endoscopy. However, it is more expensive as the capsule can only be used once. Unfortunately, this technique has been abandoned in Tunisian hospitals due to its cost being considered expensive.

Both colonoscopy and WCE allow for the detection of important lesions in the videos, such as: Loss of visibility of the vascular framework, which is indicated by the disappearance of blood vessels and the formation of fibrous tissue that impedes nutrient absorption and inflammation and bleeding, which appear as red areas on the intestinal wall and ulcers and indentations in the wall that appear white or gray. The precise collection and examination of the endoscopic video data is essential not only for identifying the disease presence and advancement, but also for categorizing the different types of IBDs and classifying the subtypes within the same disease.

\subsection{PIML: A new and growing discipline with challenges}
Physics-Informed Machine Learning (PIML), also known as Physics-Informed Neural Networks (PINN), is an emerging discipline that merges the concepts of physics with the advanced techniques of machine learning and neural networks. The objective of PIML is to harness the laws of physics to increase the precision of machine learning models, particularly in cases where the systems being modeled are governed by partial differential equations. In PIML, the governing equations of a physical system are integrated into the training process of a machine learning model, resulting in predictions that are not only accurate, but also physically meaningful and interpretable. Overfitting is prevented through this approach. PIML has been successfully applied in diverse fields such as fluid dynamics, structural mechanics, quantum mechanics, cosmology and quantitative finance. With many advancements and applications yet to be discovered, PIML is a rapidly growing field that promises exciting new possibilities. According to the Gartner AI Hype Cycle diagrams for 2021 \citep{Gartner2021} and 2022 \citep{Gartner2022}, PINN and PIML are currently in the innovation trigger phase, gaining increasing attention and applications in the scientific community. With continued growth and development, these technologies are expected to reach the peak of inflated expectations before settling into a plateau of productivity. As PINN and PIML become more widely adopted, we can expect to see their use in solving real-world problems in biology and medicine (see the figure \ref{fig:gartner_hype_cycle1}).

The establishment of neural networks in mathematics can be traced back to the seminal work of Cybenko in 1989 \citep{Cybenko1989}. In this paper, Cybenko presented the concept of universal approximation, which demonstrated that a single hidden layer feedforward neural network with a sigmoid activation function is capable of approximating any continuous bounded function with a sufficient number of hidden units. This foundational work was further advanced by the studies of Hornik and Barron \citep{HORNIK1989, Barron1994}, which provided additional insights into the concept of universal approximation. Various architectures have been developed, starting with original PINN and followed by DeepFNet, DeepONet, DeepM\&MNet.

\begin{itemize}
    \item DeepFNet \citep{deepfnet} is a neural network architecture that is well-suited for functional approximation tasks because it is flexible, able to model complex relationships, and scalable. Its hierarchical structure allows it to learn and represent multi-scale features in the data, improving its ability to generalize and make accurate predictions on unseen data. Generally, it requires significantly fewer neurons than shallow networks to achieve a given degree of function approximation.
    \item DeepONet \citep{deeponet1}: uses a deep learning approach to learn nonlinear operators. The advantage is that it can capture complex relationships between variables that are not easily modeled using linear techniques. DeepONet uses seq2seq and fractional algorithms. Seq2seq (or "sequence-to-sequence") is a type of algorithm that is used to map input sequences to output sequences. Data with long-range relationships are analyzed using fractional (or "fractionally-differentiated") approaches.

\begin{figure}[H]
   \centering
    \includegraphics[scale=0.33]{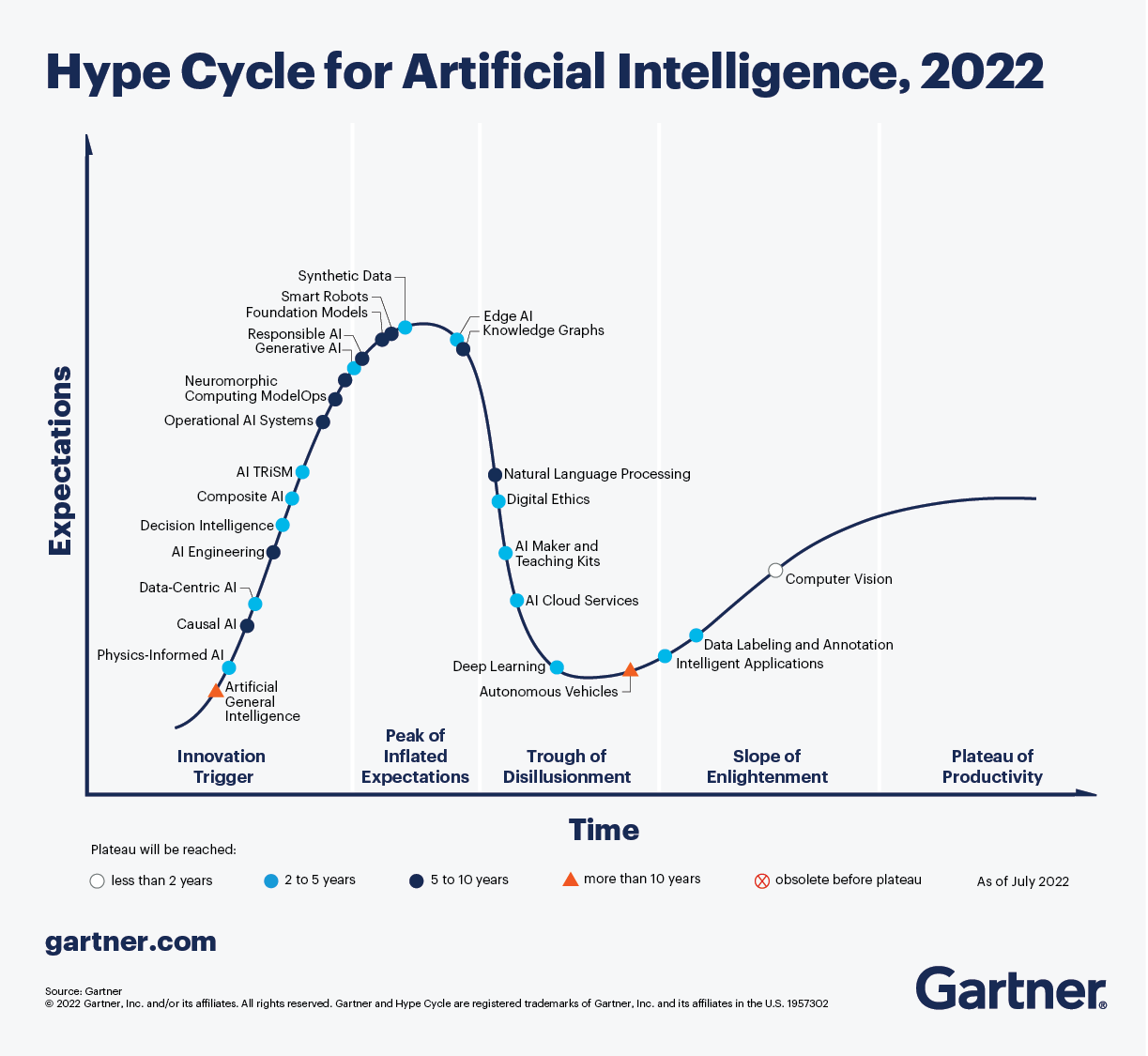}
    \caption{For the second consecutive year, the 2022 Gartner's hype cycle for artificial intelligence evokes the physics-informed AI. We notice that the PIML is actually in the innovation trigger regime with an improved outlook because the plateau of productivity will be reached between 2 to 5 years instead of 5 to 10 years.}
   \label{fig:gartner_hype_cycle1}
\end{figure}

    \item DeepM\&MNet \citep{deepmnet1, deepmnet2} is a neural network framework for simulating complex, multiphysics systems. It uses pre-trained neural networks to make predictions about the different fields in a coupled system, such as the flow, electric and concentration fields. The framework is designed to be fast and efficient, and can be used to build models with very little data. DeepM and MNet are versatile algorithms for modeling complex, multiphysics and multiscale dynamic systems. DeepM uses a multilayer perceptron architecture, while MNet uses a combination of convolutional and long short-term memory networks. Both algorithms are able to capture intricate patterns and trends in time series data.
    \end{itemize}

However, the best approach for using PINN in a particular biological system depends on the available data and knowledge of the physics of the system. We will explore three possible scenarios in which PINN can be applied in biology.
\begin{itemize}
\item First scenario: In this case, a Physics-Informed Neural Network (PINN) is used to make predictions about a system based on both data and known physics information. The neural network is trained on the data and also incorporates the known physics through the use of constraints or regularization terms in the loss function. This allows the network to make predictions that are consistent with the known physics and improve accuracy by utilizing the available data. Fluid dynamics represents a classic example where the neural network is trained on experimental or numerical data of the fluid flow and incorporates the governing equations of fluid dynamics as constraints or regularization terms in the loss function.

 \item Second scenario: In this scenario, there is a large amount of data available, but there is no physical model to describe the dynamics. For example, consider the physics of jets produced by terrestrial accelerators in heavy ion experiments such as ALICE at the LHC-CERN accelerator. The lack of a physical model can make it difficult to understand the underlying dynamics of the system. This is a good use case for traditional machine learning techniques, as there is no physics information to incorporate into the model.

 \item Third scenario: In this case, data is limited and the system is described by several physical models. The limited data and the presence of multiple physical models can make it challenging to determine which model is most appropriate for describing the system. In this case, it may be necessary to use a combination of approaches, such as combining physical models with machine learning techniques, to gain a more complete understanding of the system. It is also important to carefully validate the results and ensure that the chosen model is able to accurately describe the observed behavior. This is the case considered in this article, in which we attempt to model a biological phenomenon caused by loops of reactions and counter-reactions between bacteria and immune cells.
\end{itemize}

Having discussed the various techniques and scenarios involved in PINNs, it is now important to evaluate and quantify the performance of the model. From a general point of view, the neural network performance can be characterized into three main types: 
\begin{enumerate}
  \item Approximation error to ground truth function.
  \item Generalization to unseen data.
  \item Trainability of the model.
\end{enumerate}
In fact, the universal function approximation theorem only considers the approximation error of a neural network to the ground truth function. However, it does not consider other important factors, such as the generalization error and the model trainability. This generalization measures a model's ability to make accurate predictions on new, unseen data. Meanwhile, the model trainability is determined by factors such as its size, complexity, amount and quality of training data. These factors can influence the model's ability to be effectively trained, as larger and more complex models may require more resources and may be harder to converge. For that this theorm could not ensure the generalization and the trainability in complex biological process \citep{Lagergren-2020} or in the modeling of nonlinear two-phase transport in porous media \citep{Fuks_2020}. These limitations include the availability and quality of data and the potential for the models to fail to capture the full range of possible behaviors or phenomena. In the following points, we aim to shed light on the challenges faced in our modeling efforts.

\begin{itemize}
    \item Complexity of underlying processes: Biological processes are often characterized by complex interactions and dynamics, making it challenging to accurately model them using traditional mathematical or physical approaches. The non-linear nature of the differential equations involved, including the presence of non-linear terms such as square or cubic terms, only adds to the difficulty. These non-linear terms can even lead to blow-up phenomena, a common challenge for mathematicians working with PDEs \cite{Nadin2021}. This can also make it challenging for PINNs to learn and represent the underlying patterns and relationships in the data.
    
    \item Data availability and quality: The data used to train PINNs may be limited in quantity or quality, or may not be representative of the full range of behaviors or phenomena. This can affect the model's ability to learn and generalize, and can reduce its accuracy and reliability. Spatial data requires exhaustive examination, which can be expensive, and hospitals are often reluctant to share their data. Our attempts to contact digestive disease institutes for data resulted in a refusal to collaborate. However, we were able to find a more accessible data set called Kvasir that we plan to use for our study \cite{kvasir}.
    
    \item Limited range of behaviors : PINNs may not be able to capture the full range of behaviors and phenomena that can occur in biological systems. This is because the models are typically trained on a limited set of data and are not able to capture the full range of possible behaviors or situations that can arise.
    
    \item Necessity for simple and parsimonious PINN models: The complexity of the PINN model itself can also be a limitation. These models can be computationally expensive to train and may require a large amount of data and computational resources. This can make their use challenging in certain contexts, such as when data is limited or when computational resources are constrained, as in the case of automatic lesion detection techniques where the gastroenterologist manipulates the patient using the colonoscope and works on the software in real-time \citep{automaticdetection1, automaticdetection}.
    \end{itemize}

\subsection{Our approach}
The discussion above have provided the necessary foundation for our main work. We are attempting to predict the evolution of two bowel diseases with poor quality data collected only on the edges of the domain and by integrating physical constraints from nonlinear PDE with the simplest deep neural networks. We are inspired by the first work in PINN done by M. Raissi et al. \citep{RAISSI2019}. In this paper, the authors dealt with 4 equations:
\begin{itemize}
    \item Two-dimensional Navier-Stokes equation system
    \item Shrödinger equation
    \item Korteweg-De Vries equation
    \item Burgers' equation
\end{itemize}
Then, we will start with the simplest PDE equations and gradually add more complexity, including nonlinear terms. By following this progression, we hope to build a comprehensive model for predicting the evolution of these diseases. Therefore, we will apply this approch in these equations:
\begin{itemize}
    \item Simple partial differential equation (a Toy model)
    \item Diffusion equation or heat equation in 2D domain
    \item Fisher-KPP equation in 1D domain
    \item Korteweg-De Vries equations
    \item Traditional Turing System: nonlinear coupled system for Reaction Diffusion Equations
    \item Turing mechanism System for Crohn's disease presented in this paper \citep{Nadin2021}.
\end{itemize}

\section{Benchmarking our approach with some chosen PDEs}
This benchmarking refers to the process of evaluating the performance and accuracy of the DNN against a series of PDEs with a minimum of data. We begin by testing the DNN on a simple PDE and observe that it is able to accurately solve it with a high degree of accuracy. We will see in the next section that this resolution relates to the determination of the severity score for IBDs, incorporating the crucial aspect of spatial information distribution. Next, we apply the DNN to the Burger's equation, which is a more complex nonlinear PDE. The DNN is still able to solve this equation with a good level of accuracy. We then move on to the heat equation, which is a linear PDE and therefore relatively easier to solve. Finally, we test the DNN on a nonlinear system of Korteweg-De Vries equations.

\subsection{Neural networks architecture}
Choosing the hyperparameters of a DNN is an important step in its design and training. These parameters are the values that control the overall behaviour of the DNN, such as the number of layers, the number of neurons per layer, the learning rate and the regularization strength. Then, it is important to consider the nature of the problem being solved and the available data. For example, if the data is limited or noisy, it may be necessary to use a smaller or simpler network to avoid overfitting. The choice of optimization algorithm is another important factor in the training of the network. There are many different optimization algorithms available, each with its own strengths and weaknesses. Two commonly used optimization algorithms are L-BFGS and Adam.

\begin{figure}[H]
   \centering
    \includegraphics[scale=0.5]{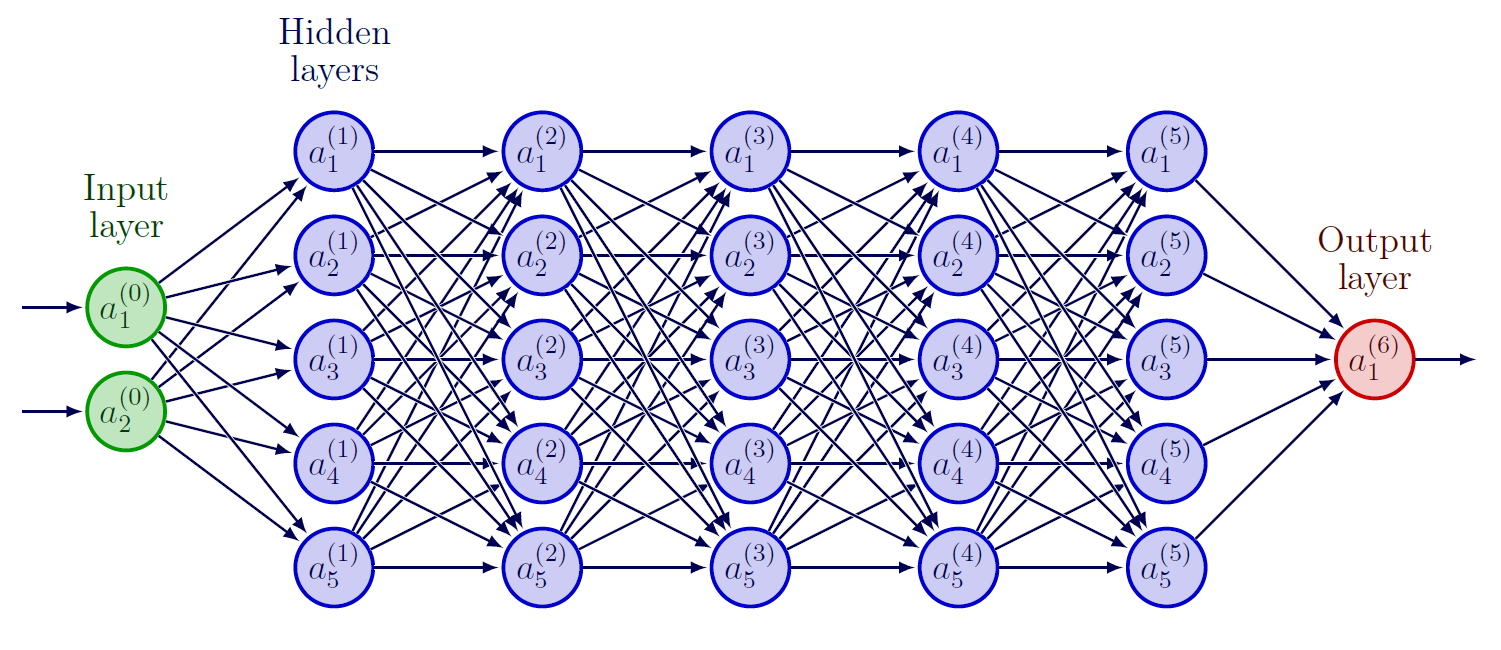}
\caption{Illustration of a DNN designed to approximate the concentration of phagocytes and bacteria in the digestive system. The network features an input layer of 2 neurons for capturing spatial and temporal information, followed by 5 hidden layers of 5 neurons each. The output layer consists of a single neuron that provides the final prediction. This architecture exhibits a fully connected structure.}
   \label{fig:gartner_hype_cycle}
\end{figure}

We can control overfitting using the dropout concept which works by randomly "dropping out" a fraction of the neurons during training, then a proportion of neurons are temporarily excluded from the network and do not contribute to the forward or backward passes. This has the effect of reducing the complexity of the DNN and forcing the remaining neurons to learn more robust and generalizable features. The notion of parsimony is important in deep learning because it helps to ensure that the models we build are as simple as possible while still being able to effectively capture the underlying patterns in the data. By using parsimony, we can avoid overfitting and build models that are more likely to perform well on new, unseen data.

\subsubsection{Adam vs LBFGS}
Stochastic gradient descent (SGD) is an optimization algorithm for finding model parameters that minimize the loss function, which measures the difference between the expected and actual output of a model \citep{Bottou1998}. There are many variations of SGD, including Adagrad, RMSprop, and Adam. Adam optimization is a stochastic gradient descent method that adaptively estimates first-order and second-order moments \citep{Kingma2015}.\\

In Adam optimization, the parameter update is given by:

$$m_w^{t+1}= \beta_1 m_w^{t}+(1-\beta_1)\nabla_w L^{t}$$
$$v_w^{t+1}= \beta_2 v_w^{t}+(1-\beta_2)(\nabla_w L^{t})^{2}$$
$$\hat{m}_w=\frac{m_w^{t+1}}{1-\beta_1^{t}}$$
$$\hat{v}_w=\frac{v_w^{t+1}}{1-\beta_2^{t}}$$
$$w^{t+1}=w^{t}- \eta\frac{\hat{m}_w}{\sqrt{\hat{v}_w}-\epsilon}$$

where $w(t)$ are the model parameters, $L^t$ is the loss function, $t$ is the current training iteration, $\beta_1$ and $\beta_2$ are the forgetting factors for gradients and second-order gradient moments, respectively.

On the other hand, the BFGS algorithm is a Quasi-Newton method for optimization, which approximates the Hessian matrix using a series of updates \citep{nocedal2006numerical, martens2011deep}. One of the most widely used Quasi-Newton methods is L-BFGS (Limited-memory BFGS \citep{Liu1989, Duchi2011}), which is more memory efficient than BFGS, as it only stores a few vectors representing the approximation of the Hessian matrix instead of the full matrix. L-BFGS is more computationally demanding than Adam, but can be faster in situations where the second-order moments are known analytically. Instead of storing the full n x n estimation of the inverse Hessian, L-BFGS stores only a few vectors that represent the approximation implicitly which make it more practical for usage in ML settings with small-mid sized datasets. This method is computationally more demanding. However when the second order is known analytically the optimization method moves faster towards the minimum. The reason lies in the fact that methods based on the first order approximate the error function at a point with a tangent hyperplane, while methods of the second order, with a quadratic hypersurface: this allows us to move closer to the error surface when we update the weights at each iteration.\\
The L-BFGS algorithm is as follows:

\begin{algorithm}
	\caption{L-BFGS} 
	\begin{algorithmic}
		\For {$i=1,2,\ldots,n$}
          \State Obtain a direction $P_k$ by solving $B_kP_k=-\nabla f(X_k)$\\
          \State Perform a one-dimensional optimization to find an acceptable stepsize $\alpha_k$ in the direction
          \State found in the first step. In an exact line search, $\alpha_k = \operatorname{argmin} f(X_k + \alpha P_k)$.
          \State In practice, an inexact line search usually suffices, with acceptable $\alpha_k$ satisfying the Wolfe
          \State conditions.\\
          \State  Set $S_k = \alpha_k P_k$ and update $X_{k+1}=X_k+S_k$\\
          \State $y_k=\nabla f(X_{k+1}) - \nabla f(X_k)$\\
          \State $B_{k+1}= B_k +\frac{Y_k Y_k^T}{s_k Y_k^T} +\frac{B_k s_k s_k^T B_k^T}{B_k s_k s_k^T}$
		\EndFor
	\end{algorithmic} 
\end{algorithm}

where $X_k$ are the model parameters, $f$ is the loss function, $B_k$ is an approximation of the Hessian matrix, and $S_k$ and $Y_k$ are the differences between the current and previous values of $X$ and $\nabla f(X)$, respectively. The L-BFGS and ADAM optimization algorithms are both commonly used in the context of training neural networks, including PINN. Here is a comparison of some key features of these algorithms:

\begin{itemize}
    \item Convergence rate: L-BFGS typically has a faster convergence rate than ADAM. However, ADAM can still be effective in practice and may be preferred in some cases due to its simplicity and ability to adapt to changing data \citep{Kingma2015}. 
    \item Memory requirements: L-BFGS requires storing a set of past gradients in memory, which can be costly for large datasets or networks. ADAM, on the other hand, only requires storing the average of the past gradients, which is typically much cheaper in terms of memory usage \citep{nocedal2006numerical}. 
    \item Sensitivity to hyperparameters: L-BFGS has relatively few hyperparameters, which can make it easier to tune. ADAM has more hyperparameters, such as the learning rate and the momentum parameters, which can be more sensitive to the choice of values and may require more careful tuning \citep{nocedal2006numerical, martens2011deep}.
    \item Robustness: L-BFGS can be sensitive to the initialization of the parameters, and may require a good initial guess to find the optimal solution. ADAM can be more robust to the initialization, but may be less sensitive to the true global minimum as shown in \citep{martens2011deep}.
\end{itemize}
Both L-BFGS and ADAM can be effective in training PINN models, and it may be useful to try both algorithms and compare their performance to determine which is the best fit for a particular problem. In our study, we found that the ADAM give best results.

\begin{table}[H]
\centering
\begin{tabular}{cc}
\multicolumn{2}{c}{\textbf{Optimization algorithms}} \\
\hline
\multicolumn{1}{c|}{\textbf{L-BFGS}} & \multicolumn{1}{c}{\textbf{ADAM}} \\
\hline
\multicolumn{1}{c|}{\textbf{$x_{k+1} = x_k + \alpha_k d_k$}} & \multicolumn{1}{c}{\textbf{$m_k = \beta_1 m_{k-1} + (1 - \beta_1) g_k$ }} \\
\multicolumn{1}{c|}{\textbf{$s_k = x_{k+1} - x_k$}} & \multicolumn{1}{c}{\textbf{$v_k = \beta_2 v_{k-1} + (1 - \beta_2) g_k^2$}} \\
\multicolumn{1}{c|}{\textbf{$y_k = g_{k+1} - g_k$}} & \multicolumn{1}{c}{\textbf{$m_k^\prime = \frac{m_k}{1 - \beta_1^k}$}} \\
\multicolumn{1}{c|}{\textbf{$\alpha_k = \frac{s_k^T y_k}{y_k^T H_k y_k}$}} & \multicolumn{1}{c}{\textbf{$v_k^\prime = \frac{v_k}{1 - \beta_2^k}$}} \\
\multicolumn{1}{c|}{\textbf{$x_{k+1} = x_k + \alpha_k d_k$}} & \multicolumn{1}{c}{\textbf{$x_{k+1} = x_k - \frac{\eta}{\sqrt{v_k^\prime} + \epsilon} m_k^\prime$}} \\
\hline

\end{tabular}
\caption{$x_k$ represents the current estimate of the solution, $d_k$ is the search direction, $\alpha_k$ is the step size, $s_k$ and $y_k$ are the difference vectors, $g_k$ is the gradient of the objective function at $x_k$, $m_k$ and $v_k$ are the first and second moments of the gradient estimates, $\beta_1$ and $\beta_2$ are the decay rates for the moving averages, $\eta$ is the learning rate, $\epsilon$ is a small constant to prevent division by zero, and $H_k$ is the approximate Hessian matrix. Note that the formulas for L-BFGS involve the computation of the Hessian matrix and its inverse, which can be computationally expensive for large datasets or networks. In contrast, the formulas for ADAM do not involve the Hessian matrix and can be more computationally efficient.}
\end{table}

\subsubsection{Root Mean Square Error(RMSE) Loss}
In the context of a biology process where the values are continuous and expected to be situated in a small range, using the mean squared error (MSE) or root mean squared error (RMSE) loss function can help ensure that the model is able to accurately predict the values within that range. Both MSE and RMSE measure the difference between the predicted values and the true values, but MSE is simply the average squared difference while RMSE is the square root of the average squared difference. Both of these loss functions penalize large errors more heavily than small errors, which can be useful for preventing the DNN model from making large errors in its predictions. 

$$ RMSE = \sqrt{\frac{ \sum_{i=1}^{n} (\widehat{x_{i}} - x_{i})^{2} } {n} } $$

With: $x_{i}$ = the $i^{th}$ observed value, $\widehat{x_i}$ = the $i^{th}$ predicted value and n is the number of available observations. 
 
\subsubsection{RELU vs Sigmoid vs Tanh}
In the context of tuning a DNN, the choice of activation function can significantly impact the performance. The Rectified Linear Unit (ReLU) activation function is widely used due to its simplicity and ability to converge faster than other activation functions. It is defined as $f(x) = max(0,x)$ and is generally used in the hidden layers of the network. One disadvantage of ReLU is that it can suffer from the "dying ReLU" problem, where the weights of the neurons become very small and the activation function becomes inactive. The Sigmoid activation function is defined as $f(x) = \frac{1}{1 + e^{-x}}$ and is often used in the output layer of binary classification problems. However, it has a slow convergence rate and can suffer from vanishing gradients, where the gradients of the weights become very small, hindering the model's ability to learn. The Hyperbolic Tangent (Tanh) activation function is defined as $f(x) = tanh(x) = 2\sigma(2x)-1$, where $\sigma(x)$ is the Sigmoid function. It is often used in the hidden layers of the network and can perform well in certain tasks, but it can also suffer from the vanishing gradients problem. The activation function choice can also depend on the range of the values being predicted, especially in a biology context. For example, if the predicted values are expected to be within a small range, such as between 0 and 1, the Sigmoid or Tanh activation functions may be more appropriate. However, if the predicted values are expected to have a larger range, the ReLU activation function may be more suitable. It is important to keep in mind that the choice of activation function is just one of many hyperparameters that can impact the performance of the model and should be tuned accordingly. In this work, we experimented with three activation functions: RELU, Sigmoid and Hyperbolic tangent. Our models converged for Sigmoid and Hyperbolic tangent but we couldn't obtain satisfying result with RELU.

\subsection{Toy model: simple partial differential equation}
Let's consider this first-order partial differential equation: 

$$ a * \frac{\partial u(x,t)}{\partial x} + b * \frac{\partial u(x,t)}{\partial t} + c * u(x,t) = 0$$

and let's fix the values of the constants a, b, and c to be a = 1, b = -2, and c = -1. The modified equation becomes:

$$ \frac{\partial u(x,t)}{\partial x} - 2 * \frac{\partial u(x,t)}{\partial t} - u(x,t) = 0$$

defined for the temporal interval $t \in [0,1]$ and the spatial interval $x \in [0,2]$. The initial condition is given by:

$$u(x,0) = 6 * e^{-3 * x}$$

and the PDE boundary conditions are given by:

$$u(x=0,t) = g_1(t)$$
$$u(x=2,t) = g_2(t)$$

(Note that during the numerical and PINN resolutions, we will choose $g_1(t) = 6 * e^{-2 * t} $ and $g_2(t) = 6 * e^{-6 -2 * t} $).

To solve this partial differential equation analytically, the method of separation of variables is used. Start by separating the variables x and t and this involves writing the solution in the form:
$$u(x,t) = X(x) * T(t)$$

Substituting this expression into the PDE gives:

$$  X'(x) * T(t) - 2 * X(x) * T'(t) - X(x) * T(t) = 0   $$

$$  ( X'(x) - X(x) ) * T(t) = 2 * X(x) * T'(t)   $$

$$  \frac{X'(x) - X(x)}{X(x)} = 2 * \frac{T'(t)}{T(t)} = C_0 $$

with $C_0$ is an arbitrary constant.

Solving for X(x) and T(t) separately gives:

Let's solve the left member, $$\frac{X'(x) - X(x)}{X(x)} = C_0$$

The integration gives:

$$X(x) = C_1 * e^{( C_0 + 1) * x }$$

The right member is written $$\frac{T'(t)}{T(t)} = \frac{C_0}{2}$$

The integration gives: 

$$T(t) = C_2 * e^{ \frac{C_0 * t}{2} }$$

The general solution is then given by:

$$u(x,t)=X(x) * T(t) =  C_1 * C_2 * e^{( C_0 + 1) * x + \frac{C_0 * t}{2} }$$

Using the initial condition at $t=0$:
$$u(x,0)=C_1 * C_2 * e^{(C_0 + 1 ) * x} = 6 * e^{-3 * x}$$

Then $C_1 * C_2 = 6 $ and $C_0 = -4$

 Therefore, the solution is:

$$u(x,t) = 6 * e^{-3 * x - 2 * t}$$

\subsection*{Analytical solution}

$$ u(x,t) = 6 * e^{-3 * x - 2 * t}, \hspace{0.5cm} x \in [0,2], \hspace{0.5cm} t \in [0,1]$$

\subsection*{Numerical schema}
Now, we are interested in numerically solving the previous PDE by using the centered finite difference method. We need to discretize the spatial and temporal domains and approximate the derivatives using finite differences. Here is an outline of the steps involved in the numerical scheme:

\begin{itemize}
    \item Choose a spatial discretization step size $\Delta x$, and a temporal discretization step size $\Delta t$.
    \item Define the grid points $x_i$ and $t_j$ as follows:

\end{itemize}

$$x_i = i \Delta x \hspace{1cm} t_j = j \Delta t$$

$$\frac{\partial u}{\partial x} \approx \frac{u_{i+1,j} - u_{i-1,j}}{2\Delta x}$$

$$\frac{\partial u}{\partial t} \approx \frac{u_{i,j+1} - u_{i,j-1}}{2\Delta t}$$

$$ u_{i,j} = \frac{u_{i+1,j} - u_{i-1,j}}{2\Delta x} - 2 \frac{u_{i,j+1} - u_{i,j-1}}{2\Delta t} - u_{i,j} $$

$$ u_{i,j} = \frac{1}{1 + 2 \frac{\Delta t}{\Delta x}} (u_{i+1,j} - 2u_{i,j} + u_{i-1,j} - 2\frac{\Delta t}{\Delta x} (u_{i,j+1} - u_{i,j-1})) $$

Substitute the initial condition $u(x_i,0) = 6 * \exp(-3 * x_i)$ to find $u_{i,0}$.\\

Substitute the boundary conditions $u(0,t_j) = 6 * e^{-2 * t_j}$ and $u(2,t_j) = 6 * e^{-6 -2 * t_j}$ to find $u_{0,j}$ and $u_{N,j}$, where $N$ is the number of grid points in the spatial domain.\\

\begin{lstlisting}[language=Python, caption={The code used to compute the numerical solution and the analytical one. The results are visualized using the matplotlib library.}]
import numpy as np
import matplotlib.pyplot as plt


def solve_pde(a, b, c, Delta_x, Delta_t):
    # Number of grid points in spatial and temporal domains
    N = int(2 / Delta_x)
    M = int(1 / Delta_t)

    # Define grid points
    x = np.linspace(0, 2, N+1)
    t = np.linspace(0, 1, M+1)

    # Initialize solution array
    u = np.zeros((N+1, M+1))

    # Set initial condition
    u[:, 0] = 6 * np.exp(-3 * x)

    # Set boundary conditions
    u[0, :] = 6 * np.exp(-2 * t)
    u[N, :] = 6 * np.exp(-6 - 2 * t)

    # Iterate over temporal domain
    for j in range(M):
        # Iterate over spatial domain
        for i in range(1, N):
            # Use finite difference equation to update solution
            u[i, j+1] = (1 / (1 + 2 * Delta_t / Delta_x)) * (u[i+1, j] - 2 * u[i, j] + u[i-1, j] - 2 * Delta_t / Delta_x * (u[i, j+1] - u[i, j-1]))

    # Return solution
    return u

# Define function to calculate analytical solution
def analytical_solution(x, t):
    return 6 * np.exp(-2 * t - 3 * x)

# Set discretization step sizes
Delta_x = 0.1
Delta_t = 0.01

# Calculate numerical solution
solution = solve_pde(2, 1, -1, Delta_x, Delta_t)

# Define grid points for plotting
x_plot = np.linspace(0, 2, int(2 / Delta_x) + 1)
t_plot = np.linspace(0, 1, int(1 / Delta_t) + 1)
X, T = np.meshgrid(x_plot, t_plot)

# Calculate analytical solution for plotting
analytical = analytical_solution(X, T)

# Plot numerical and analytical solutions
plt.figure()
plt.title("Numerical and analytical solutions")
plt.pcolor(X, T, solution, cmap="coolwarm")
plt.colorbar()
plt.contour(X, T, analytical, colors="k")
plt.xlabel("x")
plt.ylabel("t")
plt.show()

\end{lstlisting}

\subsection*{PINN approach}
In the case of numerical methods, the approach is to convert the PDE into numerical schemes, ensuring properties such as numerical convergence, consistency, and numerical stability. In the PINN approach, the problem is reformulated as a numerical optimization problem the goal is to end up with a loss function that contains most of the system's dynamic information. To achieve this, the equation is first rearranged such that all its terms are gathered on one side, which forms the first term of the cost function. The remaining two terms of the loss function consist of the initial and boundary conditions, respectively. Then, the function is the sum of three positive terms that must be minimized through an optimization algorithm. Secondly, the data is generated randomly sampled from the phase space, with points located at the boundary, within the interior of the domain, and subject to the initial conditions. 

Let us define $f(x, t)$  as :

$$ f(x,t) = \frac{\partial u(x,t)}{\partial x} - 2 * \frac{\partial u(x,t)}{\partial t} - u(x,t) \hspace{1cm} x \in [0,2], t \in [0,1] $$

In this case, $u(x, t)$ will be approximated by a neural network. The latter has as an objective to minimize the following loss :

$$ MSE = MSE_0 + MSE_b + MSE_f $$

where 

$$ MSE_0 = \frac{1}{N_0} \sum_{i=1}^{N_0} |u(x_{0}^{i}, 0) - u_{0}^{i}|^{2} $$

$$ MSE_b = \frac{1}{N_b} \sum_{i=1}^{N_b} (|u(0, t_{b}^{i}) - u_{b}^{i}|^{2} + |u(2, t_{b}^{i}) - u_{b}^{i}|^{2}) $$

$$ MSE_f = \frac{1}{N_f} \sum_{i=1}^{N_f} |f(x_{f}^{i}, t_{f}^{i})|^{2} $$

$ \{x_{0}^{i}, u_{0}^{i}\} $ denotes the initial data at $t=0$, $ \{t_{b}^{i}, u_{b}^{i}\} $ denotes to the boundary data and $ \{ x_{f}^{i}, t_{f}^{i}\} $ corresponds to collocation points on $f(x, t)$.

\begin{lstlisting}[language=Python, caption= An implementation of PINN for the toy model.]

import numpy as np
import torch
import torch.nn as nn
from torch.autograd import Variable

device = torch.device('cuda:0' if torch.cuda.is_available() else 'cpu')

class Net(nn.Module):
  def __init__(self):
    super(Net, self).__init__()
    self.hidden_layer1 = nn.Linear(2,5)
    self.hidden_layer2 = nn.Linear(5,5)
    self.hidden_layer3 = nn.Linear(5,5)
    self.hidden_layer4 = nn.Linear(5,5)
    self.hidden_layer5 = nn.Linear(5,5)
    self.output_layer = nn.Linear(5,1)

  def forward(self, x, t):
    inputs = torch.cat([x,t], axis=1)
    layer1_out = torch.sigmoid(self.hidden_layer1(inputs))
    layer2_out = torch.sigmoid(self.hidden_layer2(layer1_out))
    layer3_out = torch.sigmoid(self.hidden_layer3(layer2_out))
    layer4_out = torch.sigmoid(self.hidden_layer4(layer3_out))
    layer5_out = torch.sigmoid(self.hidden_layer5(layer4_out))
    output = self.output_layer(layer5_out)

    return output

def f(x, t, net):
  u = net(x, t)
  u_x = torch.autograd.grad(u.sum(), x, create_graph=True)[0]
  u_t = torch.autograd.grad(u.sum(), t, create_graph=True)[0]

  pde = u_x - 2*u_t - u 
  return pde

def get_boundary_data():
  x_bc = np.random.uniform(low=0.0, high=2.0, size=(500,1))
  t_bc = np.zeros((500,1))

  u_bc = 6*np.exp(-3*x_bc)
  return x_bc, t_bc, u_bc

def get_collocation_data():
  x_collocation = np.random.uniform(low=0.0, high=2.0, size=(500,1))
  t_collocation = np.random.uniform(low=0.0, high=1.0, size=(500,1))
  all_zeros = np.zeros((500,1))
  return x_collocation, t_collocation, all_zeros

def train_model(net, cost_function, optimizer, num_iterations):
  previous_validation_loss = 99999999.0
  
  for epoch in range(num_iterations):
    optimizer.zero_grad()

    # Loss based on boundaries conditions
    x_bc, t_bc, u_bc = get_boundary_data()
    pt_x_bc = Variable(torch.from_numpy(x_bc).float(), requires_grad=False).to(device)
    pt_t_bc = Variable(torch.from_numpy(t_bc).float(), requires_grad=False).to(device)
    pt_u_bc = Variable(torch.from_numpy(u_bc).float(), requires_grad=False).to(device)

    net_bc_out = net(pt_x_bc, pt_t_bc)
    mse_u = cost_function(net_bc_out, pt_u_bc)

    # Loss based on PDE
    x_collocation, t_collocation, all_zeros = get_collocation_data()
    pt_x_collocation = Variable(torch.from_numpy(x_collocation).float(), requires_grad=True).to(device)
    pt_t_collocation = Variable(torch.from_numpy(t_collocation).float(), requires_grad=True).to(device)
    pt_all_zeros = Variable(torch.from_numpy(all_zeros).float(), requires_grad=False).to(device)

    f_out = f(pt_x_collocation, pt_t_collocation, net)
    mse_pde = cost_function(f_out, pt_all_zeros)

    # Total loss
    total_loss = mse_u + mse_pde
    total_loss.backward()
    optimizer.step()

    # Validation
    if epoch % 1000 == 0:
      x_validation = np.random.uniform(low=0.0, high=2.0, size=(500,1))
      t_validation = np.random.uniform(low=0.0, high=1.0, size=(500,1))
      u_validation = 6*np.exp(-3*x_validation)

      pt_x_validation = Variable(torch.from_numpy(x_validation).float(), requires_grad=False).to(device)
      pt_t_validation = Variable(torch.from_numpy(t_validation).float(), requires_grad=False).to(device)
      pt_u_validation = Variable(torch.from_numpy(u_validation).float(), requires_grad=False).to(device)

      net_validation_out = net(pt_x_validation, pt_t_validation)
      mse_validation_loss = cost_function(net_validation_out, pt_u_validation)

\end{lstlisting}

\subsection*{Comparison between PINN, analytical solutions and finite difference method}

\begin{figure}[htp]
\centering
\includegraphics[width=.3\textwidth]{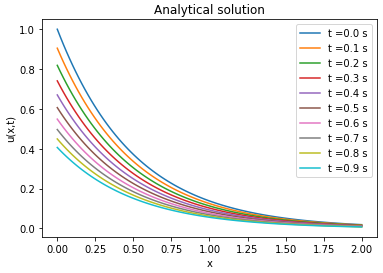}\hfill
\includegraphics[width=.3\textwidth]{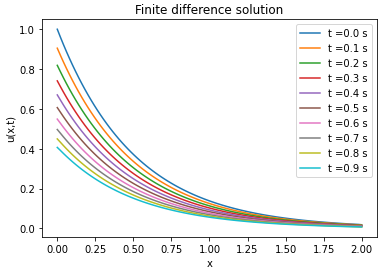}\hfill
\includegraphics[width=.3\textwidth]{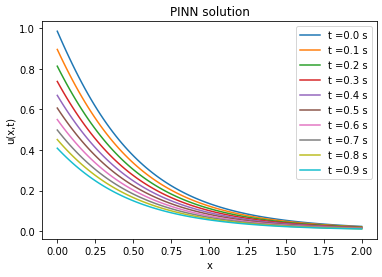}
\caption{PINN, analytical and finite difference solutions side by side.}
\label{fig:figure3}
\end{figure}

\begin{figure}[htp]

\centering
\includegraphics[width=.3\textwidth]{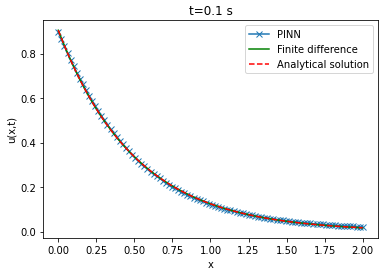}\hfill
\includegraphics[width=.3\textwidth]{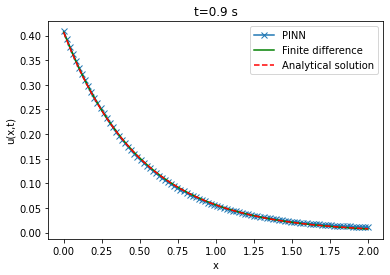}\hfill
\includegraphics[width=.3\textwidth]{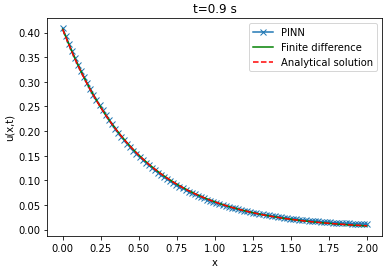}

\caption{PINN, analytical and finite difference at t=0.1 s, t=0.5 s and t = 0.9 s.}
\label{fig:figure3}

\end{figure}

\begin{table}[H]
\centering
\begin{tabular}{ll|lll|lll|}
\cline{3-8}
                                              &   & \multicolumn{3}{l|}{Analytical solution}                                 & \multicolumn{3}{l|}{Finite difference solution}                          \\ \cline{3-8} 
                                              &   & \multicolumn{1}{l|}{8}        & \multicolumn{1}{l|}{16}       & 32       & \multicolumn{1}{l|}{8}        & \multicolumn{1}{l|}{16}       & 32       \\ \hline
\multicolumn{1}{|l|}{Layers} & 2 & \multicolumn{1}{l|}{0.000472} & \multicolumn{1}{l|}{0.000416} & 0.000746 & \multicolumn{1}{l|}{0.000702} & \multicolumn{1}{l|}{0.000332} & 0.000572 \\ \cline{2-8} 
\multicolumn{1}{|l|}{}                        & 4 & \multicolumn{1}{l|}{0.000366} & \multicolumn{1}{l|}{0.000465} & 0.000412 & \multicolumn{1}{l|}{0.000614} & \multicolumn{1}{l|}{0.000710} & 0.000644 \\ \cline{2-8} 
\multicolumn{1}{|l|}{}                        & 8 & \multicolumn{1}{l|}{0.000790} & \multicolumn{1}{l|}{0.001780} & 0.001695 & \multicolumn{1}{l|}{0.000979} & \multicolumn{1}{l|}{0.001777} & 0.001548 \\ \hline
\end{tabular}
\caption{\label{tab:RSME} RMSE between PINN and analytical solution / and PINN and finite difference solution for the toy model.}
\end{table}

\begin{table}[H]
\centering
\begin{tabular}{ll|lll|}
\cline{3-5}
                                              &   & \multicolumn{3}{l|}{Neurons}                                 \\ \cline{3-5} 
                                              &   & \multicolumn{1}{l|}{8}    & \multicolumn{1}{l|}{16}   & 32   \\ \hline
\multicolumn{1}{|l|}{Layers} & 2 & \multicolumn{1}{l|}{2:05} & \multicolumn{1}{l|}{2:02} & 2:55 \\ \cline{2-5} 
\multicolumn{1}{|l|}{}                        & 4 & \multicolumn{1}{l|}{3:58} & \multicolumn{1}{l|}{3:46} & 3:43 \\ \cline{2-5} 
\multicolumn{1}{|l|}{}                        & 8 & \multicolumn{1}{l|}{4:44} & \multicolumn{1}{l|}{5:16} & 5:26 \\ \hline
\end{tabular}
\caption{\label{tab:Time }Execution time according to the number of layers and neurons for the toy model.}
\end{table}

\subsection{Burger's equation}
The Burger's equation is a nonlinear PDE that is often used to model a variety of physical, biological, and chemical phenomena, including incompressible fluid flow, population dynamics, and chemical reactions. It expresses the balance between the fluid convective transport and the diffusive transport due to viscosity. Solving the Burger's equation allows to determine the fluid velocity field at a given time and spatial location. This equation has several applications in biology like modeling blood flow in the cardiovascular system \citep{bloodmodelingburgers}, modeling pattern formation biological systems and modeling population dynamics \citep{populationburgers}. Two types of Burgers equations are considered, those without viscosity term which can be obtained by considering the particles non-interaction and whose solution can be realized with the help of the finite method, including the derivatives approximation means of Taylor developments based on the discretization of the phase space with viscosity term. The solution of the nonlinear viscosity problem based on the Cole-Hopf transform. Concerning numerical resolution done here either by a finite difference method. The first limitation of the Burgers equation is that it is a simplified model that makes certain assumptions about the system being studied. For example, it may assume that the fluid is incompressible or that the reaction rate is constant, which may not hold true in all cases. As a result, the Burgers equation may not be able to accurately capture the full complexity of a heterogeneous system. The second limitation is that the equation is a deterministic model, not take into account the inherent randomness that is often present in biological systems.

\subsubsection*{Equation}
The Burger's equation is the following:

$$ \frac{\partial u(x,t)}{\partial t} + u(x,t) \frac{\partial u(x,t)}{\partial x} - \nu \frac{\partial^2 u(x,t)}{\partial x^{2}} = 0   \hspace{1.5cm} x \in [0,1], t \in [0,0.1] $$

with the initial condition:

 $u(x,0)=\sin(\pi x)$
 
and the boundary conditions:

 $u(0, t) = u(1, t) = 0 $
 
where $\nu$ = 1 is the coefficient of kinematic viscosity
 
Using the Hopf-Cole transformation

$$u(x,t) = -2\nu \frac{\Theta_{x}}{\Theta}$$

The burger's equation transforms to the linear heat equation:

$$ \frac{\partial \Theta(x,t)}{\partial t} = \nu \frac{\partial^2 \Theta(x,t)}{\partial x^{2}}  \hspace{1.5cm} x \in [0,1], t \in [0,0.1] $$

with the initial condition

$\Theta(x,0) = \exp(\frac{\cos(\pi x) - 1}{2\pi \nu})$

and the boundary conditions

$\Theta_{x}(0,t) = \Theta_{x}(1,t)=0$

\subsection*{Analytical solution}
The Fourier solution to the burger's equation is given by

$$u(x,t) = 2\pi \nu \frac{\sum_{n=1}^{\infty} a_{n}\exp(-n^{2}\pi^{2}\nu t) n \sin(n \pi x)}{a_{0} + \sum_{n=1}^{\infty} a_{n}\exp(-n^{2}\pi^{2}\nu t) \cos(n \pi x)}$$

with 

$a_{0} = \int_{0}^{1} \exp(\frac{\cos(\pi x) - 1}{2\pi \nu}) \,dx$

$a_{n} = \int_{0}^{1} \exp(\frac{\cos(\pi x) - 1}{2\pi \nu})\cos(n \pi x) \,dx \hspace{0.5cm} n>0$

\begin{lstlisting}[language=Python, caption= The code calculates the various coefficients and plots the analytic solution at different temporal instants.]

import numpy as np
from scipy.integrate import quad
import matplotlib.pyplot as plt
from tqdm import tqdm

def A_0(x, v_d):
  return np.exp((np.cos(np.pi*x) - 1)/(2*np.pi*v_d))

def A_n(x, v_d, n):
  return 2*np.exp((np.cos(np.pi*x) - 1)/(2*np.pi*v_d))*np.cos(n*np.pi*x)

def fct_u(x,t,v_d):
  v = v_d
  Num = 0
  Den = 0
  for n in range(1,101):
    a = quad(A_n, 0, 1, args=(v,n))[0]*np.exp(-v*t*(n**2)*(np.pi**2))
    Num += a*n*np.sin(n*np.pi*x)
    Den += a*np.cos(n*np.pi*x)

  return (2*np.pi*v*Num)/(quad(A_0, 0, 1, args=(v))[0] + Den)

for i in range(0,10):
  plt.plot(fct_u(np.arange(0,1,0.001), i/100,1), label='i = ' + str(i/100))
  plt.legend()

\end{lstlisting}

\subsection*{Numerical schema}
The solution domain {(x, t) :$x \in [0, 1]$, $t \in [0, 0.1]$} is discretized into cells described by the node set $(x_{i}, t_{j})$ in which $x_{j}$ =ih, $t_{j}$ =jk (i = 0,1,...,N; j = 0,1,...,J, Nh= 1 and Jk = 0.1) h=$\Delta x$  is a spatial mesh size, k=$\Delta t$ is the time step.

A standard explicit finite difference approximation of the heat equation is

$\Theta_{i,j+1} = (1-2r)\Theta_{i,j} + 2r\Theta_{i+1,j}, \hspace{0.5cm} i=0$

$\Theta_{i,j+1} = r\Theta_{i-1,j} + (1-2r)\Theta_{i,j} + r\Theta_{i+1,j}, \hspace{0.5cm} i=1,2,...,N-1$

$\Theta_{i,j+1} = 2r\Theta_{i-1,j} + (1-2r)\Theta_{i,j}, \hspace{0.5cm} i=N$

Using the Hopf-Cole transformation

$$u(x_{i},t_{j}) = - \frac{\nu}{h}(\frac{\Theta_{i+1,j} - \Theta_{i-1,j}}{\Theta_{i,j}}), \hspace{0.5cm} i=1,...,N-1, \hspace{0.2cm} j=0,...,J$$

\begin{lstlisting}[language=Python, caption= An implementation of the finite difference method for solving partial differential equations.]

v = 1
r = 1/4
t_f = 0.1
delta_x = 0.01
delta_t = (r*delta_x**2)/v
N =  int(1/delta_x)
J = int(t_f/delta_t)

print('delta_x =', delta_x,'delta_t =',delta_t,'N =',N,'J =',J)

x = np.arange(0, 1, delta_x)
t = np.arange(0, t_f, delta_t)

theta_df = np.zeros((N,J))

#Initial condition
theta_df[:,0] = A_0(x, v)

for j in tqdm(range(0,J-1)):
  theta_df[0,j+1] = (1 - 2*r)*theta_df[0,j] + 2*r*theta_df[1,j]
  for i in range(1,N-1):
    theta_df[i,j+1] = r*theta_df[i-1,j] + (1 - 2*r)*theta_df[i,j] + r*theta_df[i+1,j]

  theta_df[N-1,j+1] = 2*r*theta_df[N-2,j] + (1 - 2*r)*theta_df[N-1,j]

u_df = np.zeros((N-2, J))

for j in tqdm(range(0,J)):
  for i in range(1,N-2):
    u_df[i,j] = -(v/delta_x)*(theta_df[i+1,j] - theta_df[i-1,j])/theta_df[i,j]

print(u_df.shape)

for i in range(0,J,500):
  plt.plot(u_df[:,i], label = 'i =' +str(i*delta_t))
  plt.legend()

for i in range(0,J,500):
  plt.plot(fct_u(x, i*delta_t,1), label = 'i =' +str(i*delta_t))
  plt.legend()

\end{lstlisting}

\subsection*{PINN approach}

Let us define $f(x, t)$  as :

$$ f = \frac{\partial u(x,t)}{\partial t} + u(x,t) \frac{\partial u(x,t)}{\partial x} - \nu \frac{\partial^2 u(x,t)}{\partial x^{2}}   \hspace{1.5cm} x \in [0,1], t \in [0,0.1] $$

In this case, $u(x, t)$ will be approximated by a neural network. The latter has as an objective to minimize the following loss :

$$ MSE = MSE_0 + MSE_b + MSE_f $$

where 

$$ MSE_0 = \frac{1}{N_0} \sum_{i=1}^{N_0} |u(x_{0}^{i}, 0) - u_{0}^{i}|^{2} $$

$$ MSE_b = \frac{1}{N_b} \sum_{i=1}^{N_b} (|u(0, t_{b}^{i}) - u_{b}^{i}|^{2} + |u(1, t_{b}^{i}) - u_{b}^{i}|^{2}) $$

$$ MSE_f = \frac{1}{N_f} \sum_{i=1}^{N_f} |f(x_{f}^{i}, t_{f}^{i})|^{2} $$

$ \{x_{0}^{i}, u_{0}^{i}\} $ denotes the initial data at $t=0$, $ \{t_{b}^{i}, u_{b}^{i}\} $ denotes to the boundary data and $ \{ x_{f}^{i}, t_{f}^{i}\} $ corresponds to collocation points on $f(x, t)$.

\begin{lstlisting}[language=Python, caption= An implementation of the PINN for Burger’s equation.]

import torch
import torch.nn as nn
import numpy as np
from random import uniform
device = torch.device('cuda:0' if torch.cuda.is_available() else 'cpu')

torch.set_default_tensor_type('torch.cuda.FloatTensor')

N_u = 4000
N_f = 10000

#Make X_uv_train
#BC x=-1  & tt t > 0
x_left = np.ones((N_u//4,1), dtype=float)*(0)
t_left = np.random.uniform(low=10**(-6), high=0.1, size=(N_u//4,1))
X_left = np.hstack((x_left, t_left))

#BC x=1 & tt t > 0
x_right = np.ones((N_u//4,1), dtype=float)*(1)
t_right = np.random.uniform(low=10**(-6), high=0.1, size=(N_u//4,1))
X_right = np.hstack((x_right, t_right))

#IC t=0 & tt x,y in [-1,1]
t_zero = np.zeros((N_u//2,1), dtype=float)
x_zero = np.random.uniform(low=0.0, high=1.0, size=(N_u//2,1))
X_zero = np.hstack((x_zero, t_zero))

X_u_train = np.vstack((X_left, X_right, X_zero))
# shuffling
index=np.arange(0,N_u)
np.random.shuffle(index)
X_u_train=X_u_train[index,:]

#Make u_train
u_left = np.zeros((N_u//4,1), dtype=float)
u_right = np.zeros((N_u//4,1), dtype=float)
u_initial = np.sin(np.pi * x_zero) 
u_train = np.vstack((u_left, u_right, u_initial))

# ==========================================
u_train=u_train[index,:]
# ==========================================
# make X_f_train 
X_f_train=np.zeros((N_f,2),dtype=float)
for row in range(N_f):
    x=uniform(0,1)
    t=uniform(10**(-6),0.1)  
    X_f_train[row,0]=x
    X_f_train[row,1]=t
   
X_f_train=np.vstack((X_f_train, X_u_train))

class PhysicsInformedNN():
  def __init__(self,X_u,u,X_f):
    # x & t from boundary conditions
    self.x_u = torch.tensor(X_u[:, 0].reshape(-1, 1),dtype=torch.float32,requires_grad=True)
    self.t_u = torch.tensor(X_u[:, 1].reshape(-1, 1),dtype=torch.float32,requires_grad=True)
    # x & t from collocation points
    self.x_f = torch.tensor(X_f[:, 0].reshape(-1, 1),dtype=torch.float32,requires_grad=True)
    self.t_f = torch.tensor(X_f[:, 1].reshape(-1, 1),dtype=torch.float32,requires_grad=True)
    # boundary solution
    self.u = torch.tensor(u, dtype=torch.float32)
    # null vector to test against f:
    self.null =  torch.zeros((self.x_f.shape[0], 1))
    # initialize net:
    self.create_net()

    self.optimizer = torch.optim.Adam(self.net.parameters(),lr=0.001)
    # typical MSE loss (this is a function):
    self.loss = nn.MSELoss()
    # loss :
    self.ls = 0
    # iteration number:
    self.iter = 0

  def create_net(self):
    self.net=nn.Sequential(
        nn.Linear(2,32), nn.Sigmoid(),
        nn.Linear(32, 32), nn.Sigmoid(),
        nn.Linear(32, 32), nn.Sigmoid(),
        nn.Linear(32, 32), nn.Sigmoid(),
        nn.Linear(32, 1))
  def net_u(self,x,t):
    u=self.net(torch.hstack((x,t)))
    return u
  def net_f(self,x,t):

    vega = 1
    u=self.net_u(x,t)
    u = u.to(device)

    u_t=torch.autograd.grad(u,t,grad_outputs=torch.ones_like(u),create_graph=True)[0]
    u_x=torch.autograd.grad(u,x,grad_outputs=torch.ones_like(u),create_graph=True)[0]
    u_xx=torch.autograd.grad(u_x,x,grad_outputs=torch.ones_like(u),create_graph=True)[0]
    u_t = u_t.to(device)
    u_x = u_x.to(device)
    u_xx = u_xx.to(device)
    

    f = u_t + u*u_x - vega*u_xx
    f = f.to(device)

    return f

  def closure(self):
    # reset gradients to zero 
    self.optimizer.zero_grad()
    # u & f predictions
    u_prediction = self.net_u(self.x_u, self.t_u)
    #
    f_prediction = self.net_f(self.x_f,self.t_f)
    #
    # losses:
    u_loss_x = self.loss(u_prediction, self.u)
    f_loss = self.loss(f_prediction, self.null)
    self.ls = u_loss_x + f_loss
    # derivative with respect to net's weights:
    self.ls.backward()
    # increase iteration count:
    self.iter += 1
    # print report:
    if not self.iter % 100:
      print('Epoch: {0:}, Loss: {1:6.8f}'.format(self.iter, self.ls))
    return self.ls    
    
  def train(self):
    """
    training loop
    """
    for epoch in range(15000):
      self.net.train()
      self.optimizer.step(self.closure)

# pass data sets to the PINN:
pinn = PhysicsInformedNN(X_u_train, u_train, X_f_train)
pinn.train()

import matplotlib.pyplot as plt
from mpl_toolkits.axes_grid1 import make_axes_locatable
#     
x = torch.linspace(0, 1, 100)
t = torch.linspace( 0, 0.1, 4000)
# x & t grids:
X,T = torch.meshgrid(x,t)
# x & t columns:
xcol = X.reshape(-1, 1)
tcol = T.reshape(-1, 1)
# one large column:
usol = pinn.net_u(xcol,tcol)
# reshape solution:
U = usol.reshape(x.numel(),t.numel())
# transform to numpy:
xnp = x.cpu().numpy()
tnp = t.cpu().numpy()
Unp = U.cpu().detach().numpy()

print(Unp.shape)

for i in range(0,4000,500):
  plt.plot(Unp[1:99,i], label = 'i =' +str(i*2.5*10**(-5)) )
  plt.legend()

\end{lstlisting}

\subsection*{Comparison between PINN, analytical solution and finite difference method}
\begin{figure}[htp]

\centering
\includegraphics[width=.5\textwidth]{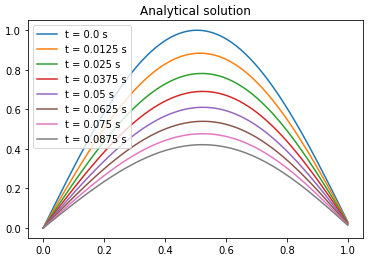}\hfill
\includegraphics[width=.5\textwidth]{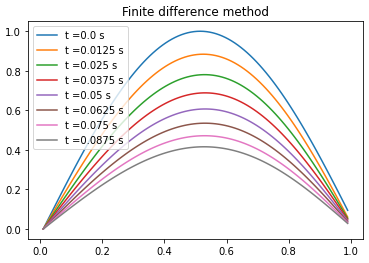}\hfill
\includegraphics[width=.5\textwidth]{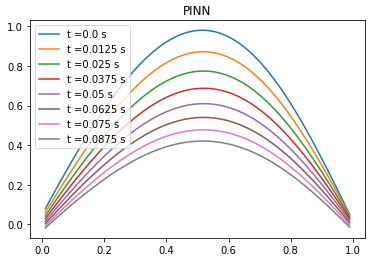}

\caption{PINN, analytical and finite difference solution side by side}
\label{fig:figure3}

\end{figure}

\begin{figure}[htp]

\centering
\includegraphics[width=.5\textwidth]{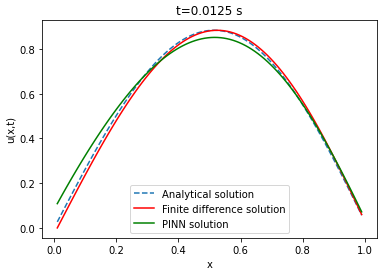}\hfill
\includegraphics[width=.5\textwidth]{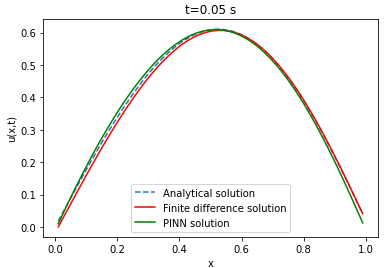}\hfill
\includegraphics[width=.5\textwidth]{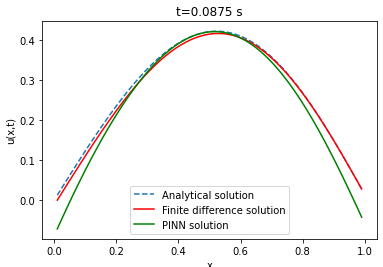}

\caption{PINN, analytical and finite difference at t=0.0125 s, t=0.05 s and t = 0.0875}
\label{fig:figure3}

\end{figure}

\begin{lstlisting}[language=Python, caption= Implementation of the comparison between the 3 methods of resolution.]

#RMSE
print("RMSE PINN vs Finite difference = ", np.sqrt(np.mean( (Unp[1:99,:] - u_df)**2) ) )
print("RMSE PINN vs Analytical solution = ", np.sqrt(np.mean((Unp[1:99,:] - u_an[1:99,:])**2) ) )
print("RMSE Finite difference vs Analytical solution = ", np.sqrt(np.mean((u_an[1:99,:] - u_df)**2)) )

#RMSE PINN vs Finite difference =  0.03899286511045842
#RMSE PINN vs Analytical solution =  0.033786413268837606
#RMSE Finite difference vs Analytical solution =  0.011699786048706274

\end{lstlisting}

\begin{table}[H]
\centering
\begin{tabular}{ll|llllll|}                                                                                                                                               \\ \cline{3-8} 
                                             &   & \multicolumn{3}{l|}{Analytical solution}                                                      & \multicolumn{3}{l|}{Finite difference}                                   \\ \cline{3-8} 
                                             &   & \multicolumn{1}{l|}{8}        & \multicolumn{1}{l|}{16}       & \multicolumn{1}{l|}{32}       & \multicolumn{1}{l|}{8}        & \multicolumn{1}{l|}{16}       & 32       \\ \hline
{Layers} & 2 & \multicolumn{1}{l|}{0.020769} & \multicolumn{1}{l|}{0.014624} & \multicolumn{1}{l|}{0.011578} & \multicolumn{1}{l|}{0.202538} & \multicolumn{1}{l|}{0.021090} & 0.018698 \\ \cline{2-8} 
\multicolumn{1}{|l|}{}                       & 4 & \multicolumn{1}{l|}{0.010308} & \multicolumn{1}{l|}{0.029504} & \multicolumn{1}{l|}{0.033058} & \multicolumn{1}{l|}{0.018142} & \multicolumn{1}{l|}{0.034560} & 0.036699 \\ \cline{2-8} 
\multicolumn{1}{|l|}{}                       & 8 & \multicolumn{1}{l|}{0.212536} & \multicolumn{1}{l|}{0.212404} & \multicolumn{1}{l|}{0.009440} & \multicolumn{1}{l|}{0.209136} & \multicolumn{1}{l|}{0.209136} & 0.018677 \\ \hline
\end{tabular}
\caption{\label{tab:Time 1}RSME between PINN and analytical solution for burger's equation}
\end{table}

\begin{table}[H]
\centering
\begin{tabular}{ll|lll|}
\cline{3-5}
                                              &   & \multicolumn{3}{l|}{Neurons}                                 \\ \cline{3-5} 
                                              &   & \multicolumn{1}{l|}{8}    & \multicolumn{1}{l|}{16}   & 32   \\ \hline
{Layers} & 2 & \multicolumn{1}{l|}{2:00} & \multicolumn{1}{l|}{2:03} & 2:38 \\ \cline{2-5} 
\multicolumn{1}{|l|}{}                        & 4 & \multicolumn{1}{l|}{2:37} & \multicolumn{1}{l|}{4:32} & 5:02 \\ \cline{2-5} 
\multicolumn{1}{|l|}{}                        & 8 & \multicolumn{1}{l|}{4:39} & \multicolumn{1}{l|}{8:04} & 9:52 \\ \hline
\end{tabular}
\caption{\label{tab:Time 1}Execution time according to the number of layers and neurons for the burger'equation}
\end{table}

 \subsection{Heat equation: good for diffusion but poor for spatial heterogeneity}
In order to compare between the three resolution approaches: the analytical approach, the numerical approach with finite differences and the PINN; we propose to start by the analytical method. We solve the heat equation by considering a constant $\alpha$ parameter.
 
 $$\frac{\partial u}{\partial t} =  \alpha(\frac{\partial^{2} u}{\partial x^{2}}+\frac{\partial^{2} u}{\partial y^{2}}) \hspace{0.2cm} x \in [-10,10],\hspace{0.2cm} y \in [-10,10],\hspace{0.2cm} t \in [0, 0.25]$$
 
with the initial condition

$u(x,y,0) = e^{-x^{2} - y^{2}}$

and the boundary conditions

$u(0,y,t)=u(x,0,t)=0$

We take $\alpha = 2$.

\subsection*{Analytical solution}
The analytical solution is 

$$ u(x,y,t) = \frac{\sqrt{2 \sigma}}{\sqrt{4 D t + 2 \sigma}} e^{- \frac{x^2 + y^2}{4 D t}}$$

With the following Fourier transform applied:

$$ u(k_x, k_y)=\frac{1}{\sqrt{2 \pi}} \int_{- \infty}^{+ \infty} u(x,y) e^{-i(k_x x+k_y y)} $$

\subsection*{Numerical schema}
The solution domain is discretized into cells described by the node set $(x_{i}, y_{j}, t_{n})$ in which $x_{j} =ih$, $y_{j}=jl$, $t_{n}=nk$ (i = 0,1,...,I; i = 0,1,...,J ; n = 0,1,...,N) $h=\Delta x$ and $l=\Delta y$ are a spatial mesh size, $k=\Delta t$ is the time step and $u(x_{i},y_{j}, t_{n})=u_{i,j}^{n}$

$$u_{i,j}^{n+1} = u_{i,j}^{n} + \alpha \frac{\Delta t}{\Delta x} \frac{u_{i+1,j}^{n} - 2u_{i,j}^{n} u_{i-1,j}^{n}}{(\Delta x)^{2}} + \alpha \frac{\Delta t}{\Delta y} \frac{u_{i,j+1}^{n} - 2u_{i,j}^{n} u_{i,j-1}^{n}}{(\Delta y)^{2}}$$

\subsection*{PINN approach}

Let us define $f(x, y, t)$  as :

 $$ f := \frac{\partial u}{\partial t} - \alpha(\frac{\partial^{2} u}{\partial x^{2}}+\frac{\partial^{2} u}{\partial y^{2}})$$

In this case, $u(x, y, t)$ will be approximated by a neural network. The latter has as an objective to minimize the following loss :

$$ MSE = MSE_0 + MSE_b + MSE_f $$

where 

$$ MSE_0 = \frac{1}{N_0} \sum_{n=1}^{N_0} |u(x_{0}^{i}, y_{0}^{i}, 0) - u_{0}^{i}|^{2} $$

$$ MSE_b = \frac{1}{N_b} \sum_{n=1}^{N_b} (|u(0, y_{b}^{i}, t_{b}^{i}) - u_{b}^{i}|^{2} + |u(x_{b}^{i}, 0, t_{b}^{i}) - u_{b}^{i}|^{2}) $$

$$ MSE_f = \frac{1}{N_f} \sum_{n=1}^{N_f} |f(x_{f}^{i}, y_{f}^{i}, t_{f}^{i})|^{2} $$

$ \{x_{0}^{i}, y_{0}^{i}, u_{0}^{i}\} $ denotes the initial data at $t=0$, $ \{t_{b}^{i}, x_{b}^{i}, y_{b}^{i}, u_{b}^{i}\} $ denotes to the boundary data and $ \{ x_{f}^{i}, y_{f}^{i}, t_{f}^{i}\} $ corresponds to collocation points on $f(x, t)$.

\subsection*{Comparison between PINN, analytical solution and finite difference method}

\begin{figure}[H]
   \centering
    \includegraphics[scale=0.6]{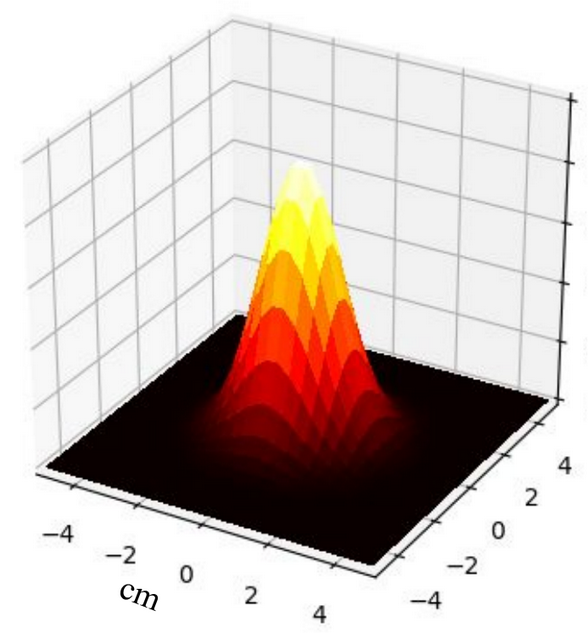}
    \includegraphics[scale=0.6]{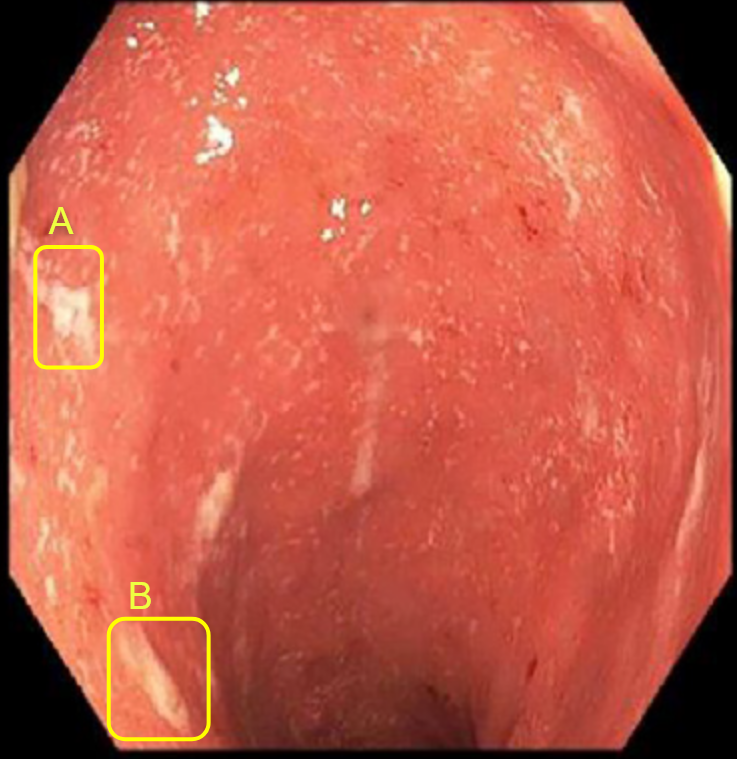}
    \caption{Colonoscopy images showing a moderate colitis. 
    The figure shows the initial state of the diffusion process solved by an analytical approach using a Fourier transform. The image is extracted from this paper \citep{Gajendran2019}.}
   \label{fig:blckbx}
\end{figure}

\begin{figure}[H]
   \centering
    \includegraphics[scale=0.6]{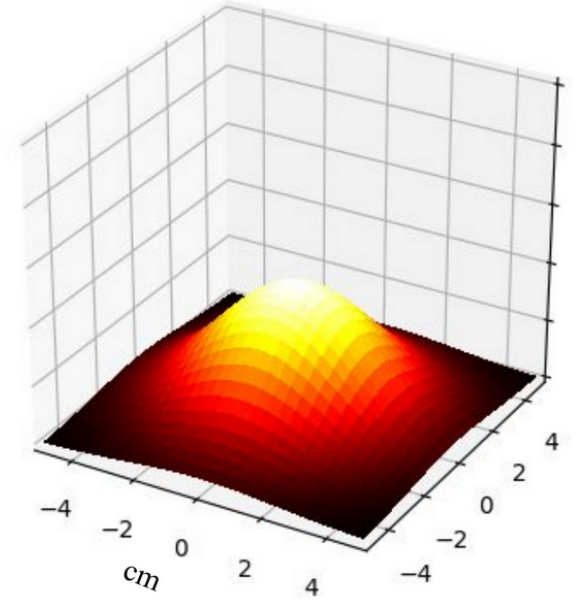}
    \includegraphics[scale=0.6]{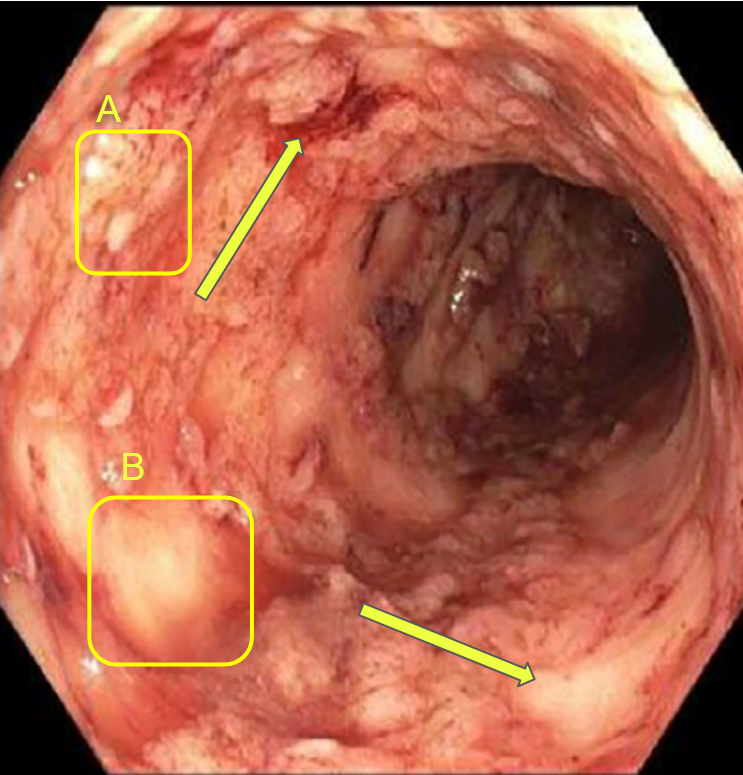}
    \caption{Colonoscopy images showing a severe colitis. The figure shows the final state of the diffusion process solved by an analytical approach using a Fourier transform. The image is extracted from this paper \citep{Gajendran2019}. The heat equation model may not be able to accurately predict the spatial distribution of the disease as shown by the two arrows.}
   \label{fig:blckbx}
\end{figure}

\begin{table}[H]
\centering
\begin{tabular}{ll|llllll|}
\cline{3-8}
\multicolumn{2}{l|}{}           & \multicolumn{3}{l|}{Analytical method} & \multicolumn{3}{l|}{Finite difference}                                \\ \cline{3-8} 
\multicolumn{2}{l|}{}                             & \multicolumn{6}{l|}{Number of neurons}                                                                                                                             \\ \cline{3-8} 
\multicolumn{2}{l|}{}                             & \multicolumn{1}{l|}{8}       & \multicolumn{1}{l|}{16}      & \multicolumn{1}{l|}{32}      & \multicolumn{1}{l|}{8}       & \multicolumn{1}{l|}{16}      & 32      \\ \hline
\multicolumn{1}{|l|}{Layers} & 2 & \multicolumn{1}{l|}{0.01104} & \multicolumn{1}{l|}{0.02231} & \multicolumn{1}{l|}{0.00507} & \multicolumn{1}{l|}{0.01102} & \multicolumn{1}{l|}{0.02230} & 0.00506 \\ \cline{2-8} 
\multicolumn{1}{|l|}{}                        & 4 & \multicolumn{1}{l|}{0.00289} & \multicolumn{1}{l|}{0.02391} & \multicolumn{1}{l|}{0.00601} & \multicolumn{1}{l|}{0.00289} & \multicolumn{1}{l|}{0.02391} & 0.00599 \\ \cline{2-8} 
\multicolumn{1}{|l|}{}                        & 8 & \multicolumn{1}{l|}{0.04588} & \multicolumn{1}{l|}{0.04613} & \multicolumn{1}{l|}{0.04307} & \multicolumn{1}{l|}{0.04586} & \multicolumn{1}{l|}{0.04612} & 0.04308 \\ \hline
\end{tabular}
\caption{Comparaison of the RMSE between the (analytical method and the finite difference) and the PINN using different parameters}
\end{table}
\begin{table}[H]
\centering
\begin{tabular}{cc|ccc|}
\cline{3-5}
                                               &    & \multicolumn{3}{c|}{Time of execution}                        \\ \cline{3-5} 
                                               &    & \multicolumn{3}{c|}{Layers}                                  \\ \cline{3-5} 
                                               &    & \multicolumn{1}{c|}{2}    & \multicolumn{1}{c|}{4}    & 8    \\ \hline
\multicolumn{1}{|c|}{Neurons} & 8  & \multicolumn{1}{c|}{1:56} & \multicolumn{1}{c|}{3:22} & 5:40 \\ \cline{2-5} 
\multicolumn{1}{|c|}{}                         & 16 & \multicolumn{1}{c|}{1:39} & \multicolumn{1}{c|}{3:19} & 5:58 \\ \cline{2-5} 
\multicolumn{1}{|c|}{}                         & 32 & \multicolumn{1}{c|}{1:34} & \multicolumn{1}{c|}{2:17} & 9:48 \\ \hline

\end{tabular}
\caption{Time of training by minutes for the PINN model using different parameters}
\end{table}
 
\begin{figure}[H]
   \centering
    \includegraphics[scale=0.55]{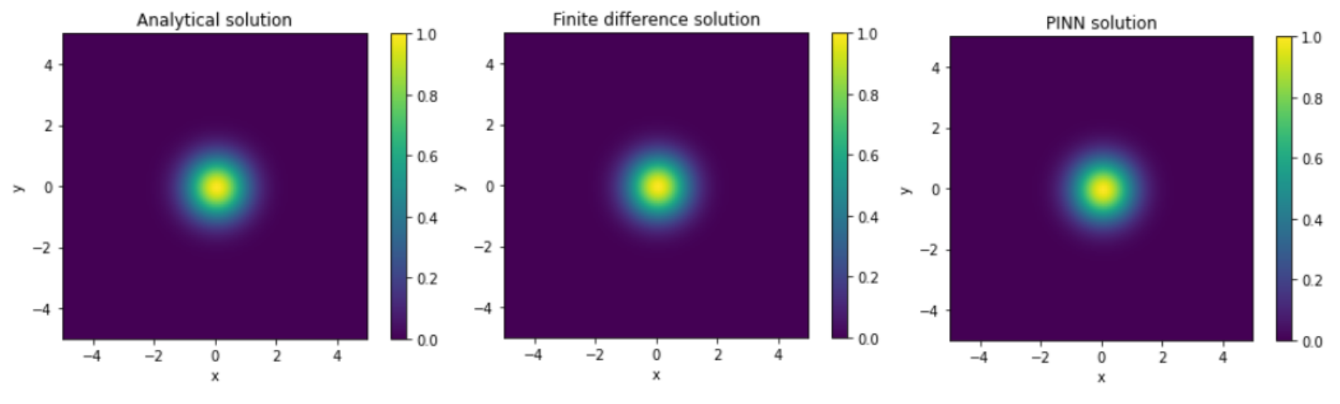}
    \caption{An illustration showing the comparison of three different methods for solving the heat equation, including an analytical solution, PINN and a finite difference method at time t=0.}
   \label{fig:Heat_Equation_t0}
\end{figure}

\begin{figure}[H]
   \centering
    \includegraphics[scale=0.55]{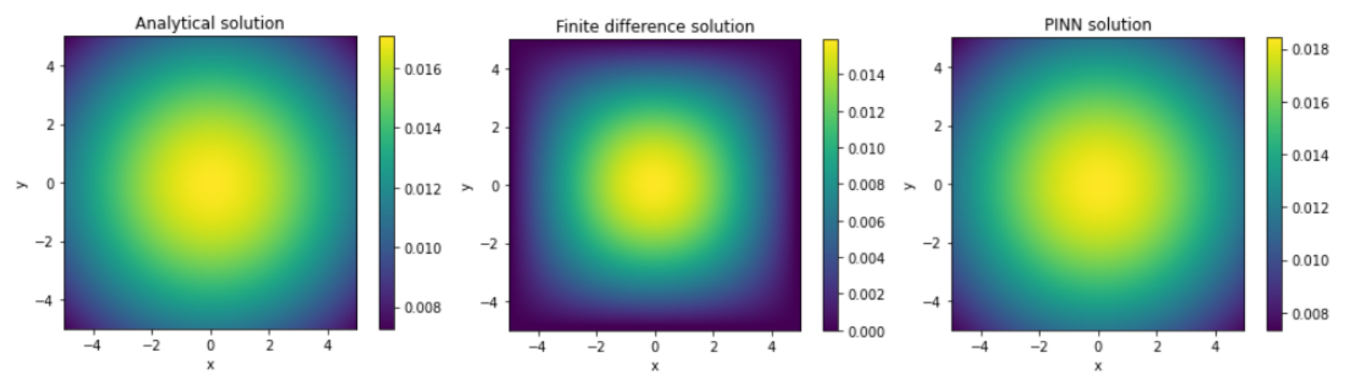}
    \caption{An illustration showing the comparison of three different methods for solving the heat equation, including an analytical solution, PINN, and a finite difference method at time t=1.}
   \label{fig:Heat_Equation_t1}
\end{figure}

The RMSE between the analytical solution and the solution obtained from the PINN is found to be on average of the order of $10^{-2}$. This indicates that the PINN method is able to approximate the analytical solution with a certain level of accuracy. Furthermore, the RMSE between the finite difference solution and the PINN solution is also found to be of the same order. However, a direct correlation between the RMSE and the increase in the number of layers and neurons per layer in the PINN is not observed. Nevertheless, some configurations of the PINN are able to approximate the analytical solution as well as the finite difference, demonstrating the effectiveness of this method for solving the heat equation.

\subsection{System of Korteweg-De Vries Equations}
The Korteweg-de Vries (KDV) equation is a non-linear partial differential equation that describes solitary waves, also known as solitons, and arises from the Navier-Stokes equation, by neglecting the dispersion term. Solitons, are responsible for tsunamis and tidal bores, which can cause significant damage and threaten marine safety. To prevent such disasters, geophysicists study the production process and formation conditions of these waves. Solitary waves are rare maritime occurrences that were first observed by John Scott Russell in 1834. They are exceedingly dangerous due to their unexpected nature and amplitude, which far exceeds the height of swell waves. Characterizing the propagation mode of solitons can help identify them, as they propagate while preserving nearly all of the initial energy. Solitons can be explained using nonlinear models, as they do not conform to linear approaches like the Airy model. The speed of a solitary wave is proportional to its amplitude, a phenomenon known as wave front stiffening, which causes its nonlinear action. The Korteweg de Vries equation is derived from the assumption that the non-linear effect is added to the linear effect reflecting the wave's dispersion, and the soliton is the compensation of the two effects. Soliton solutions can be obtained using numerical analysis, such as the finite difference method or the Lax-Wendrof approach, which can solve the KDV and Burger's equations. Solitons can be found in various fields of physics, including solid mechanics and optics, and have real-world applications such as data transmission in telecommunications. PINN approach can be used to solve such equations \citep{Xiao-KDV}.

\subsection*{Equation}
The coupled Korteweg-De Vries Equations are given by

$$\large{
\begin{cases}
      \frac{\partial u}{\partial t} = 6au\frac{\partial u}{\partial x} - 2bv\frac{\partial v}{\partial x} - a\frac{\partial^{3} u}{\partial x^{3}}, \hspace{0.3cm} x \in [-250,250], \hspace{4mm} t \in [0,10]\\
      \frac{\partial v}{\partial t} = -3au\frac{\partial v}{\partial x} + \frac{\partial^{3} v}{\partial x^{3}} 
\end{cases}
}$$

with the initial conditions:

$u(x,0) = F(x), \hspace{4mm} v(x,0) = G(x) \hspace{2mm} \text{with x} \in [-250,250]$

and the boundary conditions:

$u(-250,t)=v(-250,t)=0$

$u(250,t)=v(250,t)=0$

\subsection*{Analytical solution}
The analytical solution of the equation is the couple (u,v) with

$$u(x,t) = 2 \lambda^{2} sech^{2}(\xi)$$

$$v(x,t) = \frac{1}{2w^{1/2}} sech(\xi) $$

where:

$ \xi = \lambda(x - \lambda^{2}t) + \frac{1}{2 \log(w)}$

$ w = \frac{-b}{8(4a+1)\lambda^{4}}$

In this paper, we take $a=- \frac{1}{8}$, b=-3, $\lambda = \frac{1}{2}$ 

\subsection*{Numerical schema}
The solution domain is discretized into cells described by the node set $(x_{i}, t_{n})$ in which $x_{j}=ih$, $t_{n}=nk$ (i = 0,1,...,I; n = 0,1,...,N) $h=\Delta x$  is a spatial mesh size, $k=\Delta t$ is the time step and $u(x_{i}, t_{n})=u_{i}^{n}$

$$u_{i}^{n+1} = u_{i}^{n} - \frac{3}{4} \frac{\Delta t}{\Delta x}u_{i}^{n}(u_{i}^{n} - u_{i-1}^{n}) + 6 \frac{\Delta t}{\Delta x}v_{i}^{n}(v_{i}^{n} - v_{i-1}^{n}) + \frac{1}{16} \frac{\Delta t}{(\Delta x)^{3}} (u_{i+2}^{n}-2u_{i+1}^{n}+2u_{i-1}^{k}-u_{i-2}^{k})$$

$$v_{i}^{n+1} = v_{i}^{n} - 3 \frac{\Delta t}{\Delta x}u_{i}^{n}(v_{i}^{k} - v_{i-1}^{k}) + \frac{1}{2}\frac{\Delta t}{(\Delta x)^{3}}(v_{i+2}^{k}-2v_{i+1}^{k}+2v_{i-1}^{k}-v_{i-2}^{k}) $$

\begin{lstlisting}[language=Python, caption=An implementation of the PINN resolution for the KdV system.]

import torch
import numpy as np
from tqdm import tqdm
import matplotlib.pyplot as plt
import torch.nn as nn
from random import uniform
device = torch.device('cuda:0' if torch.cuda.is_available() else 'cpu')
torch.set_default_tensor_type('torch.cuda.FloatTensor')
print(device)

def omega(a,b,lambd):
  return -b/(8*(4*a+1)*lambd**4)

def xi(x,t,a,b,lambd):
  return lambd*(x-t*lambd**2) + 1/(2*np.log(omega(a,b,lambd)))

def fct_u(x,t,a,b,lambd):
  return (2*lambd**2)/(np.cosh(xi(x,t,a,b,lambd))**2)

def fct_v(x,t,a,b,lambd):
  return 1/(2*np.sqrt(omega(a,b,lambd))*np.cosh(xi(x,t,a,b,lambd)))

a = -1/8
b = -3
lambd = 0.5

N_uv = 1000
N_f = 15000

#Make X_uv_train
#BC x=-10  & tt t > 0
x_left = np.ones((N_uv//4,1), dtype=float)*(-250)
t_left = np.random.uniform(low=0.001, high=10.0, size=(N_uv//4,1))
X_left = np.hstack((x_left, t_left))

#BC x=10 & tt t > 0
x_right = np.ones((N_uv//4,1), dtype=float)*(250)
t_right = np.random.uniform(low=0.001, high=10.0, size=(N_uv//4,1))
X_right = np.hstack((x_right, t_right))

#IC t=0 & tt x,y in [-10,10]
t_zero = np.zeros((N_uv//2,1), dtype=float)
x_zero = np.random.uniform(low=-250.0, high=250.0, size=(N_uv//2,1))
X_zero = np.hstack((x_zero, t_zero))

X_uv_train = np.vstack((X_left, X_right, X_zero))
# shuffling
index=np.arange(0,N_uv)
np.random.shuffle(index)
X_uv_train=X_uv_train[index,:]

#Make u_train
u_left = np.zeros((N_uv//4,1), dtype=float)
u_right = np.zeros((N_uv//4,1), dtype=float)
u_initial = fct_u(x_zero,0,a,b,lambd)
u_train = np.vstack((u_left, u_right, u_initial))

#Make v_train
v_left = np.zeros((N_uv//4,1), dtype=float)
v_right = np.zeros((N_uv//4,1), dtype=float)
v_initial = fct_v(x_zero,0,a,b,lambd)
v_train = np.vstack((v_left, v_right, v_initial))

# ==========================================
u_train=u_train[index,:]
v_train=v_train[index,:]
# ==========================================
# make X_f_train 
X_f_train=np.zeros((N_f,2),dtype=float)
for row in range(N_f):
    x=uniform(-250,250) 
    t=uniform(0,10)  
    X_f_train[row,0]=x
    X_f_train[row,1]=t
   
X_f_train=np.vstack((X_f_train, X_uv_train))

class PhysicsInformedNN():
  def __init__(self,X_uv,u,v,X_f):
    # x & t from boundary conditions
    self.x_uv = torch.tensor(X_uv[:, 0].reshape(-1, 1),dtype=torch.float32,requires_grad=True)
    self.t_uv = torch.tensor(X_uv[:, 1].reshape(-1, 1),dtype=torch.float32,requires_grad=True)
    # x & t from collocation points
    self.x_f = torch.tensor(X_f[:, 0].reshape(-1, 1),dtype=torch.float32,requires_grad=True)
    self.t_f = torch.tensor(X_f[:, 1].reshape(-1, 1),dtype=torch.float32,requires_grad=True)
    # boundary solution
    self.u = torch.tensor(u, dtype=torch.float32)
    self.v = torch.tensor(v, dtype=torch.float32)
    # null vector to test against f:
    self.null =  torch.zeros((self.x_f.shape[0], 1))
    # initialize net:
    self.create_net()

    self.optimizer = torch.optim.Adam(self.net.parameters(),lr=0.001)
    # typical MSE loss (this is a function):
    self.loss = nn.MSELoss()
    # loss :
    self.ls = 0
    # iteration number:
    self.iter = 0

  def create_net(self):
    self.net=nn.Sequential(
        nn.Linear(2,32), nn.Sigmoid(),
        nn.Linear(32, 32), nn.Sigmoid(),
        nn.Linear(32, 32), nn.Sigmoid(),
        nn.Linear(32, 32), nn.Sigmoid(),
        nn.Linear(32, 32), nn.Sigmoid(),
        nn.Linear(32, 32), nn.Sigmoid(),
        nn.Linear(32, 32), nn.Sigmoid(),
        
        nn.Linear(32, 2))
  def net_uv(self,x,t):
    uv=self.net(torch.hstack((x,t)))
    return uv


  def net_fg(self,x,t):
    uv = self.net_uv(x,t)
    u = uv[:,0].reshape(-1,1).to(device)
    v = uv[:,1].reshape(-1,1).to(device)

    #u partial derivatives
    u_t = torch.autograd.grad(u,t, grad_outputs=torch.ones_like(u), create_graph=True)[0]
    u_x = torch.autograd.grad(u,x, grad_outputs=torch.ones_like(u), create_graph=True)[0]
    u_xx = torch.autograd.grad(u_x,x, grad_outputs=torch.ones_like(u), create_graph=True)[0]
    u_xxx = torch.autograd.grad(u_xx,x, grad_outputs=torch.ones_like(u), create_graph=True)[0]

    #v partial derivatives
    v_t = torch.autograd.grad(v,t, grad_outputs=torch.ones_like(u), create_graph=True)[0]
    v_x = torch.autograd.grad(v,x, grad_outputs=torch.ones_like(u), create_graph=True)[0]
    v_xx = torch.autograd.grad(v_x,x, grad_outputs=torch.ones_like(u), create_graph=True)[0]
    v_xxx = torch.autograd.grad(v_xx,x, grad_outputs=torch.ones_like(u), create_graph=True)[0]

    f = u_t - 6*a*u*u_x -2*b*v*v_x - a*u_xxx
    g = v_t +3*u*v_x + v_xxx

    f = f.to(device)
    g = g.to(device)

    return f,g

  def closure(self):

    # reset gradients to zero 
    self.optimizer.zero_grad()
    
    # u & f predictions
    uv_prediction = self.net_uv(self.x_uv, self.t_uv)
    u_prediction = uv_prediction[:,0].reshape(-1,1)
    v_prediction = uv_prediction[:,1].reshape(-1,1)
    #
    f_prediction_u, f_prediction_v = self.net_fg(self.x_f, self.t_f)

    #
    # losses:
    u_loss = self.loss(u_prediction, self.u)
    v_loss = self.loss(v_prediction, self.v)
    f_loss_u = self.loss(f_prediction_u, self.null)
    f_loss_v = self.loss(f_prediction_v, self.null)

    self.ls = u_loss + v_loss + f_loss_u + f_loss_v
    
    # derivative with respect to net's weights:
    self.ls.backward()

    # increase iteration count:
    self.iter += 1

    # print report:
    if not self.iter % 100:
      print('Epoch: {0:}, Loss: {1:6.8f}'.format(self.iter, self.ls))
      return self.ls    
        
  def train(self):
    for epoch in range(15000):
      self.net.train()
      self.optimizer.step(self.closure)

pinn = PhysicsInformedNN(X_uv_train,u_train,v_train, X_f_train)
pinn.train()

import matplotlib.pyplot as plt
from mpl_toolkits.axes_grid1 import make_axes_locatable
#     
x = torch.linspace(-250, 250, 500)
t = torch.linspace( 0, 10, 500)
# x & t grids:
X,T = torch.meshgrid(x,t)
# x & t columns:
xcol = X.reshape(-1, 1)
tcol = T.reshape(-1, 1)
# one large column:
sol = pinn.net_uv(xcol, tcol)
usol = sol[:,0].reshape(-1,1)
vsol = sol[:,1].reshape(-1,1)
# reshape solution:
U = usol.reshape(x.numel(), t.numel())
V = vsol.reshape(x.numel(), t.numel())
# transform to numpy:
xnp = x.cpu().numpy()
tnp = t.cpu().numpy()
Unp = U.cpu().detach().numpy()
Vnp = V.cpu().detach().numpy()

plt.plot(np.linspace(-250,250,500), fct_u(np.linspace(-250,250,500),100*0.02,a,b,lambd), label='analytic', color="red")
plt.plot(np.linspace(-250,250,500),Unp[:,100], label='pinn', linestyle='dashed')
plt.xlim(-30,30)
plt.legend()

plt.plot(np.linspace(-250,250,500), fct_u(np.linspace(-250,250,500),0.02*300,a,b,lambd), label='analytic', color='red')
plt.plot(np.linspace(-250,250,500),Unp[:,300], label='pinn', linestyle='dashed')
plt.xlim(-50,50)
plt.legend()

plt.plot(np.linspace(-250,250,500), fct_v(np.linspace(-250,250,500),25*0.02,a,b,lambd), label='analytic', color='red')
plt.plot(np.linspace(-250,250,500),Vnp[:,25], label='pinn', linestyle='dashed')
plt.xlim(-50,50)
plt.legend()

x = np.linspace(-250,250,500)
t = np.linspace(0,10,500)

ms_x, ms_t = np.meshgrid(x,t)

U_an = fct_u(ms_x, ms_t, a,b, lambd)
V_an = fct_v(ms_x, ms_t, a,b, lambd)

#RMSE analytic & PINN
np.sqrt(np.mean((Unp-U_an.T)**2))
# 0.00152934398676765

#RMSE analytic & PINN
np.sqrt(np.mean((Vnp-V_an.T)**2))
#0.001979420025691176

a = -1/8
b = -3
lambd = 0.5

delta_x = 1
delta_t = 0.02

x = np.arange(-250,250,delta_x)
t = np.arange(0,10,delta_t)

alpha = delta_t/delta_x
beta = delta_t/(delta_x**3)

n = len(x)
m = len(t)

u_df = np.zeros((n,m))
v_df = np.zeros((n,m))

#For u
#BC
u_df[0,:] = 0
u_df[-1,:] = 0

#IC
u_df[:,0] = fct_u(x,0,a,b,lambd)

#For v
#BC
v_df[0,:] = 0
v_df[-1,:] = 0

#IC
v_df[:,0] = fct_v(x,0,a,b,lambd)

for k in tqdm(range(0,m-1)):
  for i in range(2,n-3):
    u_df[i,k+1] = u_df[i,k] + 6*a*alpha*u_df[i,k]*(u_df[i,k]-u_df[i-1,k]) - 2*b*alpha*v_df[i,k]*(v_df[i,k]-v_df[i-1,k]) - 0.5*a*beta*(u_df[i+2,k]-2*u_df[i+1,k]+2*u_df[i-1,k]-u_df[i-2,k])
    v_df[i,k+1] = v_df[i,k] - 3*alpha*u_df[i,k]*(v_df[i,k]-v_df[i-1,k]) + 0.5*beta*(v_df[i+2,k]-2*v_df[i+1,k]+2*v_df[i-1,k]-v_df[i-2,k]) 

#RMSE PINN & finite difference
np.sqrt(np.mean((Unp-u_df)**2))
# 0.010952436812997277

#RMSE analytic & finite difference
np.sqrt(np.mean((U_an.T-u_df)**2))
# 0.010989140837703614

#RMSE PINN & finite difference
np.sqrt(np.mean((Vnp-v_df)**2))
# 0.006990164623215043

#RMSE analytic & finite difference
np.sqrt(np.mean((V_an-v_df)**2))
# 0.015797690668562524

\end{lstlisting}

\subsection*{Comparison beetween PINN, analytical solution and finite difference method}
\begin{figure}[H]%
    \centering
    \subfloat{{\includegraphics[width=8cm]{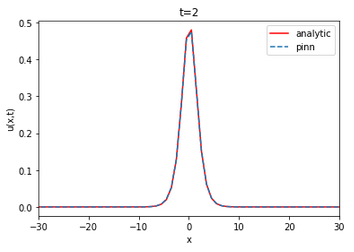} }}%
    \qquad
    \subfloat{{\includegraphics[width=8cm]{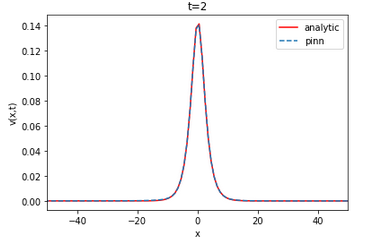} }}%
    \caption{The figure on the left shows 'u' while the right shows 'v' at t=2. Both solutions are plotted on the analytical method }%
    \label{fig:example}%
\end{figure}

\begin{figure}[H]%
    \centering
    \subfloat{{\includegraphics[width=8cm]{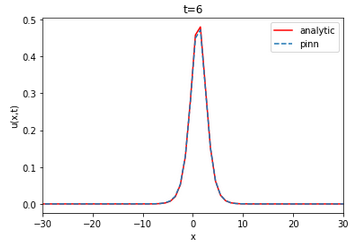} }}%
    \qquad
    \subfloat{{\includegraphics[width=8cm]{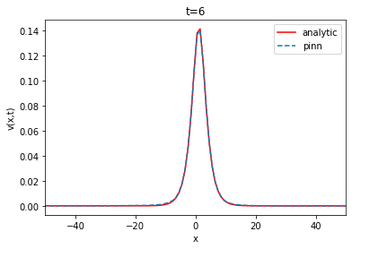} }}%
    \caption{The figure on the left shows 'u' while the right shows 'v' at t=6. Both solutions are plotted on the analytical method.}%
    \label{fig:example}%
\end{figure}
\begin{table}[H]%
\begin{tabular}{ll|cccccc|cccccc|}
\cline{3-14}
                                              &                         & \multicolumn{6}{c|}{Analytical}                                                                                                                            & \multicolumn{6}{c|}{Finite difference}                                                                                                                  \\ \cline{3-14} 
                                              &                         & \multicolumn{6}{c|}{Layers}                                                                                                                                & \multicolumn{6}{c|}{Layers}                                                                                                                             \\ \cline{3-14} 
                                              &                         & \multicolumn{2}{c|}{2}                                   & \multicolumn{2}{c|}{4}                                    & \multicolumn{2}{c|}{8}              & \multicolumn{2}{c|}{2}                                  & \multicolumn{2}{c|}{4}                                   & \multicolumn{2}{c|}{8}             \\ \cline{3-14} 
                                              &                         & \multicolumn{1}{c|}{u}     & \multicolumn{1}{c|}{v}      & \multicolumn{1}{c|}{u}      & \multicolumn{1}{c|}{v}      & \multicolumn{1}{c|}{u}     & v      & \multicolumn{1}{c|}{u}     & \multicolumn{1}{c|}{v}     & \multicolumn{1}{c|}{u}      & \multicolumn{1}{c|}{v}     & \multicolumn{1}{c|}{u}     & v     \\ \hline
\multicolumn{1}{|c|}{N} & \multicolumn{1}{c|}{8}  & \multicolumn{1}{c|}{0.007} & \multicolumn{1}{c|}{0.002}  & \multicolumn{1}{c|}{0.008}  & \multicolumn{1}{c|}{0.0037} & \multicolumn{1}{c|}{0.036} & 0.012  & \multicolumn{1}{c|}{0.01}  & \multicolumn{1}{c|}{0.006} & \multicolumn{1}{c|}{0.0101} & \multicolumn{1}{c|}{0.005} & \multicolumn{1}{c|}{0.01}  & 0.009 \\ \cline{2-14} 
\multicolumn{1}{|c|}{}                        & \multicolumn{1}{c|}{16} & \multicolumn{1}{c|}{0.005} & \multicolumn{1}{c|}{0.0019} & \multicolumn{1}{c|}{0.0028} & \multicolumn{1}{c|}{0.007}  & \multicolumn{1}{c|}{0.036} & 0.012  & \multicolumn{1}{c|}{0.009} & \multicolumn{1}{c|}{0.006} & \multicolumn{1}{c|}{0.0102} & \multicolumn{1}{c|}{0.006} & \multicolumn{1}{c|}{0.033} & 0.009 \\ \cline{2-14} 
\multicolumn{1}{|c|}{}                        & \multicolumn{1}{c|}{32} & \multicolumn{1}{c|}{0.005} & \multicolumn{1}{c|}{0.0063} & \multicolumn{1}{c|}{0.0029} & \multicolumn{1}{c|}{0.0012} & \multicolumn{1}{c|}{0.037} & 0.0065 & \multicolumn{1}{c|}{0.01}  & \multicolumn{1}{c|}{0.006} & \multicolumn{1}{c|}{0.0107} & \multicolumn{1}{c|}{0.006} & \multicolumn{1}{c|}{0.033} & 0.009 \\ \hline
\end{tabular}
\caption{Comparaison of the RMSE between the (analytical method and the finite difference) and the PINN using different parameters for the Korteweg-De Vries Equations.}
\end{table}

The value of the RMSE between the analytical solution and the finite difference is approximately 0.022733 and 0.011562 for u and v, respectively. The RMSE in relation to the number of layers and neurons is on average of the order of $10^{-3}$ for u and v. The neural network is able to approximate the finite difference solution of the coupled KDV equation. The RMSE value does not vary linearly with the increase or decrease of the number of layers and neurons per layer.

\section{PINN framework for the IBDs spread modeling by PDEs}
\begin{figure}[H]
   \centering
    \includegraphics[scale=0.6,angle=90]{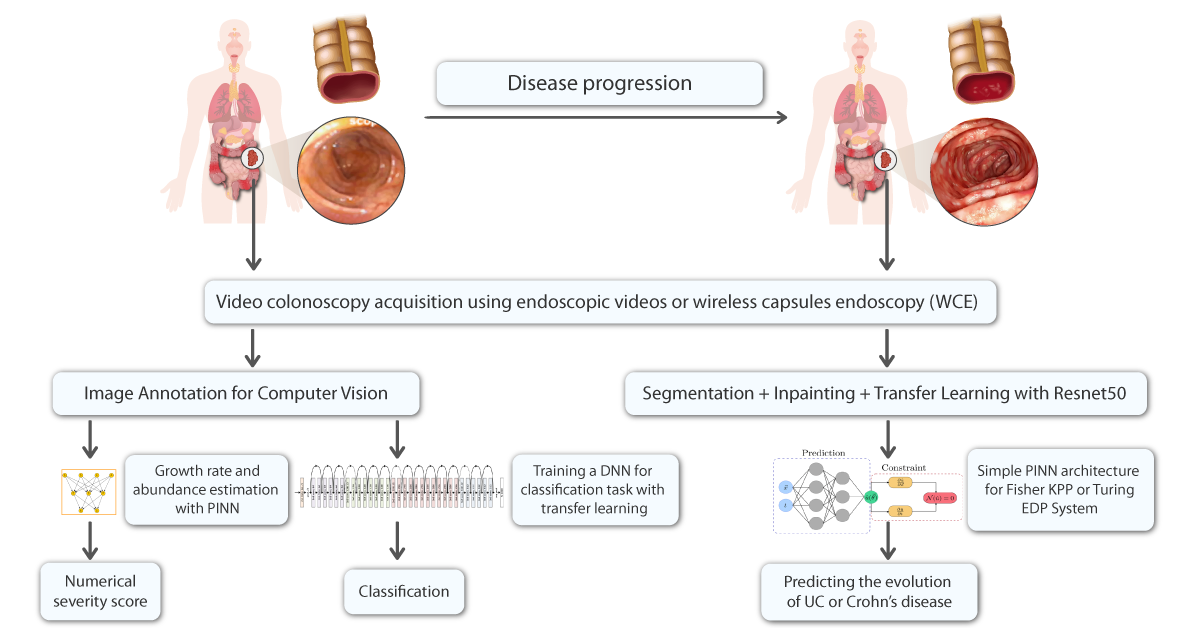}
    \caption{The experimental framework studied in our paper.}
   \label{fig:experimental_framework}
\end{figure}

In this section, we will discuss the combination of the concepts from image analysis and mathematical modeling. Using the PINN method, both of these areas are treated. Indeed, The figure \ref{fig:experimental_framework} shows a global approach that unifies the mathematical models which exploit the spatial information of the two diseases. 
\begin{itemize}
    \item The estimation of a score of the disease severity with the use of first-order ODE for spatial distribution. PINN is used to solve these equations and to estimate the score.
    \item A transfer learning model for the classification of the visual appearance of the different type of lesions. The deep learning model can use an image dataset annotated by gastroenterologists.
    \item Based on the spatial information extracted from images, PINN is used to predict the evolution and the spread of the two diseases.
    \item PINN is used to predict the lesions distributions along the colon.
\end{itemize}

\subsection{First-order ODE for spatial distribution}
In this part, an ordinary differential equation (ODE) has been used to describe the spatial distribution of colonic lesions in individuals with ulcerative colitis. This model is found effective in describing the disease state and there parameters were correlated with severity assessments provided by gastroenterologists \citep{safaaarticle, safaaposter}. Our contribution consists in finding an PINN architecture allowing to solve this equation by finding the two parameters the growth rate $\alpha$ and the abundance $c$ for the following first-order ODE model:

$$z'(s) = \alpha * z(s) $$ 

with $s \in [0,1]$

then the analytical solution is 

$$z(s) = c * e^{\alpha * s}$$

\begin{lstlisting}[language=Python, caption=A pytorch code predicting the solution of the problem.]
import torch
import torch.nn as nn
import torch.optim as optim

# Define the neural network architecture
class Net(nn.Module):
  def __init__(self, n_input, n_hidden_1, n_hidden_2, n_output):
    super(Net, self).__init__()
    self.fc1 = nn.Linear(n_input, n_hidden_1)
    self.fc2 = nn.Linear(n_hidden_1, n_hidden_2)
    self.fc3 = nn.Linear(n_hidden_2, n_output)
  
  def forward(self, x):
    x = self.fc1(x)
    x = self.fc2(x)
    x = self.fc3(x)
    return x

# Define the training loss
def loss(predicted_z, target_z):
  return torch.nn.functional.mse_loss(predicted_z, target_z)

# Define the physics-informed constraints
def ode_constraint(predicted_z, alpha):
  return predicted_z[1] - alpha * predicted_z[0]

def initial_condition_constraint(predicted_z, initial_z):
  return predicted_z[0] - initial_z

def train(target_z, initial_z, n_hidden_1, n_hidden_2, learning_rate, n_iter):
  # Create the neural network
  net = Net(2, n_hidden_1, n_hidden_2, 1)

  # Create the Adam optimizer
  optimizer = optim.Adam(net.parameters(), lr=learning_rate)

  # Train the neural network
  for i in range(n_iter):
    # Zero the gradients
    optimizer.zero_grad()

    # Forward pass
    predicted_z = net(t)

    # Compute the loss
    l = loss(predicted_z, target_z)

    # Compute the gradients of the loss
    l.backward()

    # Update the weights
    optimizer.step()

    # Compute the physics-informed constraints
    ode_error = ode_constraint(predicted_z, alpha)
    initial_condition_error = initial_condition_constraint(predicted_z, initial_z)

    # Compute the total error
    total_error = l + ode_error + initial_condition_error

    # Print the error every 100 iterations
    if i % 100 == 0:
      print(total_error)

# Set the hyperparameters
target_z = ...
initial_z = ...
n_hidden_1 = ...
n_hidden_2 = ...
learning_rate = ...
n_iter = ...

# Train the neural network
train(target_z, initial_z, n_hidden_1, n_hidden_2, learning_rate, n_iter)
\end{lstlisting}

\subsection{A possible explanation of the Crohn’s disease with a Turing mechanism}
\subsubsection{A first Turing mechanism presented in \citep{Nadin2021}}
Crohn’s disease is a chronic inflammatory bowel disease characterized by patchy inflammation throughout the gastrointestinal tract. In this study \citep{Nadin2021}, the authors propose a reaction-diffusion system that uses bacteria and phagocytic cells to model the dysfunctional immune responses that cause IBD. They demonstrate that, under specific conditions, the system can generate activator-inhibitor dynamics that lead to the formation of spatially periodic and time-indelible Turing patterns. This is the first time Turing patterns have been applied to an inflammation model, and the study compares the model parameters with realistic parameters from the literature.

The model represents the intestine as an interval on the real axis, taking into account only two components: external bacteria (microbiota, pathogens, or antigens) and immune cells (phagocytes). In a healthy gut, the immune response controls the inflammatory response, while in disease states, the intestinal immune system becomes unbalanced, resulting in excessive migration of immune cells to damaged areas and increased epithelial barrier permeability. These changes allow for further infiltration of the microbiota, which can exacerbate inflammation. A complex network of interactions between these factors initiates the inflammatory cascade that leads to the Crohn’s disease.

$$ \frac{\partial \beta(t,x,y)}{\partial t} - d_{b} \Delta \beta(t,x,y)= r_{b}(1 - \frac{\beta(t,x,y)}{b_{i}})\beta(t,x,y) - \frac{a \beta(t,x,y) \gamma(t,x,y)}{s_{b} + \beta(t,x,y)} + f_{e}(1 - \frac{\beta(t,x,y)}{b_{i}})\gamma(t,x,y) $$

$$ \frac{\partial \gamma(t,x,y)}{\partial t} - d_{c}\Delta \gamma(t,x,y) = f_{b} \beta(t,x,y) - r_{c} \gamma(t,x,y) $$

With the following initial conditions

$$\beta(0,x,y) = \beta_0(x,y)$$ and $$\gamma(0,x,y) = \gamma_0(x,y)$$ 

In our study, the boundary conditions of Neumann have been considered. In the paper, a periodic boundary conditions have been considered instead.

\begin{table}[H]
\centering
\begin{tabular}{|c|  c | c| }
    \hline
  parameter & description & value \\
  \hline
  $r_b$ & The bacteria reproduction rate per minute & $0.0347\;min^{-1}$  \\
  \hline
  $r_c$ & The phagocytes intrinsic death rate per minute & $0.02\;min^{-1}$  \\
  \hline
  $d_b$ & The bacteria diffusion rate per minute & $10^{-13}\;m^{2}*min^{-1}$  \\
  \hline
  $d_c$ & The phagocytes diffusion rate per minute & $10^{-10}\;m^{2}*min^{-1}$  \\
  \hline
  $b_i$ & The bacteria density in the lumen & $10^{17}\;m^{3}$  \\
  \hline
  $f_b$ & The rate of the immune response par minute & $0.002\;min^{-1}$  \\
  \hline
  $a$ & the product between $s_b$ and $p_c$ & $0.3129\;min^{-1}$  \\
  \hline
  $s_b$ & a cofficient of proportionality between $p_c$ and $a$ & $10^{15}\;m^{3}$  \\
  \hline
  $f_e$ & The rate per minute of the epithelium porosity  & $0.0856\;min^{-1}$  \\
  \hline

\end{tabular}
\caption{
\label{tab:turing_parameters} The table shows the value of the different parameters used in the Turing system. The different values are extracted from \citep{Nadin2021}.}
\end{table}

\begin{lstlisting}[language=Python, caption=The code exposes the resolution of the 1D system by a finite difference schema.]
import numpy as np
import matplotlib.pyplot as plt

def reac_diff_solver_1D():
    L = 3000
    N = 3000
    T = 200000
    dt = 1

    x = np.linspace(0, L, N)
    dx = x[1] - x[0]
    t = np.arange(0, T+dt, dt)
    t_num = len(t)

    d_b = 10**0
    d_c = 10**5

    # r_c = 0.0002 # To get a Turing process
    r_c = 2  # This value kill the Turing process
    b_i = 10**17  # To get a Turing process
    r_b = 0.0347
    k = 0.1  # (c*=k*b*)
    f_b = k*r_c

    s_b = 10**15
    theta = 0.3  # (b*=theta*b_i)
    alpha = 0.3129

    f_e = alpha*theta*b_i/((s_b+theta*b_i)*(1-theta))-r_b/k

    Kb = dt/(dx**2)*d_b
    Kc = dt/(dx**2)*d_c

    b0 = 1*10**15*(1495 < x)*(x < 1505)

    c0 = np.zeros_like(b0)

    Ab = np.diag((1+2*Kb)*np.ones(N)) - np.diag(Kb*np.ones(N-1), 1) - np.diag(Kb*np.ones(N-1), -1)
    Ab[0, 0] = 1 + Kb
    Ab[-1, -1] = 1 + Kb

    Ac = np.diag((1+2*Kc+dt*r_c)*np.ones(N)) - np.diag(Kc*np.ones(N-1), 1) - np.diag(Kc*np.ones(N-1), -1)
    Ac[0, 0] = 1 + Kc + dt*r_c
    Ac[-1, -1] = 1 + Kc + dt*r_c

    B = np.zeros((t_num, N))
    B[0, :] = b0

    C = np.zeros((t_num, N))
    C[0, :] = c0

    source_b = np.zeros(N)
    source_c = np.zeros(N)
    MD_b = np.zeros(N)
    MD_c = np.zeros(N)
    
    plt.figure()
    plt.plot(10**(-6)*x, B[0, :], "blue", 10**(-6)*x, C[0, :], "red")
    for j in range(1, t_num):
        
        MD_b = B[j-1, :] + dt*f_e*C[j-1, :]*(1 - B[j-1, :]/b_i) + dt*r_b*(1 - B[j-1, :]/b_i)*B[j-1, :] - dt*alpha*B[j-1, :]*C[j-1, :]/(s_b + B[j-1, :])
        Ab_extended = Ab
        B[j, :] = np.linalg.solve(Ab_extended, MD_b)
        MD_c = C[j-1, :] + dt*f_b*B[j, :]
        C[j, :] = np.linalg.solve(Ac, MD_c)

        plt.plot(10**(-6)*x, B[j, :], "blue", 10**(-6)*x, C[j, :], "red")
        plt.xlim([0, 3*10**(-3)])
        plt.ylim([0, 10**(17)])
        plt.legend(["bacteria at time t="+str(t[j])])
        plt.pause(0.00001)

    plt.figure()
    plt.plot(10**(-6)*x, B[-1, :], "blue", 10**(-6)*x, 100*C[-1, :], "red", 10**(-6)*x, B[0, :], "yellow")
    plt.xlabel("spatial variable $x$ (meters)")
    plt.ylabel("temporal variables")
    plt.legend(["bacteria", "chemoattractant", "initial bacteria"])

reac_diff_solver_1D()
\end{lstlisting}

\begin{lstlisting}[language=Python, caption=The code exposes the resolution of the 2D system ny a finite difference method.]
import numpy as np
import matplotlib.pyplot as plt
from mpl_toolkits.mplot3d import Axes3D

def reac_diff_solver_2D():
    L = 3000
    N = 100
    T = 20000

    x = np.linspace(0, L, N)
    y = x
    dx = np.max(np.abs(np.diff(x)))
    dt = 1
    R = 0.5
    tempIt = int(T/dt)

    Xi, Yi = np.meshgrid(x, y)

    d_b = 10**0
    d_c = 10**3
    r_c = 0.02
    b_i = 10**17
    r_b = 0.0347
    k = 0.1
    f_b = k*r_c
    s_b = 10**15
    theta = 0.3
    alpha = 0.3129

    f_e = alpha*theta*b_i/((s_b+theta*b_i)*(1-theta))-r_b/k

    Rb = dt/(2*dx**2)*d_b
    Rc = dt/(2*dx**2)*d_c

    b0 = 10**15*np.double(np.abs(Xi-L/2)<50)
    c0 = 0*b0

    X = np.reshape(Xi, (N)**2, 1)
    Y = np.reshape(Yi, (N)**2, 1)

    B_aux = np.array([b0[:,0], b0, b0[:, -1]])
    B_aux = np.vstack([B_aux[0,:], B_aux, B_aux[-1,:]])
    B_vector = np.reshape(B_aux, (N+2)**2, 1)
    C_aux = np.array([c0[:,0], c0, c0[:, -1]])
    C_aux = np.vstack([C_aux[0,:], C_aux, C_aux[-1,:]])
    C_vector = np.reshape(C_aux, (N+2)**2, 1)

    t = 0

    Mb = np.diag((1+2*Rb)*np.ones((N+2)**2,1)) - np.diag(Rb*np.ones((N+2)**2-1,1), -1) - np.diag(Rb*np.ones((N+2)**2-1,1), 1)
    Mb = np.asarray(Mb).tolist()
    Mc = np.diag((1+2*Rc+dt*r_c)*np.ones((N+2)**2,1)) - np.diag(Rc*np.ones((N+2)**2-1,1), -1) - np.diag(Rc*np.ones((N+2)**2-1,1), 1)
    Mc = np.asarray(Mc).tolist()
    Ab = np.diag((1-2*Rb)*np.ones((N+2)**2,1)) + np.diag(Rb*np.ones((N+2)**2-(N+2),1), (N+2)) + np.diag(Rb*np.ones((N+2)**2-(N+2),1), -(N+2))
    Ab = np.asarray(Ab).tolist()
    Ac = np.diag((1-2*Rc)*np.ones((N+2)**2,1)) + np.diag(Rc*np.ones((N+2)**2-(N+2),1), (N+2)) + np.diag(Rc*np.ones((N+2)**2-(N+2),1), -(N+2))
    Ac = np.asarray(Ac).tolist()

    Mbinv = np.linalg.inv(Mb)
    Gb = Mbinv*Ab
    Mcinv = np.linalg.inv(Mc)
    Gc = Mcinv*Ac

    fig = plt.figure(1)
    ax1 = fig.add_subplot(121, projection='3d')
    bsurf = ax1.plot_surface(Xi, Yi, b0)
    ax2 = fig.add_subplot(122, projection='3d')
    csurf = ax2.plot_surface(Xi, Yi, c0)
    ptit = plt.title(f't = {t}')
    plt.xlim([0, L])
    plt.ylim([0, L])

    for it in range(1, tempIt+1):
        t = t+dt
        B_vector = Gb*B_vector
        C_vector = Gc*C_vector

        B_aux = np.reshape(B_vector, (N+2, N+2))
        C_aux = np.reshape(C_vector, (N+2, N+2))

        B_aux[1:N+1, 1:N+1] = B_aux[1:N+1, 1:N+1] + dt*(f_b*C_aux[1:N+1, 1:N+1] - theta*B_aux[1:N+1, 1:N+1] + b_i*(1-theta))
        C_aux[1:N+1, 1:N+1] = C_aux[1:N+1, 1:N+1] + dt*(r_c*C_aux[1:N+1, 1:N+1] + f_e*B_aux[1:N+1, 1:N+1])

        B_vector = np.reshape(B_aux, (N+2)**2, 1)
        C_vector = np.reshape(C_aux, (N+2)**2, 1)

    if it % 50 == 0:
        bsurf.set_zdata(np.reshape(B_vector, (N+2, N+2))[1:N+1, 1:N+1])
        csurf.set_zdata(np.reshape(C_vector, (N+2, N+2))[1:N+1, 1:N+1])
        ptit.set_text(f't = {t}')
        plt.draw()
        plt.pause(0.001)
        plt.show()    
\end{lstlisting}

\begin{lstlisting}[language=Python, caption=The python code contains the PINN implementation for the first Turing system.]
import torch
import numpy as np
import torch.nn as nn
from random import uniform, random
import random
import matplotlib.pyplot as plt
device = torch.device('cuda:0' if torch.cuda.is_available() else 'cpu')
torch.set_default_tensor_type('torch.cuda.FloatTensor')
print(device)

rb = 0.0347
rc = 2
db = 10**(0)
dc = 10**(5)
bi = 10**(7)
k=0.1
fb = k*rc
alpha = 0.3129
sb = 10**(5)
theta = 0.3
fe = alpha*theta*bi/((sb+theta*bi)*(1-theta))-rb/k

N_uv = 14000
N_f =  20000

#Make X_uv_train
#BC x=-10  & tt t > 0
x_left = np.zeros((N_uv//4,1), dtype=float)
t_left = np.random.uniform(low=0.0, high=1500.0, size=(N_uv//4,1))
X_left = np.hstack((x_left, t_left))

#BC x=10 & tt t > 0
x_right = np.ones((N_uv//4,1), dtype=float)*(3000)*10**(-6)
t_right = np.random.uniform(low=0.0, high=1500.0, size=(N_uv//4,1))
X_right = np.hstack((x_right, t_right))

#IC t=0 & tt x,y in [-10,10]
t_zero = np.zeros((N_uv//2,1), dtype=float)
x_zero = np.random.uniform(low=0.0, high=3000*10**(-6), size=(N_uv//2,1))
X_zero = np.hstack((x_zero, t_zero))

X_uv_train = np.vstack((X_left, X_right, X_zero))
# shuffling
index=np.arange(0,N_uv)
np.random.shuffle(index)
X_uv_train=X_uv_train[index,:]

#Make u_train
u_left = np.zeros((N_uv//4,1), dtype=float)
u_right = np.zeros((N_uv//4,1), dtype=float)
#u_initial = list_indicator(np.sort(x_zero, axis=0),1400*10**(-6),1600*10**(-6)).reshape(N_uv//2,1)*10**(5)
#u_initial = np.zeros((N_uv//2,1), dtype=float)
#u_initial[N_uv//4 -50: N_uv//4 +50] = 10**(5)
u_initial = np.exp(-((x_zero*10**(6) - N_uv//4)**2)/10000)*10**(5)
u_train = np.vstack((u_left, u_right, u_initial))


#Make v_train
v_left = np.zeros((N_uv//4,1), dtype=float)
v_right = np.zeros((N_uv//4,1), dtype=float)
v_initial = np.zeros((N_uv//2,1), dtype=float)
v_train = np.vstack((v_left, v_right, v_initial))

# ==========================================
u_train=u_train[index,:]
v_train=v_train[index,:]
# ==========================================
# make X_f_train 
X_f_train=np.zeros((N_f,2),dtype=float)
for row in range(N_f):
    #x=random.randint(0,3000)*10**(-6) 
    x = uniform(0,3000*10**(-6))
    #t=random.randint(0,5000)  
    t = uniform(0,1500)
    X_f_train[row,0]=x
    X_f_train[row,1]=t
   
X_f_train=np.vstack((X_f_train, X_uv_train))

class PhysicsInformedNN():
  def __init__(self,X_uv,u,v,X_f):
    # x & t from boundary conditions
    self.x_uv = torch.tensor(X_uv[:, 0].reshape(-1, 1),dtype=torch.float32,requires_grad=True)
    self.t_uv = torch.tensor(X_uv[:, 1].reshape(-1, 1),dtype=torch.float32,requires_grad=True)
    # x & t from collocation points
    self.x_f = torch.tensor(X_f[:, 0].reshape(-1, 1),dtype=torch.float32,requires_grad=True)
    self.t_f = torch.tensor(X_f[:, 1].reshape(-1, 1),dtype=torch.float32,requires_grad=True)
    # boundary solution
    self.u = torch.tensor(u, dtype=torch.float32)
    self.v = torch.tensor(v, dtype=torch.float32)
    # null vector to test against f:
    self.null =  torch.zeros((self.x_f.shape[0], 1))
    # initialize net:
    self.create_net()

    self.optimizer = torch.optim.Adam(self.net.parameters(),lr=0.001)
    # typical MSE loss (this is a function):
    self.loss = nn.MSELoss()
    # loss :
    self.ls = 0
    # iteration number:
    self.iter = 0

  def create_net(self):
    self.net=nn.Sequential(
        nn.Linear(2,16), nn.Sigmoid(),
        nn.Linear(16, 16), nn.Sigmoid(),
        nn.Linear(16, 16), nn.Sigmoid(),
        nn.Linear(32, 32), nn.Sigmoid(),
        nn.Linear(32, 32), nn.Sigmoid(),
        nn.Linear(32, 32), nn.Sigmoid(),
        nn.Linear(32, 32), nn.Sigmoid(),
        nn.Linear(32, 2))
        
  def net_uv(self,x,t):
    uv=self.net(torch.hstack((x,t)))
    return uv


  def net_fg(self,x,t):

    rb = 0.0347
    rc = 2
    db = 10**(0)
    dc = 10**(5)
    bi = 10**(7)
    k=0.1
    fb = k*rc
    alpha = 0.3129
    sb = 10**(5)
    theta = 0.3
    fe = alpha*theta*bi/((sb+theta*bi)*(1-theta))-rb/k

    uv = self.net_uv(x,t)
    u = uv[:,0].reshape(-1,1).to(device)
    v = uv[:,1].reshape(-1,1).to(device)

    #u partial derivatives
    u_t = torch.autograd.grad(u,t, grad_outputs=torch.ones_like(u), create_graph=True)[0]
    u_x = torch.autograd.grad(u,x, grad_outputs=torch.ones_like(u), create_graph=True)[0]
    u_xx = torch.autograd.grad(u_x,x, grad_outputs=torch.ones_like(u), create_graph=True)[0]
    

    #v partial derivatives
    v_t = torch.autograd.grad(v,t, grad_outputs=torch.ones_like(u), create_graph=True)[0]
    v_x = torch.autograd.grad(v,x, grad_outputs=torch.ones_like(u), create_graph=True)[0]
    v_xx = torch.autograd.grad(v_x,x, grad_outputs=torch.ones_like(u), create_graph=True)[0]
    

    f = u_t - db*u_xx - rb*(1 - u/bi)*u + (alpha*u*v)/(sb + u) - fe*(1 - u/bi)*v
    g = v_t - dc*v_xx - fb*u + rc*v

    f = f.to(device)
    g = g.to(device)

    return f,g

  def closure(self):

    # reset gradients to zero 
    self.optimizer.zero_grad()
    
    # u & f predictions
    uv_prediction = self.net_uv(self.x_uv, self.t_uv)
    u_prediction = uv_prediction[:,0].reshape(-1,1)
    v_prediction = uv_prediction[:,1].reshape(-1,1)
    #
    f_prediction_u, f_prediction_v = self.net_fg(self.x_f, self.t_f)

    #
    # losses:
    u_loss = self.loss(u_prediction, self.u)
    v_loss = self.loss(v_prediction, self.v)
    f_loss_u = self.loss(f_prediction_u, self.null)
    f_loss_v = self.loss(f_prediction_v, self.null)

    self.ls = u_loss + v_loss + f_loss_u + f_loss_v
    
    # derivative with respect to net's weights:
    self.ls.backward()

    # increase iteration count:
    self.iter += 1

    # print report:
    if not self.iter % 100:
      print('Epoch: {0:}, Loss: {1:6.12f}'.format(self.iter, self.ls))
      return self.ls    
        
  def train(self):
    for epoch in range(15000):
      self.net.train()
      self.optimizer.step(self.closure)
\end{lstlisting}

\subsubsection{A second Turing mechanism with cubic term}
Let's discuss another activator-inhibitor dynamic in a Turing system, where the diffusion rate of the activator is much slower than that of the inhibitor. In this system, the inhibitor suppresses the production of both components, while the activator component must increase its own production. The system is subject to small perturbations that encourage the emergence of large-scale patterns, according to the Turing model.

The system is described by two coupled non-linear partial differential equations. The first equation describes the time evolution of the activator component, u, as a function of time t and two spatial dimensions x and y. The left-hand side of the equation represents the rate of change of u with respect to time (irreversible over time), while the right-hand side consists of four terms. The first term represents the diffusion of u with a diffusion coefficient a, while the second term represents the self-enhancement of u (a non-linear term with a cubic dependency). The third term represents the inhibition of u by v, and the fourth term is a constant offset, c. The second equation describes the time evolution of the inhibitor component, v, with an additional time constant, $\tau$. The left-hand side of the equation represents the rate of change of v with respect to time, while the right-hand side consists of three terms. The first term represents the diffusion of v with a diffusion coefficient b, while the second term represents the production of v by u. The third term represents the inhibition of v by itself. The coupling of the two equations via the inhibition term of the activator and the production term of the inhibitor results in the formation of spatial patterns in the system.
$$\frac{\partial u(t,x,y)}{\partial t} = a * \Delta u(t,x,y) + u(t,x,y) - u^3(t,x,y) - v(t,x,y) + c$$
$$ \tau \frac{\partial v(t,x,y)}{\partial t} = b * \Delta v(t,x,y) + u(t,x,y) - v(t,x,y)$$

\begin{lstlisting}[language=Python, caption=The resolution of the system of equations by a finite difference method. The code is modified from \citep{rossant2018ipython}.]
import numpy as np
import matplotlib.pyplot as plt

# Constants
a = 2.8e-4
b = 5e-3
tau = .1
k = -.005

# Grid size and step sizes
size = 100
dx = 2. / size

# Total time and number of iterations
T = 15.0
dt = .001
n = int(T / dt)

# Initialize U and V as random arrays
U = np.random.rand(size, size)
V = np.random.rand(size, size)

# Function to compute the Laplacian of an array
def laplacian(Z):
    Ztop = Z[0:-2, 1:-1]
    Zleft = Z[1:-1, 0:-2]
    Zbottom = Z[2:, 1:-1]
    Zright = Z[1:-1, 2:]
    Zcenter = Z[1:-1, 1:-1]
    return (Ztop + Zleft + Zbottom + Zright - 4 * Zcenter) / dx**2

# Function to plot the patterns
def show_patterns(U, ax=None):
    ax.imshow(U, cmap=plt.cm.Reds,
              interpolation='bilinear',
              extent=[-1, 1, -1, 1])
    ax.set_axis_off()
    
# Set up the plot
fig, axes = plt.subplots(3, 3, figsize=(8, 8))
step_plot = n // 15

# Iterate and update the variables
for i in range(n):
    # Compute the Laplacians of U and V
    deltaU = laplacian(U)
    deltaV = laplacian(V)
    # Take the values of U and V inside the grid
    Uc = U[1:-1, 1:-1]
    Vc = V[1:-1, 1:-1]
    # Update U and V
    U[1:-1, 1:-1], V[1:-1, 1:-1] = (
        Uc + dt * (a * deltaU + Uc - Uc**3 - Vc + k),
        Vc + dt * (b * deltaV + Uc - Vc) / tau
    )
    # Neumann conditions: set derivatives at the edges to zero
    for Z in (U, V):
        Z[0, :] = Z[1, :]
        Z[-1, :] = Z[-2, :]
        Z[:, 0] = Z[:, 1]
        Z[:, -1] = Z[:, -2]

    # Plot the state of the system at 15 different times
    if i % step_plot == 0 and i < 15 * step_plot:
        ax = axes.flat[i // step_plot]
        show_patterns(U, ax=ax)
        ax.set_title(f'$t={i * dt:.2f}$')
\end{lstlisting}

\begin{figure}[H]
   \centering
    \includegraphics[scale=1]{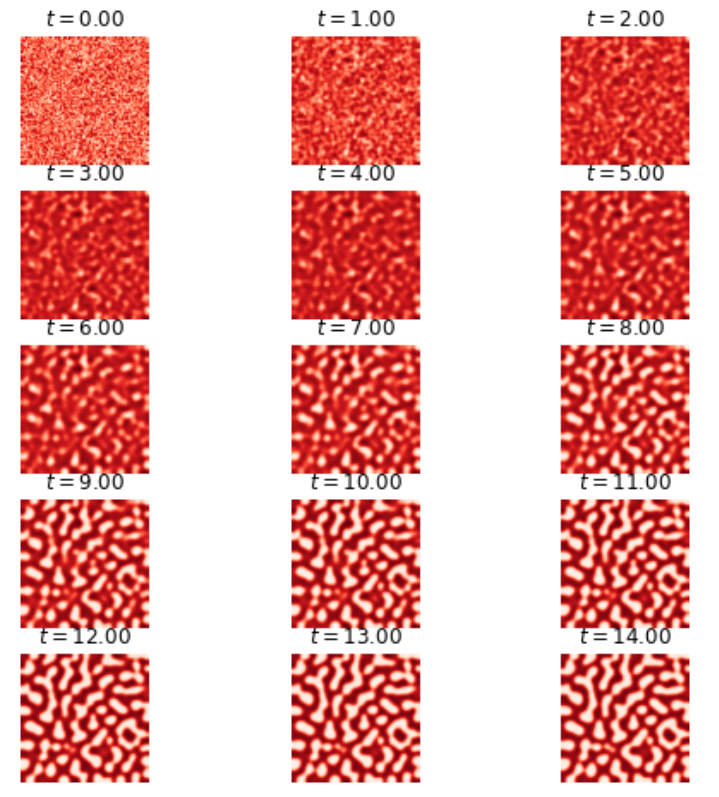}
    \caption{The Crohn's disease structure formation with a finite difference method. In this simulation, the Neumann boundary conditions have been considered (The solution normal derivative values at the domain boundaries are fixed to zero). The time intervals used do not necessarily correspond to the real elapsed time in the real case, and are only used to describe the dynamics of the systems.}
   \label{fig:evolution}
\end{figure}

\begin{figure}[H]
\begin{minipage}[h]{0.47\linewidth}
\begin{center}
\includegraphics[width=1\linewidth]{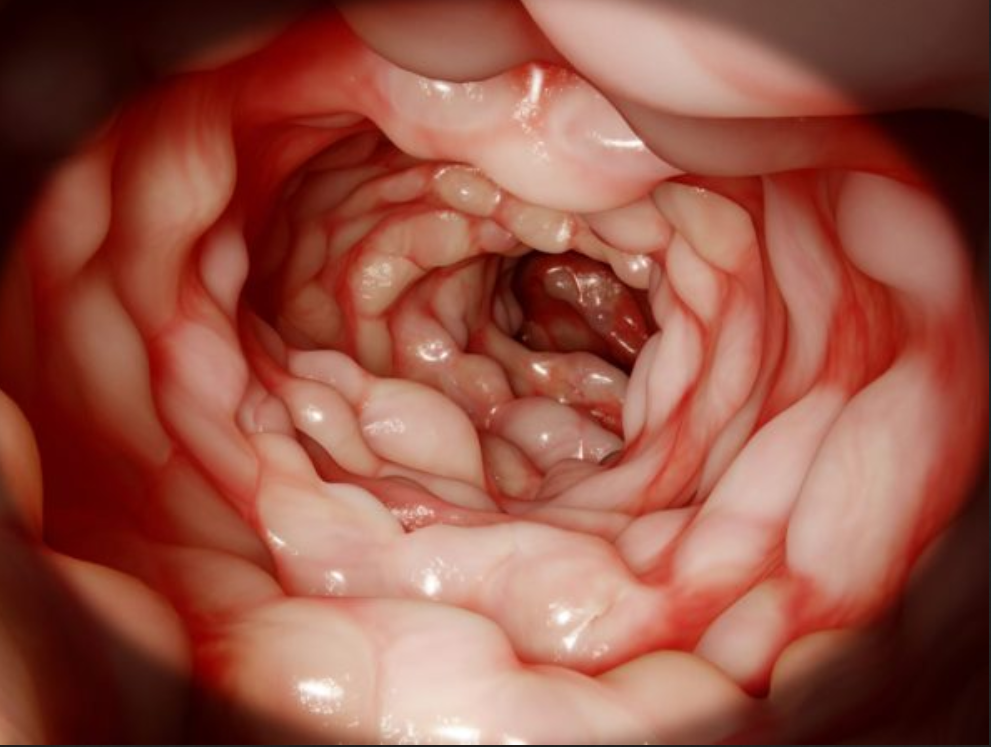} 
\label{qwe1}
\end{center} 
\end{minipage}
\hfill
\vspace{0.2 cm}
\begin{minipage}[h]{0.47\linewidth}
\begin{center}
\includegraphics[width=1\linewidth]{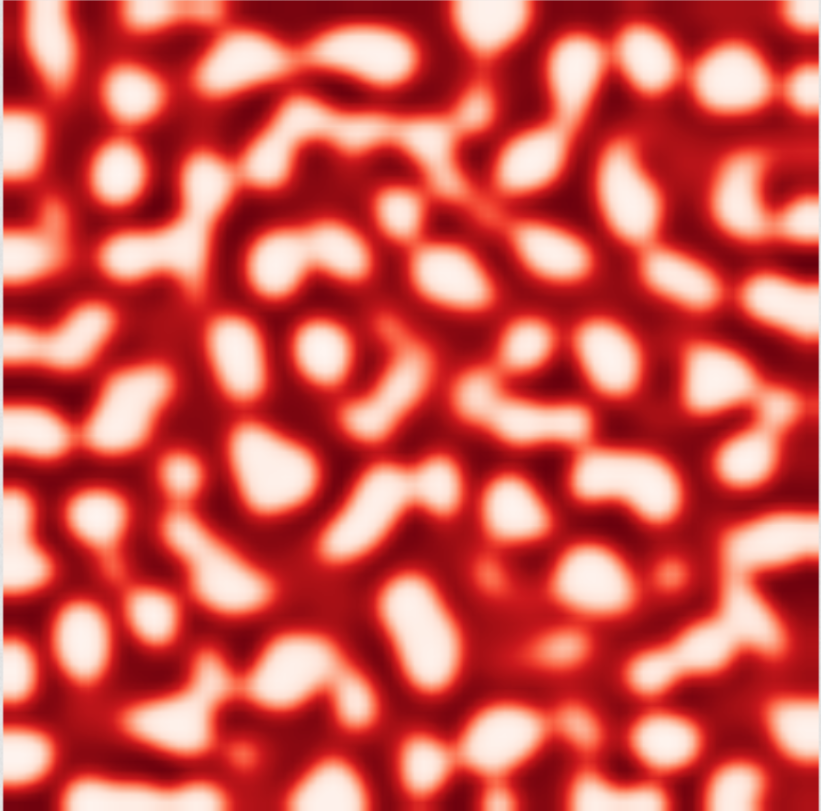} 
\label{qwe1}
\end{center}
\end{minipage}
\caption{The two figures show a strong similarity between the real disease pattern (on the left) and the simulated pattern (on the right). This suggests that the Turing nonlinear system of partial differential equations, solved using the finite difference schema, was able to accurately capture the behavior of the real disease (the underlying dynamics including the interactions between different factors such as the concentration of bacteria and the presence of phagocytes and the spread of the disease in this case).}
\label{fig:example}
\end{figure}

\begin{lstlisting}[language=Python, caption=The code used implements the PINN architecture aiming to solve the second Turing system.]
import torch
import numpy as np
import torch.nn as nn
from random import uniform
import matplotlib.pyplot as plt
from mpl_toolkits.axes_grid1 import make_axes_locatable
device = torch.device('cuda:0' if torch.cuda.is_available() else 'cpu')
torch.set_default_tensor_type('torch.cuda.FloatTensor')
print(device)
#
N_uv = 400
N_f = 10000

#Make X_uv_train
#BC x=-1, tt y, t>0
x_left = np.ones((N_uv//8,1), dtype=float)*(-1)
y_left = np.random.uniform(low=-1.0, high=1.0, size=(N_uv//8,1))
t_left = np.random.uniform(low=0.0001, high=10.0, size=(N_uv//8,1))
X_left = np.hstack((x_left, y_left, t_left))

#BC x=1, tt y, t>0
x_right = np.ones((N_uv//8,1), dtype=float)*(1)
y_right = np.random.uniform(low=-1.0, high=1.0, size=(N_uv//8,1))
t_right = np.random.uniform(low=0.0001, high=10.0, size=(N_uv//8,1))
X_right = np.hstack((x_right, y_right, t_right))

#BC y=1, tt x, t>0
x_upper = np.random.uniform(low=-1.0, high=1.0, size=(N_uv//8,1))
y_upper = np.ones((N_uv//8, 1), dtype=float)*(1)
t_upper = np.random.uniform(low=0.0001, high=10.0, size=(N_uv//8,1))
X_upper = np.hstack((x_upper, y_upper, t_upper))

#BC y=-1, tt x, t>0
x_lower = np.random.uniform(low=-1.0, high=1.0, size=(N_uv//8,1))
y_lower = np.ones((N_uv//8,1), dtype=float)*(-1)
t_lower = np.random.uniform(low=0.0001, high=10.0, size=(N_uv//8,1))
X_lower = np.hstack((x_lower, y_lower, t_lower))

#IC t=0, tt x,y
x_zero = np.random.uniform(low=-1.0, high=1.0, size=(N_uv//2,1))
y_zero = np.random.uniform(low=-1.0, high=1.0, size=(N_uv//2,1))
t_zero = np.zeros((N_uv//2,1), dtype=float)
X_zero = np.hstack((x_zero, y_zero, t_zero))

X_uv_train = np.vstack((X_left, X_right, X_upper, X_lower, X_zero))

#Shuffling
index = np.arange(0,N_uv)
np.random.shuffle(index)
X_uv_train = X_uv_train[index,:]

#Make u_train
u_left = np.zeros((N_uv//8,1), dtype=float)
u_right = np.zeros((N_uv//8,1), dtype=float)
u_upper = np.zeros((N_uv//8,1), dtype=float)
u_lower = np.zeros((N_uv//8,1), dtype=float)
u_initial = np.random.uniform(low=0.0, high=1.0, size=(N_uv//2,1))
#u_initial = np.exp(-x_zero**2 - y_zero**2)
u_train = np.vstack((u_left, u_right, u_upper, u_lower, u_initial))

#Make v_train
v_left = np.zeros((N_uv//8,1), dtype=float)
v_right = np.zeros((N_uv//8,1), dtype=float)
v_upper = np.zeros((N_uv//8,1), dtype=float)
v_lower = np.zeros((N_uv//8,1), dtype=float)
v_initial = np.random.uniform(low=0.0, high=1.0, size=(N_uv//2,1))
#v_initial = np.exp(-x_zero**2 - y_zero**2)
v_train = np.vstack((v_left, v_right, v_upper, v_lower, v_initial))

#shuffling
u_train = u_train[index,:]
v_train = v_train[index,:]

#Make X_f_train
X_f_train = np.zeros((N_f,3), dtype=float)
for i in range(N_f):
  x=uniform(-1,1)
  y=uniform(-1,1)
  t=uniform(0,10)
  X_f_train[i,0] = x
  X_f_train[i,1] = y
  X_f_train[i,2] = t

X_f_train = np.vstack((X_f_train, X_uv_train))
#
class PhysicsInformedNN():
  def __init__(self,X_uv,u,v,X_f):
    # x & t from boundary conditions
    self.x_uv = torch.tensor(X_uv[:, 0].reshape(-1, 1),dtype=torch.float32,requires_grad=True)
    self.y_uv = torch.tensor(X_uv[:, 1].reshape(-1, 1),dtype=torch.float32,requires_grad=True)
    self.t_uv = torch.tensor(X_uv[:, 2].reshape(-1, 1),dtype=torch.float32,requires_grad=True)
    # x & t from collocation points
    self.x_f = torch.tensor(X_f[:, 0].reshape(-1, 1),dtype=torch.float32,requires_grad=True)
    self.y_f = torch.tensor(X_f[:, 1].reshape(-1, 1),dtype=torch.float32,requires_grad=True)
    self.t_f = torch.tensor(X_f[:, 2].reshape(-1, 1),dtype=torch.float32,requires_grad=True)
    # boundary solution
    self.u = torch.tensor(u, dtype=torch.float32)
    self.v = torch.tensor(v, dtype=torch.float32)
    # null vector to test against f:
    self.null =  torch.zeros((self.x_f.shape[0], 1))
    # initialize net:
    self.create_net()

    self.optimizer = torch.optim.Adam(self.net.parameters(),lr=0.001)
    # typical MSE loss (this is a function):
    self.loss = nn.MSELoss()
    # loss :
    self.ls = 0
    # iteration number:
    self.iter = 0

  def create_net(self):
    self.net=nn.Sequential(
        nn.Linear(3,16), nn.Sigmoid(),
        nn.Linear(16, 16), nn.Sigmoid(),
        nn.Linear(16, 16), nn.Sigmoid(),
        nn.Linear(16, 16), nn.Sigmoid(),
        nn.Linear(20, 20), nn.Sigmoid(),
        nn.Linear(32, 32), nn.Sigmoid(),
        nn.Linear(32, 32), nn.Sigmoid(),
        nn.Linear(32, 2))
    
  def net_uv(self,x,y,t):
    uv=self.net(torch.hstack((x,y,t)))
    return uv


  def net_fg(self,x,y,t):

    a = 2.8e-4
    b = 5e-3
    c = -0.005
    tau=0.1

    uv = self.net_uv(x,y,t)
    u = uv[:,0].reshape(-1,1).to(device)
    v = uv[:,1].reshape(-1,1).to(device)

    
    #u partial derivatives
    u_t = torch.autograd.grad(u,t, grad_outputs=torch.ones_like(u), create_graph=True)[0]
    u_x = torch.autograd.grad(u,x, grad_outputs=torch.ones_like(u), create_graph=True)[0]
    u_xx = torch.autograd.grad(u_x,x, grad_outputs=torch.ones_like(u), create_graph=True)[0]
    u_y = torch.autograd.grad(u,y, grad_outputs=torch.ones_like(u), create_graph=True)[0]
    u_yy = torch.autograd.grad(u_y,y, grad_outputs=torch.ones_like(u), create_graph=True)[0]
  

    #v partial derivatives
    v_t = torch.autograd.grad(v,t, grad_outputs=torch.ones_like(u), create_graph=True)[0]
    v_x = torch.autograd.grad(v,x, grad_outputs=torch.ones_like(u), create_graph=True)[0]
    v_xx = torch.autograd.grad(v_x,x, grad_outputs=torch.ones_like(u), create_graph=True)[0]
    v_y = torch.autograd.grad(v,y, grad_outputs=torch.ones_like(u), create_graph=True)[0]
    v_yy = torch.autograd.grad(v_y,y, grad_outputs=torch.ones_like(u), create_graph=True)[0]
  

    f = u_t - a*(u_xx + u_yy) - u + u**3 + v - c
    g = v_t - (b*(v_xx + v_yy) + u - v)/tau

    f = f.to(device)
    g = g.to(device)

    return f,g

  def closure(self):

    # reset gradients to zero 
    self.optimizer.zero_grad()
    
    # u & f predictions
    uv_prediction = self.net_uv(self.x_uv, self.y_uv, self.t_uv)
    u_prediction = uv_prediction[:,0].reshape(-1,1)
    v_prediction = uv_prediction[:,1].reshape(-1,1)
    #
    f_prediction_u, f_prediction_v = self.net_fg(self.x_f, self.y_f, self.t_f)

    #
    # losses:
    u_loss = self.loss(u_prediction, self.u)
    v_loss = self.loss(v_prediction, self.v)
    f_loss_u = self.loss(f_prediction_u, self.null)
    f_loss_v = self.loss(f_prediction_v, self.null)

    self.ls = u_loss + v_loss + f_loss_u + f_loss_v
    
    # derivative with respect to net's weights:
    self.ls.backward()

    # increase iteration count:
    self.iter += 1

    # print report:
    if not self.iter % 100:
      print('Epoch: {0:}, Loss: {1:6.8f}'.format(self.iter, self.ls))
      return self.ls    
        
  def train(self):
    for epoch in range(6000):
      self.net.train()
      self.optimizer.step(self.closure)

pinn = PhysicsInformedNN(X_uv_train,u_train,v_train, X_f_train)
pinn.train()
#     
x = torch.linspace(-1, 1, 200)
y = torch.linspace(-1, 1, 200)
t = torch.linspace( 0, 10, 200)
# x & t grids:
X,Y,T = torch.meshgrid(x,y,t)
# x & t columns:
xcol = X.reshape(-1, 1)
ycol = Y.reshape(-1, 1)
tcol = T.reshape(-1, 1)
# one large column:
sol = pinn.net_uv(xcol, ycol, tcol)
usol = sol[:,0].reshape(-1,1)
vsol = sol[:,1].reshape(-1,1)
# reshape solution:
U = usol.reshape(x.numel(), y.numel(), t.numel())
V = vsol.reshape(x.numel(), y.numel(), t.numel())
# transform to numpy:
xnp = x.cpu().numpy()
ynp = y.cpu().numpy()
tnp = t.cpu().numpy()
Unp = U.cpu().detach().numpy()
Vnp = V.cpu().detach().numpy()

# Plot a heat map for the final result
#plt.imshow(Unp[:,:,-1])  
#plt.imshow(Vnp[:,:,-1])

\end{lstlisting}
\begin{table}[H]
\centering
\begin{tabular}{lc|clllll|}
\cline{3-8}
\multicolumn{2}{l|}{}            & \multicolumn{6}{c|}{Finite difference solution}                                                                                                                      \\ \cline{3-8} 
\multicolumn{2}{l|}{}                             & \multicolumn{6}{c|}{Neurons}                                                                                                                                           \\ \cline{3-8} 
\multicolumn{2}{l|}{}                             & \multicolumn{2}{c|}{8}                                  & \multicolumn{2}{c|}{16}                                 & \multicolumn{2}{c|}{32}                             \\ \cline{3-8} 
\multicolumn{2}{l|}{}                             & \multicolumn{1}{c|}{U}     & \multicolumn{1}{c|}{V}     & \multicolumn{1}{c|}{U}     & \multicolumn{1}{c|}{V}     & \multicolumn{1}{c|}{U}     & \multicolumn{1}{c|}{V} \\ \hline
\multicolumn{1}{|c|}{Layers} & 2 & \multicolumn{1}{l|}{0.987} & \multicolumn{1}{l|}{1.233} & \multicolumn{1}{l|}{0.994} & \multicolumn{1}{l|}{1.165} & \multicolumn{1}{l|}{0.857} & 1.137                  \\ \cline{2-8} 
\multicolumn{1}{|c|}{}                        & 4 & \multicolumn{1}{l|}{0.993} & \multicolumn{1}{l|}{1.195} & \multicolumn{1}{l|}{0.993} & \multicolumn{1}{l|}{1.092} & \multicolumn{1}{l|}{1.115} & 1.238                  \\ \cline{2-8} 
\multicolumn{1}{|c|}{}                        & 8 & \multicolumn{1}{l|}{0.989} & \multicolumn{1}{l|}{1.132} & \multicolumn{1}{l|}{1.003} & \multicolumn{1}{l|}{1.120} & \multicolumn{1}{l|}{1.238} & 1.125                  \\ \hline
\end{tabular}
\caption{\label{tab:turing_parameters_rmse}RSME between the PINN and the finite difference solution for the Turing reaction-diffusion system.}
\end{table}
\begin{figure}[H]
   \centering
   \includegraphics[scale=0.55]{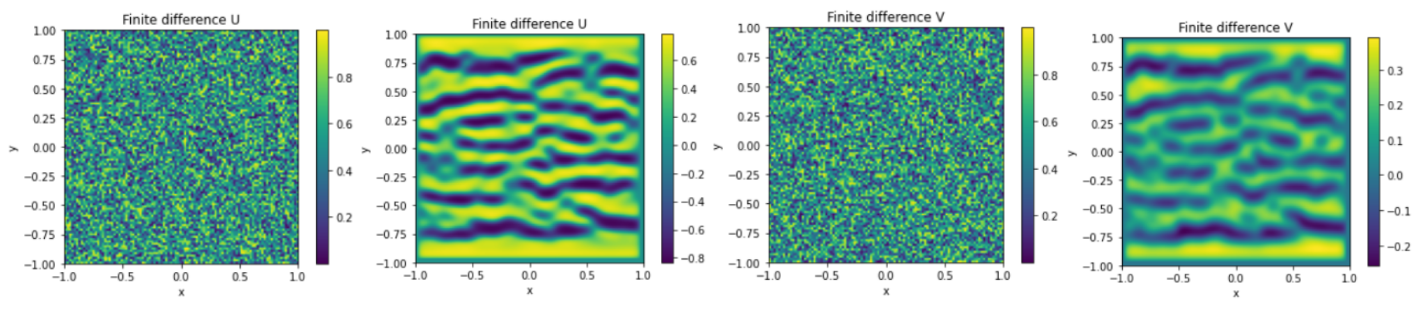}
   \caption{Solution using finite difference method of the Turing system at t=0 and t=10.}
   \label{fig:turing_pinn}
\end{figure}
\begin{figure}[H]
   \centering
   \includegraphics[scale=0.7]{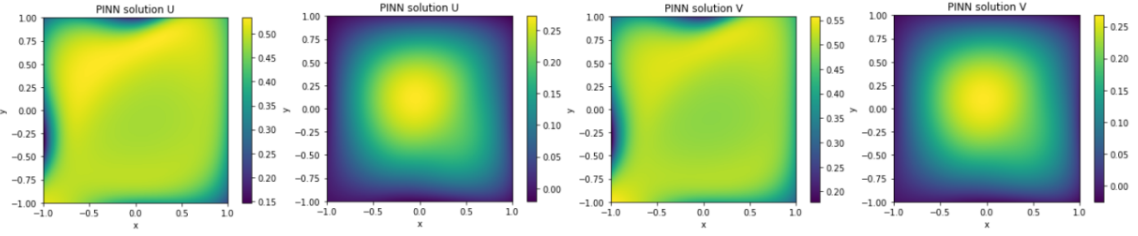}
   \caption{The figure shows that the PINN architecture fail to model the complex dynamic of the Crohn's disease.}
   \label{fig:second_turing_pinn_failing_PINN}
\end{figure}
The root mean squared error (RMSE) between the solution obtained by using the physics-informed neural network (PINN) and the finite difference solution tends towards 1 for u and v. At first glance, this low RMSE might indicate that the neural network has converged to a solution. However, as evidenced in figure \ref{fig:second_turing_pinn_failing_PINN}, this is not the case in practice. This equation models the formation of patterns as a result of interaction between different pigments, but the two solutions do not show the same patterns. Therefore, in this case, the PINN diverges and fails to approximate the numerical solution. Continued research aims to uncover the root causes behind the inability of PINNs to adequately capture the intricate explosion phenomena in IBDs.

\subsection{Fisher-KPP equation: modeling the bacteria/phagocytes couple with one PDE}
The Fisher \citep{fisher} Kolmogorov-Petrovsky-Piskunov \citep{kolmogorov} equation is a reaction-diffusion equation that plays a significant role in describing various chemical, physical and biological phenomena. It is often used to model the propagation of a single wave, such as the spread of a disease or the front of a chemical reaction. In the case of ulcerative colitis and Crohn's disease, the Fisher KPP equation can be used to model the spread of these intestinal diseases through the incorporation of factors such as the concentration of bacteria and the presence of phagocytes. One advantage of using the Fisher-KPP equation to model the IBDSs spread is that it is a relatively simple equation that can be solved analytically or numerically. This makes it a good choice for modeling the spread of these diseases, especially in cases where there is limited data available \citep{Lagergren-2020}.

\begin{equation*}
\begin{cases}
u(x,t) = \frac{bacteria(x,t)}{bacteria(x,t) +phagocytes(x,t) } \\
bacteria(x,t) +phagocytes(x,t) = N
\end{cases}
\end{equation*}

\subsection*{Fisher-KPP Equation}
$$ \frac{\partial{ u(x,t)}}{\partial t} = D\frac{\partial^2{u(x,t)}}{\partial{x^2}} + ru(x,t)(1-u(x,t)), \hspace{0.5cm} x \in [-50,50] \hspace{0.2cm} t \in [0,10], \hspace{0.2cm} $$

\begin{equation*}
u(x,0) =
\begin{cases}
1 & \text{if } x \leq 0 \\
0 & \text{if } x > 0
\end{cases}
\end{equation*}

And the boundary conditions u(-50,t)=1 and u(50,t)=0 with D=1 and r=1. The diffusion term: $D\frac{\partial^2{u(x,t)}}{\partial{x^2}}$ represents the rate at which bacteria in the medium migrate through a linear diffusion process with diffusivity D. The reaction term: $ru(x,t)(1-u(x,t))$ represents the bacteria proliferation rate in the medium, which is assumed to be proportional to the $u(x,t)$, and the remaining carrying capacity of the environment, $(1-u(x,t))$. The parameter r represents the growth rate and the quantity $ru(x,t)(1-u(x,t))$ models the bacteria logistic growth. The term $(1-u(x,t))$ represents the limiting factor, which means that the growth rate decreases as both bacteria and phagocytes approaches the carrying capacity of the biological medium.

\subsection*{Numerical schema}
The solution domain is discretized into cells described by the node set $(x_{i}, t_{n})$ in which $x_{j}=ih$, $t_{n}=nk$ (i = 0,1,...,I; n = 0,1,...,N) $h=\Delta x$  is a spatial mesh size, $k=\Delta t$ is the time step and $u(x_{i}, t_{n})=u_{i}^{n}$

$$u_{i}^{n+1} = u_{i}^{n} + D \frac{\Delta t}{(\Delta x)^{2}}(u_{i+1}^{n} - 2u_{i}^{n} + u_{i-1}^{n}) + r \Delta t u_{i}^{n}(1-u_{i}^{n})$$

\begin{lstlisting}[language=Python, caption=Implementation of the Fisher-KPP resolution with a finite difference schema.]
import numpy as np
from tqdm import tqdm
import matplotlib.pyplot as plt
from mpl_toolkits.mplot3d import Axes3D
from matplotlib import cm
from matplotlib.ticker import LinearLocator, FormatStrFormatter
from mpl_toolkits.axes_grid1 import make_axes_locatable

D=1
r=1
delta_x = 0.1
delta_t = 0.001
alpha = D*delta_t/(delta_x**2)
beta = r*delta_t
x = np.arange(-50,50,delta_x)
t = np.arange(0,1,delta_t)
m=len(x)
n=len(t)
u_df = np.zeros((len(t), len(x)))
plt.plot(x, np.heaviside(-x,0))
#Initial condition
u_df[0,:] = np.heaviside(-x,0)

#Bondaries conditions
u_df[:,0] = 1
u_df[:,-1] = 0

#Filling u matrix for finite difference schema
for k in tqdm(range(0,n-1)):
  for i in range(1,m-2):
    u_df[k+1,i] = u_df[k,i] + alpha*(u_df[k,i+1] - 2*u_df[k,i] + u_df[k,i-1]) + beta*u_df[k,i]*(1-u_df[k,i])

plt.plot(x, u_df[100,:])

#Plotting u for finite difference
for i in range(0,1000,200):
  plt.plot(x, u_df[i,:], label='t='+str(i*0.001)+'s')
  plt.xlim(left=-10, right=10)
  plt.legend()
  plt.xlabel('x')
  plt.ylabel('u(x,t)')


# 3D plotting in the phase space
fig = plt.figure()
ax = fig.gca(projection='3d')

x = np.arange(-50,50, delta_x)
t = np.arange(0,1, delta_t)

ms_x, ms_t = np.meshgrid(x,t)


surf = ax.plot_surface(ms_x, ms_t, u_df , cmap=cm.coolwarm, linewidth=0, antialiased=False)

ax.zaxis.set_major_locator(LinearLocator(10))
ax.zaxis.set_major_formatter(FormatStrFormatter('%.02f'))

ax.set_xlabel('x')
ax.set_ylabel('t')
#ax.set_zlabel('u(x,t)')

fig.colorbar(surf, shrink=0.5, aspect=5)

plt.show()
\end{lstlisting}

\begin{figure}[H]
   \centering
   \includegraphics[scale=0.9]{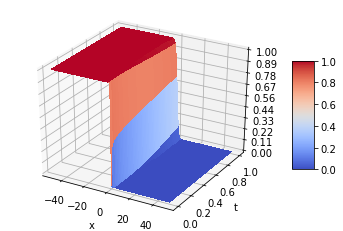}
   \caption{Solution of the Fisher-KPP equation using a finite difference method in the phase space.}
   \label{fig:fisherkpp3dfinitedifference}
\end{figure}

\subsection*{Comparison}
\begin{figure}[htp]
\centering
\includegraphics[width=.5\textwidth]{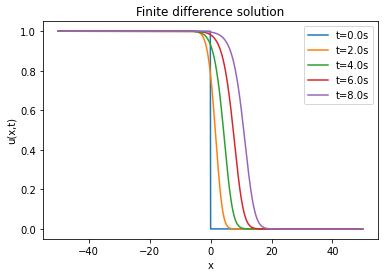}\hfill
\includegraphics[width=.5\textwidth]{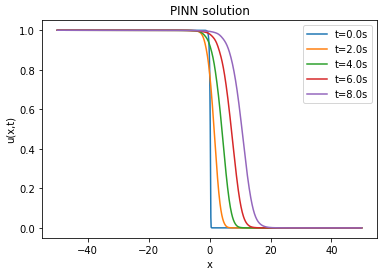}
\caption{PINN, analytical and finite difference solution side by side}
\label{fig:figure3}
\end{figure}

\begin{figure}[htp]
\centering
\includegraphics[width=.5\textwidth]{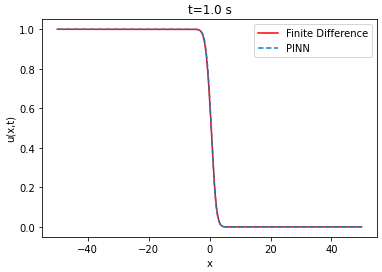}\hfill
\includegraphics[width=.5\textwidth]{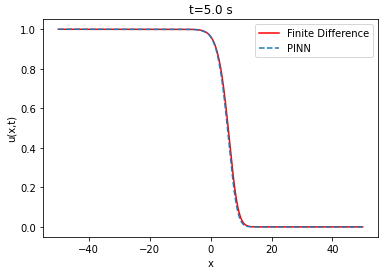}\hfill
\includegraphics[width=.5\textwidth]{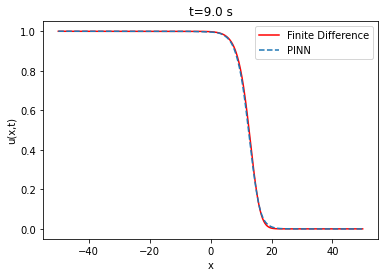}

\caption{PINN and finite difference solution at t=1s, t=5s and t=9s}
\label{fig:figure3}

\end{figure}

\begin{table}[H]
\centering
\begin{tabular}{ll|lll|}
\cline{3-5}
                                              &   & \multicolumn{3}{l|}{Analytical solution}                                 \\ \cline{3-5} 
                                              &   & \multicolumn{1}{l|}{8}        & \multicolumn{1}{l|}{16}       & 32       \\ \hline
{Layers} & 2 & \multicolumn{1}{l|}{0.028321} & \multicolumn{1}{l|}{0.074042} & 0.040836 \\ \cline{2-5} 
\multicolumn{1}{|l|}{}                        & 4 & \multicolumn{1}{l|}{0.037587} & \multicolumn{1}{l|}{0.009196} & 0.027296 \\ \cline{2-5} 
\multicolumn{1}{|l|}{}                        & 8 & \multicolumn{1}{l|}{0.043136} & \multicolumn{1}{l|}{0.022766} & 0.013606 \\ \hline
\end{tabular}
\caption{\label{tab:Time 1}RSME between PINN and analytical solution for Fisher-KPP equation}
\end{table}

\begin{table}[H]
\centering
\begin{tabular}{ll|lll|}
\cline{3-5}
                                              &   & \multicolumn{3}{l|}{Neurons}                                 \\ \cline{3-5} 
                                              &   & \multicolumn{1}{l|}{8}    & \multicolumn{1}{l|}{16}   & 32   \\ \hline
{Layers} & 2 & \multicolumn{1}{l|}{1:39} & \multicolumn{1}{l|}{1:43} & 2:06 \\ \cline{2-5} 
\multicolumn{1}{|l|}{}                        & 4 & \multicolumn{1}{l|}{2:41} & \multicolumn{1}{l|}{2:39} & 4:02 \\ \cline{2-5} 
\multicolumn{1}{|l|}{}                        & 8 & \multicolumn{1}{l|}{4:24} & \multicolumn{1}{l|}{4:48} & 8:00 \\ \hline
\end{tabular}
\caption{\label{tab:Time 1}Execution time according to the number of layers and neurons for the Fisher-KPP equation}
\end{table}

The lower the RMSE, the better the approximation. In our case, the RMSE in relation to the number of layers and neurons is on average of the order of $10^{-2}$. Thus, the PINN is able to approximate the finite difference solution of the Fisher-KPP equation. Note that increasing or decreasing the number of layers and neurons does not systematically improve the RMSE.

\begin{figure}[H]
   \centering
   \includegraphics[scale=0.45]{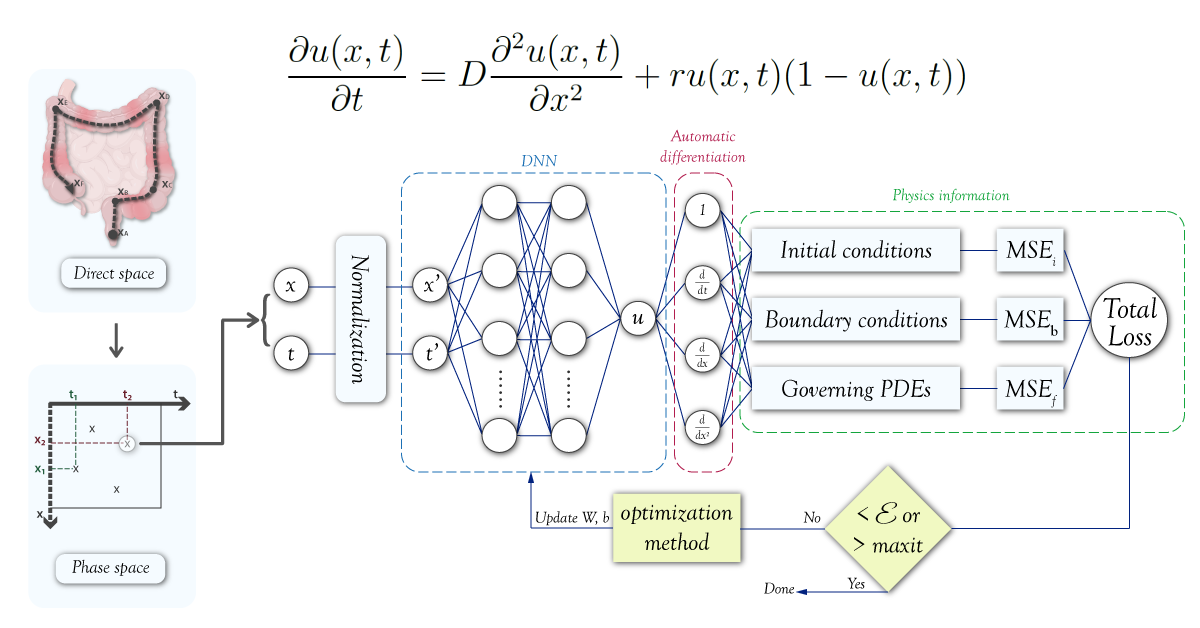}
   \caption{The figure shows the general architecture of the resolution with PINN for the Fisher-KPP equation.}
   \label{fig:general_architecture_pinn}
\end{figure}

\begin{lstlisting}[language=Python, caption=The PINN architecture code for the Fisher-KPP equation.]
import torch
import torch.nn as nn
from random import uniform
device = torch.device('cuda:0' if torch.cuda.is_available() else 'cpu')
torch.set_default_tensor_type('torch.cuda.FloatTensor')
N_u =2000
N_f =15000

#Make X_u_train
#Boundary condition: x=-10 , tt t>0
x_left = np.ones((N_u//4,1))*(-10)
t_left = np.random.uniform(low=0, high=1, size=(N_u//4,1))
X_left = np.hstack((x_left, t_left))

#Boundary condition: x=10 , tt t>0
x_right = np.ones((N_u//4,1))*(10)
t_right = np.random.uniform(low=0, high=1, size=(N_u//4,1))
X_right = np.hstack((x_right, t_right))

#Initial condition: t=0 , tt x in [-10,10]
x_zero = np.random.uniform(low=-10, high=10, size=(N_u//2,1))
t_zero = np.zeros((N_u//2,1))
X_zero = np.hstack((x_zero, t_zero))

X_u_train=np.vstack((X_right,X_left,X_zero))
# shuffling
index=np.arange(0,N_u)
np.random.shuffle(index)
X_u_train=X_u_train[index,:] 

#Make u_train
# make u_train
u_right=np.zeros((N_u//4,1),dtype=float) # boundary condition x=-10
u_left=np.ones((N_u//4,1),dtype=float)  # boundary condition x=10
u_initial=np.heaviside(-x_zero,0) # initial condition about the function 
u_train=np.vstack((u_right,u_left,u_initial)) # u_train tensor preparation 
u_train=u_train[index,:]

# make X_f_train 
X_f_train=np.zeros((N_f,2),dtype=float)
for row in range(N_f):
  x=uniform(-10,10) 
  t=uniform(0,1)  
  X_f_train[row,0]=x
  X_f_train[row,1]=t
X_f_train=np.vstack((X_f_train, X_u_train)) 

class PhysicsInformedNN():
  def __init__(self,X_u,u,X_f):
    # x & t from boundary conditions
    self.x_u = torch.tensor(X_u[:, 0].reshape(-1, 1),dtype=torch.float32,requires_grad=True)
    self.t_u = torch.tensor(X_u[:, 1].reshape(-1, 1),dtype=torch.float32,requires_grad=True)
    # x & t from collocation points
    self.x_f = torch.tensor(X_f[:, 0].reshape(-1, 1),dtype=torch.float32,requires_grad=True)
    self.t_f = torch.tensor(X_f[:, 1].reshape(-1, 1),dtype=torch.float32,requires_grad=True)
    # boundary solution
    self.u = torch.tensor(u, dtype=torch.float32)
    # null vector to test against f:
    self.null =  torch.zeros((self.x_f.shape[0], 1))
    # initialize net:
    self.create_net()

    self.optimizer = torch.optim.Adam(self.net.parameters(),lr=0.001)
    # typical MSE loss (this is a function):
    self.loss = nn.MSELoss()
    # loss :
    self.ls = 0
    # iteration number:
    self.iter = 0

  def create_net(self):
    self.net=nn.Sequential(
        nn.Linear(2,32), nn.Sigmoid(),
       
        nn.Linear(32, 32), nn.Sigmoid(),
        
        nn.Linear(32, 32), nn.Sigmoid(),
        nn.Linear(32, 32), nn.Sigmoid(),
        nn.Linear(32, 1))
  def net_u(self,x,t):
    u=self.net(torch.hstack((x,t)))
    return u
  def net_f(self,x,t):
    u=self.net_u(x,t)
    u = u.to(device)

    u_t=torch.autograd.grad(u,t,grad_outputs=torch.ones_like(u),create_graph=True)[0]
    u_x=torch.autograd.grad(u,x,grad_outputs=torch.ones_like(u),create_graph=True)[0]
    u_xx=torch.autograd.grad(u_x,x,grad_outputs=torch.ones_like(u),create_graph=True)[0]
    u_t = u_t.to(device)
    u_x = u_x.to(device)
    u_xx = u_xx.to(device)
    

    f = u_t - u_xx - u*(1-u)
    f = f.to(device)

    return f

  def closure(self):
    # reset gradients to zero 
    self.optimizer.zero_grad()
    # u & f predictions
    u_prediction = self.net_u(self.x_u, self.t_u)
    #
    f_prediction = self.net_f(self.x_f,self.t_f)
    #
    # losses:
    u_loss_x = self.loss(u_prediction, self.u)
    f_loss = self.loss(f_prediction, self.null)
    self.ls = u_loss_x + f_loss
    # derivative with respect to net's weights:
    self.ls.backward()
    # increase iteration count:
    self.iter += 1
    # print report:
    if not self.iter % 100:
      print('Epoch: {0:}, Loss: {1:6.8f}'.format(self.iter, self.ls))
    return self.ls    
    
  def train(self):
    """
    training loop
    """
    for epoch in range(20000):
      self.net.train()
      self.optimizer.step(self.closure)
     
# pass data sets to the PINN:
pinn = PhysicsInformedNN(X_u_train, u_train, X_f_train)
pinn.train()

import matplotlib.pyplot as plt
from mpl_toolkits.axes_grid1 import make_axes_locatable
#     
x = torch.linspace(-50, 50, 1000)
t = torch.linspace( 0, 1, 1000)
# x & t grids:
X,T = torch.meshgrid(x,t)
# x & t columns:
xcol = X.reshape(-1, 1)
tcol = T.reshape(-1, 1)
# one large column:
usol = pinn.net_u(xcol,tcol)
# reshape solution:
U = usol.reshape(x.numel(),t.numel())
# transform to numpy:
xnp = x.cpu().numpy()
tnp = t.cpu().numpy()
Unp = U.cpu().detach().numpy()

#Plotting u for PINN method
for i in range(0,1000,200):
  plt.plot(np.arange(-50,50,0.1),Unp[:,i], label='t='+str(i*0.001)+'s')
  plt.xlim(left=-10, right=10)
  plt.legend()
  plt.xlabel('x')
  plt.ylabel('u(x,t)')


# A comparison between PINN and the finite difference method
plt.plot(np.arange(-50,50,0.1),Unp[:,500], label='PINN', linestyle='dashed')
plt.plot(np.arange(-50,50,0.1), u_df[500,:], label='Finite Difference', color='red')
plt.xlim(left=-10, right=10)
plt.legend()
plt.xlabel('x')
plt.ylabel('u(x,t)')
     
\end{lstlisting}

\begin{table}[H]
\centering
\begin{tabular}{|c|c|c|}
\hline
\textbf{Phenomenon}& \textbf{Modelling} & \textbf{Remarks}                                                                  \\ \hline
unidimensional modelling       & First order differential equation & \adjustbox{stack=ll}{Simplistic approach\\ allows to have a \\score using a low quality data}                \\ \hline
Diffusion Modelling             & Heat equation                     & \adjustbox{stack=ll}{Unique/ works well without\\ data/Does not take into \\consideration the \\ pigmentation} \\ \hline
Modelling of viscous diffusion  & Burgers equation                  & Works well without data                                                           \\ \hline
Front wave transport Modelling  & Fisher KPP                        & Works well without data                                                           \\ \hline
\adjustbox{stack=ll}Modelling with non linear terms & Kordweg de Vries               & Works well without data                                                           \\ \hline
\adjustbox{stack=ll}{Turing pattern\\ Modelling}        & Turing equation                   & \adjustbox{stack=ll}{Numerical instability  
\\problem
 }                                                     \\ \hline
\end{tabular}
\caption{The table shows the different partial differential equations studied in this paper.}
\end{table}

\section{Spatial distribution extraction by computer vision}
The extraction of the spatial distribution of ulcerative colitis and Crohn's disease using computer vision can be very useful for our study. By using image processing algorithms, it is possible to accurately map out the areas affected by the two diseases. This can help doctors better understand the progression of the disease and adjust treatment accordingly. In addition, it can also allows for faster and more accurate evaluation of treatment response, which can be particularly useful in the context of testing new medications. The traditional detection methods based on the expertise of gastroenterologists are time and resource intensive. As a result, early detection and treatment of these diseases can reduce casualties and improve the patient's quality of life later on. With recent advances in Deep Learning, powerful approaches for both detection and classification that can handle complex environments have been developed. In this paper, we propose a deep learning based architecture for object classification in the context of Crohn disease. The goal is to assist the PINN framework developed above with input data for the PDEs initial conditions and boundary conditions. The proposed solution combines deep learning and tweaked transfer learning models for object classification and detection with balanced data for each image class. It can operate in a more complex environment and takes into consideration the state of the input. Its aim is to automatically detect damages, locate them and classify the disease type.

\subsection{Transfer learning for image transformation}
\begin{figure}[H]
   \centering
   \includegraphics[scale=0.9]{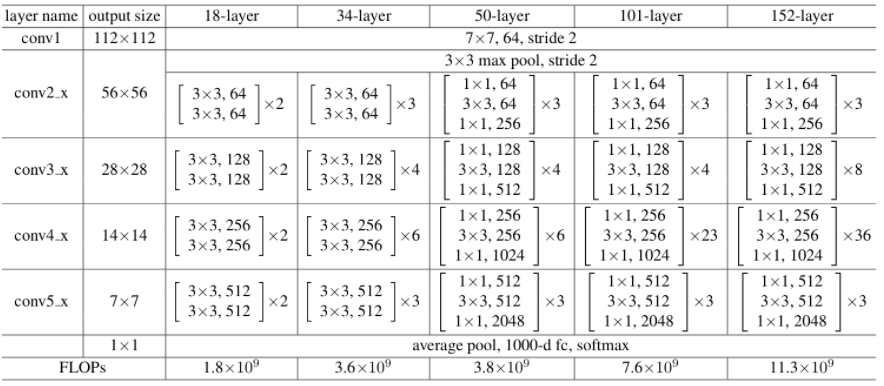}
   \caption{Resnet architecture variants.}
   \label{fig:resnet50}
\end{figure}

Transfer learning is a deep learning technique that allows the use of pre-trained models to perform image transformation tasks. The pre-trained models have already learned general features from a large dataset, making them suitable for transfer learning to related tasks. In the context of spatial distribution extraction, transfer learning can be used to improve the accuracy of feature extraction for specific image classes. The residual network ResNet \citep{deep_residual} has been used for this transfer learning taks. Its architecture is available in many different forms, each with a different number of layers, a variant with 50 layers of neural networks is referred known as Resnet50. The Resnet50 and Resnet34 architectures differ significantly in one key area. In this case, the building block was changed into a bottleneck design due to worries about the amount of time needed to train the layers. In order to create the Resnet 50 architecture, each of the Resnet34's 2-layer bottleneck blocks was changed to a 3-layer bottleneck block. Comparing this model to the 34-layer ResNet model, the accuracy of this model is noticeably higher (as shown in the figure \ref{fig:resnet50}).

On the other hand, the discrete gradient algorithm known as inpainting and the clustering techniques can then be used to fill in missing data and group similar data, respectively. Principal component analysis (PCA) can also be applied to reduce the dimensionality of the extracted features, enabling more efficient processing of the data. Ultimately, these image transformation techniques enable accurate and efficient computer vision solutions for analyzing complex biological images.

The data used to test this approach is the Kvasir dataset \citep{kvasir} which is a collection of annotated medical images of the gastrointestinal (GI) tract, designed for the purpose of computer-aided disease detection. The dataset is important for research in the medical domain of detection and retrieval, especially for single- and multi-disease computer aided detection. It contains images classified into three important anatomical landmarks and three clinically significant findings, as well as two categories of images related to endoscopic polyp removal. The sorting and annotation of the dataset are done by medical doctors. The Kvasir dataset may improve medical practice and refine health care systems globally since it includes sufficient numbers of images to be used for different tasks such as machine learning, deep learning, image retrieval, and transfer learning. The Kvasir dataset is collected using endoscopic equipment at Vestre Viken Hospital Trust in Norway and is carefully annotated by medical experts from VV and the Cancer Registry of Norway (CRN). The CRN is responsible for the national cancer screening programmes with the goal to prevent cancer death by discovering cancers or pre-cancerous lesions as early as possible. The Kvasir dataset is containing 8 classes of 2000 images per class. As a first step we opted to work with binary classification by feeding the model with both the normal images and the images having the disease.

\begin{lstlisting}[language=Python, caption=The classificastion code using a transfer learning process with the resnet50 pretrained module.]
import os
import random
import numpy as np
import torch
import torchvision
import torchvision.transforms as transforms
import matplotlib.pyplot as plt

def train_val_split(dataset, val_split=0.2, shuffle=True, random_seed=None):
    if shuffle:
        if random_seed is not None:
            random.seed(random_seed)
        dataset_size = len(dataset)
        indices = list(range(dataset_size))
        random.shuffle(indices)
    else:
        indices = list(range(len(dataset)))
    
    split = int(np.floor(val_split * len(dataset)))
    train_indices, val_indices = indices[split:], indices[:split]
    
    train_sampler = torch.utils.data.SubsetRandomSampler(train_indices)
    val_sampler = torch.utils.data.SubsetRandomSampler(val_indices)
    
    return train_sampler, val_sampler

def get_dataloaders(data_dir, image_size=(512, 512), batch_size=16, val_split=0.2, shuffle=True, random_seed=None):
    data_transforms = {
        'train': transforms.Compose([
            transforms.Resize(image_size),
            transforms.ToTensor()
        ]),
        'val': transforms.Compose([
            transforms.Resize(image_size),
            transforms.ToTensor()
        ])
    }

    image_datasets = {
        x: torchvision.datasets.ImageFolder(
            os.path.join(data_dir, x), 
            transform=data_transforms[x]
        )
        for x in ['train', 'val']
    }
    
    train_sampler, val_sampler = train_val_split(image_datasets['train'], val_split, shuffle, random_seed)
    
    dataloaders = {
        'train': torch.utils.data.DataLoader(
            image_datasets['train'], batch_size=batch_size, sampler=train_sampler
        ),
        'val': torch.utils.data.DataLoader(
            image_datasets['val'], batch_size=batch_size, sampler=val_sampler
        )
    }
    
    return dataloaders

def visualize_random_images(dataloader, classes):
    # Get a batch of training data
    inputs, labels = next(iter(dataloader))

    # Make a grid from batch
    out = torchvision.utils.make_grid(inputs)

    plt.imshow(out.permute(1, 2, 0))
    plt.title([classes[label] for label in labels])
    plt.show()

dataloaders = get_dataloaders(data_dir)
visualize_random_images(dataloaders['train'], dataloaders['train'].dataset.classes)

device = torch.device("cuda" if torch.cuda.is_available() 
                                  else "cpu")
model = models.resnet50(pretrained=True)
print(model)

for param in model.parameters():
    param.requires_grad = False
    
model.fc = nn.Sequential(nn.Linear(2048, 1024),
                                 nn.ReLU(),
                                 nn.Dropout(0.2),
                                 nn.Linear(1024, 2),
                                 nn.LogSoftmax(dim=1))
criterion = nn.NLLLoss()
optimizer = optim.Adam(model.fc.parameters(), lr=0.001)
model.to(device)

epochs = 100
steps = 0
running_loss = 0
print_every = 10
train_losses, test_losses = [], []
for epoch in range(epochs):
    for inputs, labels in trainloader:
        steps += 1
        inputs, labels = inputs.to(device), labels.to(device)
        optimizer.zero_grad()
        logps = model.forward(inputs)
        #print(logps, labels.view(-1, 1))
        loss = criterion(logps, labels)
        loss.backward()
        optimizer.step()
        running_loss += loss.item()
        
        if steps % print_every == 0:
            test_loss = 0
            accuracy = 0
            model.eval()
            with torch.no_grad():
                for inputs, labels in testloader:
                    inputs, labels = inputs.to(device),labels.to(device)
                    logps = model.forward(inputs)
                    batch_loss = criterion(logps, labels)
                    test_loss += batch_loss.item()
                    
                    ps = torch.exp(logps)
                    top_p, top_class = ps.topk(1, dim=1)
                    equals = top_class == labels.view(*top_class.shape)
                    accuracy +=torch.mean(equals.type(torch.FloatTensor)).item()
            train_losses.append(running_loss/len(trainloader))
            test_losses.append(test_loss/len(testloader))                    
            print(f"Epoch {epoch+1}/{epochs}.. "
                  f"Train loss: {running_loss/print_every:.3f}.. "
                  f"Test loss: {test_loss/len(testloader):.3f}.. "
                  f"Test accuracy: {accuracy/len(testloader):.3f}")
            running_loss = 0
            model.train()
torch.save(model, 'sicknessmodel.pth')

plt.plot(train_losses, label='Training loss')
plt.plot(test_losses, label='Validation loss')
plt.legend(frameon=False)
plt.show()

from sklearn.metrics import f1_score
f1_score(y_true, y_pred, average=None)

\end{lstlisting}

\begin{figure}[H]
  \centering
   \includegraphics[scale=1.5]{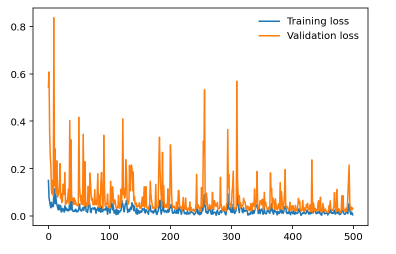}
   \caption{Resnet50 loss evolution in function of the number of epochs. The loss function used is the cross-entropy loss.}
   \label{fig:blckbx}
\end{figure}

\subsection{Inpainting and clustering}
One major issue our data had is the the endoscopic camera's light in work. In order to tackle this problem we tried firstly to inpaint the images. We have considered the areas with light on them as having missing values, so the inpainting idea, which is the task of reconstructing missing regions in an image, comes into play. It is an important problem in computer vision and an essential functionality in many imaging and graphics applications, e.g. object removal, image restoration, manipulation, re-targeting, compositing, and image-based rendering. The technique consisted of finding the white pixels in the images and dilute the surrounding pixels using a certain threshold on the pixels' values by working with a mask between 221 and 255, in order to conserve the maximum amount of information. After inpainting the images, we noticed that they still contain some noise in them. So in order to tackle this problem, we tried kmeans \citep{kmeans} for clustering. Clustering algorithms are unsupervised algorithms, meaning they don't use labelled data. They are used to assign data points from a population to different groups where data points belonging to the same group have similar traits. In our instance, clustering the image enabled us to blend the several image segments together in order to lessen the noise that persisted even after the inpainting.

\begin{lstlisting}[language=Python, caption=Implementation of the morphological mask and the inpainting transformation.]
import cv2
import numpy as np
import matplotlib.pyplot as plt

def inpaint_image(img_path, morph_size=(7,7)):
    # Load the image
    img = cv2.imread(img_path)
    if img is None:
        raise ValueError("Failed to read image from the given path")
        
    # Convert the image to grayscale
    gray = cv2.cvtColor(img, cv2.COLOR_BGR2GRAY)
    
    # Threshold the grayscale image to create a mask
    mask = cv2.threshold(gray, 220, 255, cv2.THRESH_BINARY)[1]
    
    # Perform morphological closing on the mask
    kernel = cv2.getStructuringElement(cv2.MORPH_ELLIPSE, morph_size)
    mask = cv2.morphologyEx(mask, cv2.MORPH_CLOSE, kernel, iterations=1)
    
    # Inpaint the image using the mask
    result = cv2.inpaint(img, mask, 21, cv2.INPAINT_TELEA) 
    
    # Convert the result to RGB
    image = cv2.cvtColor(result, cv2.COLOR_BGR2RGB)
    
    return image

# Specify the path to the input image
img_path = './kvasirpytorch/kvasir-dataset/ulcerative-colitis/049c2045-5259-47d8-8f9e-bbbecd81789f.jpg'

# Inpaint the image
image = inpaint_image(img_path)

# Display the result
plt.imshow(image)
plt.axis('off')

\end{lstlisting}

\begin{figure}[H]
\begin{minipage}[h]{0.47\linewidth}
\begin{center}
\includegraphics[width=1\linewidth]{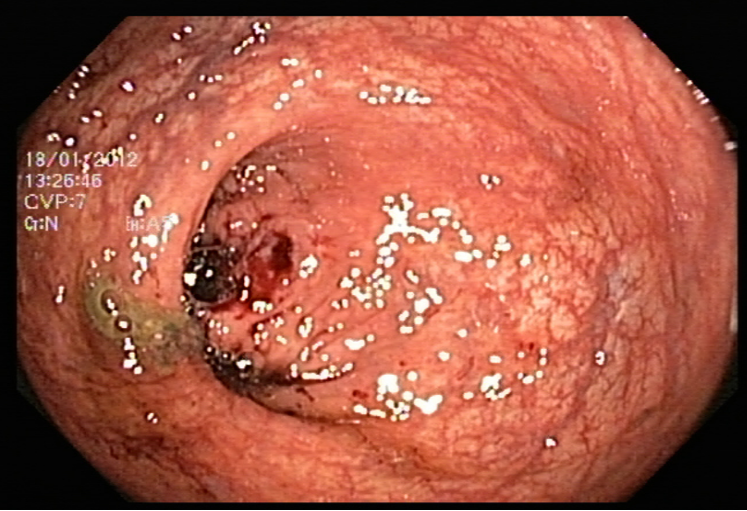} 
\label{qwe1}
\end{center} 
\end{minipage}
\hfill
\vspace{0.2 cm}
\begin{minipage}[h]{0.47\linewidth}
\begin{center}
\includegraphics[width=1\linewidth]{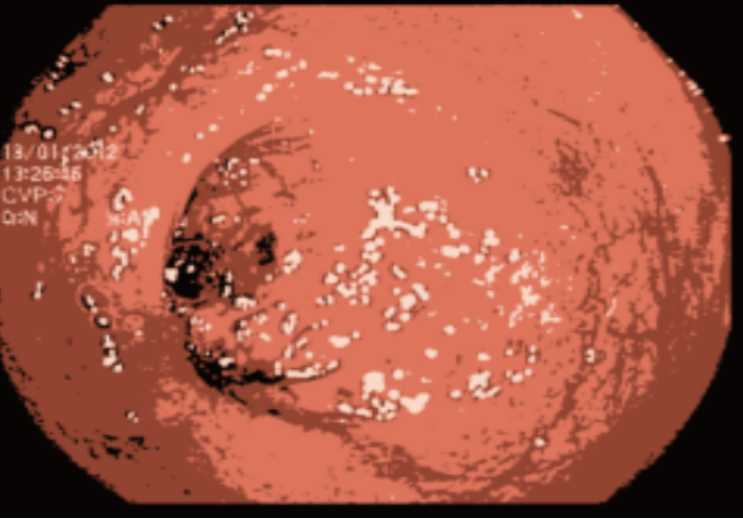} 
\label{qwe1}
\end{center}
\end{minipage}
\vfill
\vspace{0.2 cm}
\begin{minipage}[h]{0.47\linewidth}
\begin{center}
\includegraphics[width=1\linewidth]{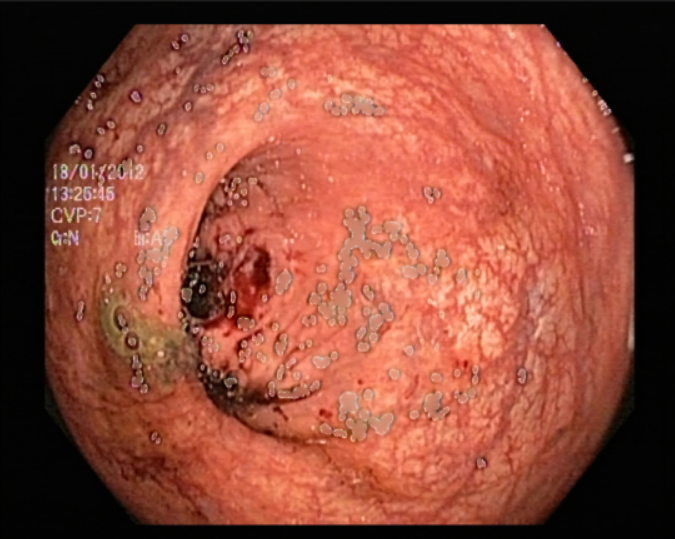} 
\label{qwe1}
\end{center}
\end{minipage}
\hfill
\begin{minipage}[h]{0.47\linewidth}
\begin{center}
\includegraphics[width=1\linewidth]{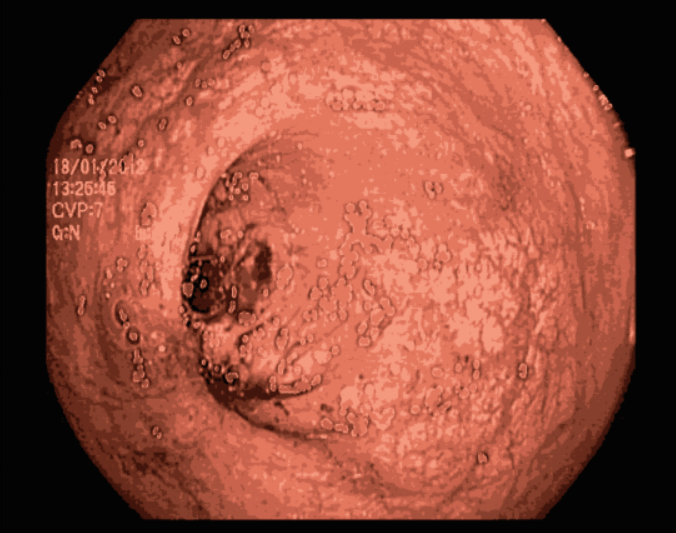} %
\label{qwe1}
\end{center}
\end{minipage}
\caption{In order from top and from left to right, the figures present: the original and real image extracted directly from the Kvasir database, the image after application of a segmentation, after application of the algorithm of inpainting and finally after applying segmentation.}
\label{fig:example}
\end{figure}

\subsection{Image classification with PCA}
Image classification using principal component analysis (PCA) was performed in this study to explore an unconventional approach to solving computer vision problems. While deep learning is commonly used to address these challenges, we opted to apply gradient boosting. Initially, we extracted the palettes of each image to obtain a correspondence table of selected colors in the RGB color space. We extracted the first five dominant color sets and transformed the images into vectors, which were arranged in a dataframe. However, before proceeding with the classification process, we applied PCA to reduce the dimensionality of the data and ensure that our points were represented optimally, thus increasing the chances of successful classification.

\begin{figure}[H]
   \centering
    \includegraphics[scale=0.8]{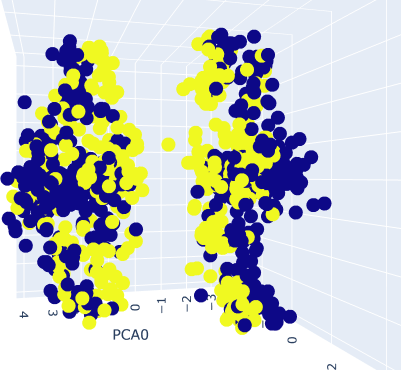}
    \caption{PCA plot of the points}
    \label{fig:blckbx}
\end{figure}

To identify the optimal number of palettes for the images, we plotted the accuracy of the model against the number of palettes used. Based on the results, we selected the optimal number of palettes for further tests. We then evaluated our gradient boosting method, specifically XGBoost \citep{xgboost}, a distributed tree boosting algorithm that utilizes the gradient boosting framework to create efficient, flexible, and portable machine learning algorithms. Overall, this study demonstrates the potential of using gradient boosting and PCA to solve computer vision problems and offers new insights into the classification of image data. This table shows the confusion matrix of our results.

\begin{center}
\begin{tabular}{ c c c }
 P/A & Positive & Negative \\ 
 Positive & 66 & 10 \\  
 Negative & 6 & 68    
\end{tabular}
\end{center}

\begin{figure}[H]
   \centering
    \includegraphics[scale=0.47]{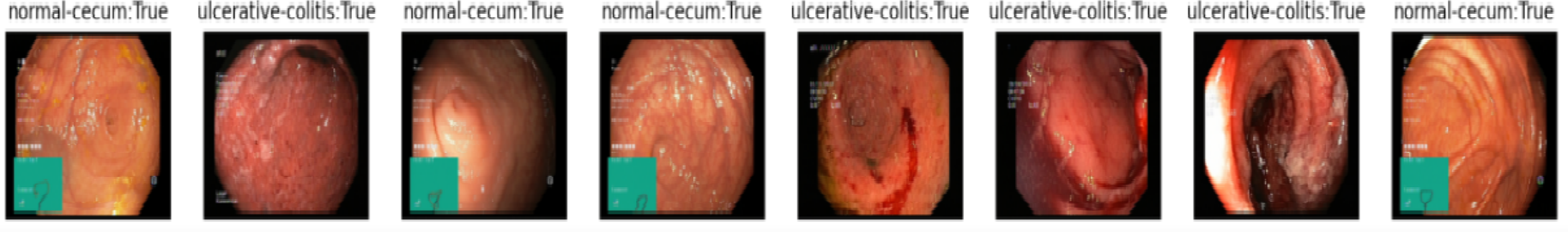}
    \caption{Results of the classification algorithm.}
   \label{fig:resultsclassification}
\end{figure}

\subsection{Computer vision solution perspectives}
In this part, we are discussing how the previous subsections are going to introduce the complete architecture of the solution. While the partial differential equations gives us a solution depending on time, classification models only provide the current state of the disease. The proposed solution would be a tailored deep learning architecture having both the regression and the classification tasks in order to get the full picture on the evolution of the disease. The regression part of the solution consists of a convolutional long short term memory (ConvLSTM). This network is provided a sequence of images depending on time and its output is the next sequence of the disease meaning the next state. With this architecture we get both the current state and its evolution both on time and surface.

\begin{figure}[H]
   \centering
    \includegraphics[scale=0.7]{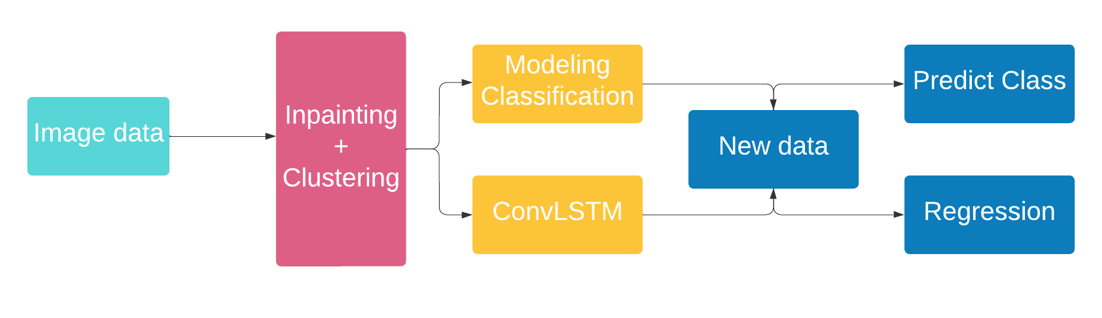}
    \caption{Computer vision proposed architecture }
    \label{fig:blckbx}
\end{figure}\textbf{}
While the previous propositions might seem a bit shallow, we propose another approach which is image segmentation.
This could be treated also as a regression problem.
We apply an image segmentation model on our images that would return the surface that has the disease on it.
The information we get is later treated as a tabular data and with sufficient data we could apply a state of the art regression model in order to determine the evolution of the surface. Keep in mind that the information provided by the segmentation model should be treated carefully because of the amount of disturbance that could be provided by the mask on the detected object.



 \section{Conclusion}
 This study was carried out with the ultimate goal of being producible and usable by the scientific community in a challenging Tunisian context. Digestive cancers are a significant public health concern due to their high mortality rates. Besides, the mortality depends on the stage at the time of diagnosis: high mortality for late stage detection vs survival rate for pre-cancerous stage detection. Ulcerative Colitis and Crohn's disease may indicate a potential evolution towards cancer. However, the current diagnostic methods for these diseases are complicated: requires expert gastroenterologists, expensive equipments, and has a major impact on the quality of life of patients (invasive) making early detection challenging. Which brings us to try to answer several open questions: Can we improve on current methods, make it less expensive and with less heavy intervention then improving the patient's life quality? 
 
 This paper aimed to demonstrate the potential of AI in providing a more cost-effective and less invasive solution for detection by solving partial differential equations that model the propagation of these diseases. Our findings provide open-source data and codes, promoting transparency and encouraging further research. Our preliminary study used unsupervised learning to model ulcerative colitis and Crohn's disease due to the lack of good quality data. Further investigations may be necessary to better understand the modeling of these diseases and the potential to combine computer vision techniques and regression models. We have shown that is possible to solve the Fisher-KPPk equation with low quality data and with a deep neural network. The difficulty was noticed rather in comparison with the Turing non-linear and coupled PDEs system. For that, an investigation is necessary to understand the phenomenon of explosion badly modeled. Finally, we believe that this work was an opportunity to bring together three communities: gastro-enterologists, mathematicians, and AI specialists through our proposed experimental framework \ref{fig:experimental_framework}.

\bibliographystyle{unsrt}
\bibliography{references}

\end{document}